\documentclass[10pt]{article} %
\usepackage[preprint]{tmlr}
\ifdefined\pdfsuppressptexinfo\pdfsuppressptexinfo=-1\fi

\usepackage{amsmath,amsfonts,bm}

\def\eqref#1{equation~\ref{#1}}

\def\1{\bm{1}}

\DeclareMathAlphabet{\mathsfit}{\encodingdefault}{\sfdefault}{m}{sl}
\SetMathAlphabet{\mathsfit}{bold}{\encodingdefault}{\sfdefault}{bx}{n}

\usepackage{graphicx}

\usepackage{amsmath,amssymb}
\usepackage{booktabs}
\usepackage{multirow}
\usepackage{graphicx}
\usepackage{caption}
\usepackage{xcolor}
\usepackage[most]{tcolorbox}
\usepackage{enumitem}
\usepackage{wrapfig}
\usepackage{booktabs}
\usepackage{booktabs,multirow,threeparttable,pifont}
\newcommand{\cmark}{\ding{51}}
\newcommand{\xmark}{\ding{55}}

\newtcolorbox{findbox}{
  enhanced,
  colback=blue!3!white,
  colframe=teal!65!black,
  boxrule=1pt,
  arc=6pt,
  left=8pt,
  right=8pt,
  top=5pt,
  bottom=5pt,
  fontupper=\itshape,
  before skip=6pt,
  after skip=6pt
}

\usepackage{hyperref}
\usepackage{url}

\title{What do MLLMs know about ``Waldo''? Evaluating Visual Reasoning under Multi-Instance Multi-Object Challenges}

\author{\name Saumya Mundra \email saumyamundra@gmail.com \\
      \addr IIIT Hyderabad
      \AND
      \name Ajoy Mondal \email ajoy.mondal@iiit.ac.in \\
      \addr IIIT Hyderabad
      \AND
      \name Vineeth N Balasubramanian \email vineethnb@iith.ac.in \\
      \addr Microsoft Research India \\
      IIT Hyderabad
      \AND
      \name C. V. Jawahar \email jawahar@iiit.ac.in \\
      \addr IIIT Hyderabad}

\def\month{MM}  %
\def\year{YYYY} %
\def\openreview{\url{https://openreview.net/forum?id=XXXX}} %
\usepackage{subcaption}
\begin{document}

\maketitle

\begin{abstract}
Recent advances in Multimodal Large Language Models (MLLMs) have led to significant progress in visual reasoning capabilities. However, existing benchmarks curate image-question pairs pertaining to object categories with few instances, enabling trivial object referencing and limited visual ambiguity. In this work, we introduce \textbf{MIMO-Bench}, a benchmark for evaluating MLLMs in a \textbf{M}ulti-\textbf{I}nstance per \textbf{M}ulti-\textbf{O}bject categories (MIMO) setting. Each image-question pair in our benchmark pertains to instances belonging to object categories with multiple instances. The presence of such similar feature distractors causes hindrance in basic visual reasoning tasks, necessitating the referring expression to be composed of multiple visual subtasks to uniquely and unambiguously identify the target object. MIMO-Bench comprises 3,802 image-question pairs across attribute binding and numerical counting tasks. Despite impressive performances on existing benchmarks, both SOTA open-source (InternVL, QwenVL) and proprietary (Gemini, GPT-5) models exhibit substantially lower performance on MIMO-Bench, highlighting their limitations in the MIMO setting. The benchmark will be made publicly available.

\end{abstract}

\vspace{-0.5cm}
\section{Introduction} \label{sec:intro}

Answering a rudimentary question such as \textit{``What is the defect on the bottle?''} is a trivial task for humans when the queried image contains one bottle. However, asking the same question upon introducing multiple bottles in the image immediately raises ambiguity, leading humans to naturally ask ``Which bottle?''. One way to counter such ambiguity is to ensure that our reference to the bottle is as detailed as possible, leading us to refine our question to \textit{``What is the defect on the blue handled bottle placed on the topmost shelf of the rightmost steel rack?''}. Humans can still answer accurately by using pre-attentive visual features~\citep{wolfe1989guided} and sequentially integrating information from the question and the visual scene~\citep{TREISMAN198097}. However, performing this task under the same time constraints often leads to illusory conjunction such as misbinding attributes like color or position.

Multimodal Large Language Models (MLLMs)~\citep{liu2023visual,wang2024qwen2,zhu2025internvl3,alayrac2022flamingo} have unified vision and language understanding, enabling remarkable progress across diverse tasks, including text-centric~\citep{mathew2021docvqa,mathew2022infographicvqa}, STEM-based~\citep{lu2023mathvista,he2024olympiadbench}, and real-world vision-centric reasoning~\citep{tong2024cambrian1,NEURIPS2024_2f8ee6a3,fu2024blink,liu2024mmbench,NEURIPS2024_1e69ff56}. Vision-centric reasoning, in particular, depends heavily on accurate visual grounding; the ability to locate and reason about specific objects in complex visual scenes.

Benchmarks such as MMStar~\citep{NEURIPS2024_2f8ee6a3}, BLINK~\citep{fu2024blink}, CV-Bench~\citep{tong2024cambrian1}, and NaturalBench~\citep{NEURIPS2024_1e69ff56} have advanced the evaluation of vision-centric reasoning, posing tasks such as \textit{numerical counting} (e.g., ``Count the number of table lamps in the image?''), \textit{spatial reasoning} (e.g., ``Is the person to the left of the pole?''), and \textit{attribute recognition} (e.g., ``What is the primary color of the car?''). However, image-question pairs in most existing benchmarks pertain to object categories containing very few instances; either because the curated images contain only a few (often only one) instances per object category, or because the questions are targeted at objects belonging to single-instance categories, despite multi-instance object categories being present. As a result, target objects are easily distinguishable, and simple referring expressions such as \texttt{<the\_object\_category>} (e.g., ``the man'' or ``the car'') suffice. While this design facilitates a controlled isolated evaluation of reasoning ability, it limits the assessment of fine-grained reasoning. As modern MLLMs such as InternVL3~\citep{zhu2025internvl3}, Qwen3-VL~\citep{bai2025qwen3vltechnicalreport}, and GPT-5~\citep{openai2025gpt5} perform basic vision tasks well, a natural question arises: how do MLLMs perform in settings featuring \textbf{M}ulti-\textbf{I}nstance per \textbf{M}ulti-\textbf{O}bject categories (\textbf{MIMO})?

\begin{figure*}[!ht]
\vspace{-0.2cm}
\caption{\textbf{Existing vs. MIMO setting.} MIMO introduces dense multi-instance scenes with more visual distractors and richer referring expressions.}
\vspace{-0.2cm}
\centerline{
\includegraphics[width=0.9\linewidth]{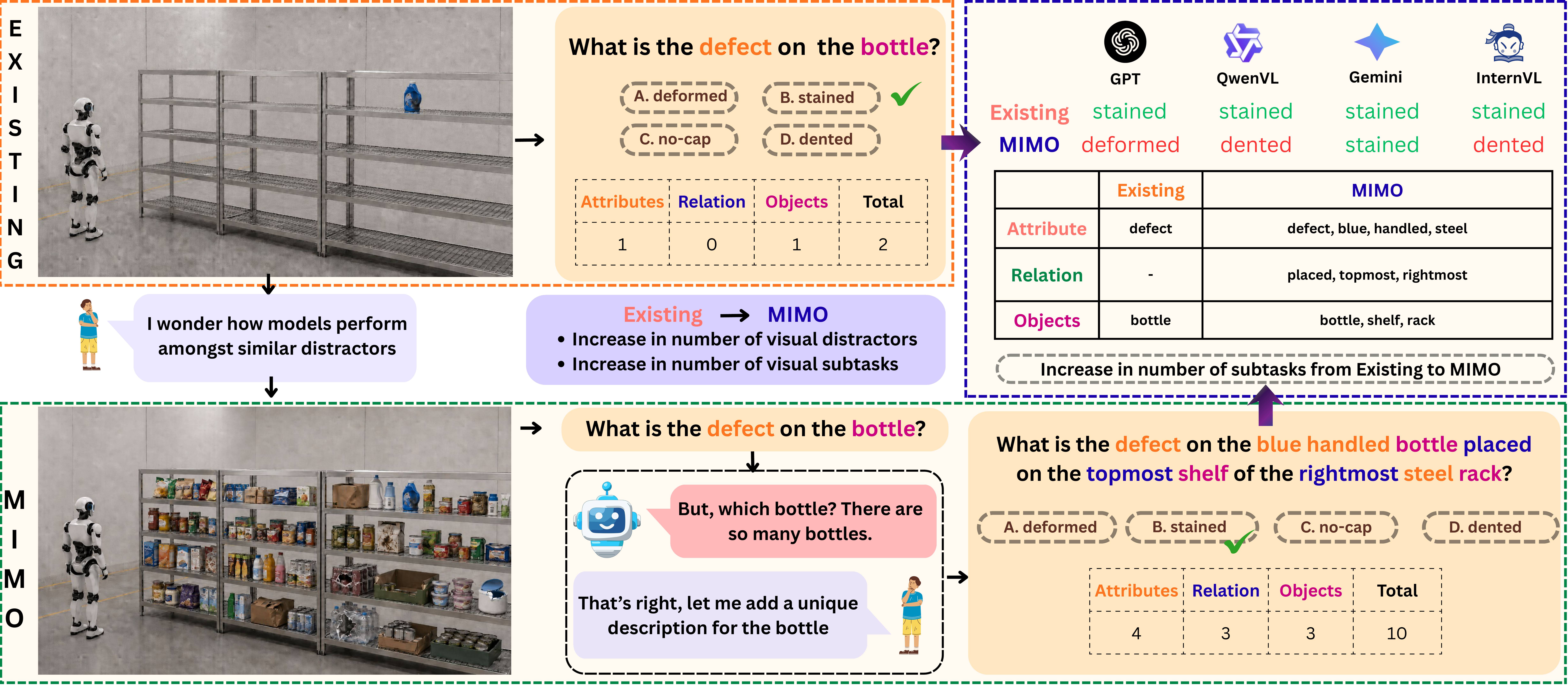}}
\vspace{-0.3cm}
\label{fig:existing_vs_mimo}
\end{figure*}

The MIMO setting is characterized by a target instance co-occurring with other instances of the same object category that act as similar-feature distractors. Presence of such distractors hurt primary visual reasoning tasks such as attribute binding and numerical counting~\citep{NEURIPS2024_cdcc6d47,izadi2025visualstructureshelpsvisual}. This reflects the classical binding problem~\citep{Roskies1999Binding} in visual cognition and suggests that a single visual reasoning task in MIMO is harder. To add to this, the MIMO setting necessitates that the referring expression be composed of multiple visual subtasks. For example, we can not simply refer to the bottle in the MIMO-image in Fig.~\ref{fig:existing_vs_mimo} as ``\textit{the bottle}''. Rather, we combine \textit{attributes/actions} and \textit{relations} of neighboring \textit{objects} to refer to the bottle uniquely as ``the \textit{blue} \textit{handled} \textit{bottle} \textit{placed} \textit{on} the \textit{top-most} \textit{shelf} \textit{of} the \textit{right-most} \textit{steel} \textit{rack}''. Such referring expressions impose multiple visual subtasks~\citep{Johnson_2017_CVPR}, including object recognition, attribute/action recognition, and relational reasoning. As a result, both the number of visual subtasks and the difficulty of each subtask amplifies in MIMO (Fig.~\ref{fig:existing_vs_mimo}).

\noindent To evaluate MLLMs under the MIMO setting, we introduce \textbf{MIMO-Bench}, a benchmark comprising 3,802 image-question pairs for attribute binding and numerical counting. MIMO-Bench consists of two primary sources of difficulty: (i) \textit{presence of similar distractors} induced by the existence of multiple instances per multiple object categories and (ii) \textit{complex referring expressions} that pose multiple visual subtasks. Standard accuracy~\citep{tong2024cambrian1,NEURIPS2024_2f8ee6a3} captures neither whether a model consistently binds all of an instance's attributes nor whether its counting errors are global or localized. Thus, we define \textbf{Group Accuracy} for attribute binding and, for numerical counting, complement \textbf{Mean Relative Accuracy (MRA)}~\citep{Yang_2025_CVPR} with \textbf{Global MRA} and \textbf{Local MRA}. To isolate the role of referring expressions, we replace them with \textbf{visual prompts} that mark the target directly, and we further show that performance degrades as scene distractors increase. Our contributions and findings are as follows:

\textbf{\textit{1. We introduce MIMO-Bench}}, a benchmark of 3,402 attribute-binding and 400 numerical-counting questions for evaluating MLLMs in dense Multi-Instance per Multi-Object scenarios.

\textbf{\textit{2. Current MLLMs struggle substantially on MIMO-Bench}}, with state-of-the-art zero-shot performance on real images reaching only about $52.8\%$ $\mathcal{I}_{\mathrm{ACC}}$ on attribute binding and $56.7\%$ $\mathcal{I}_{\mathrm{MRA}}$ on numerical counting.

\textbf{\textit{3. Referring expressions and similar-feature distractors are key bottlenecks}}, as visual prompting improves $\mathcal{I}_{\mathrm{ACC}}$ by approximately 15.5 points on average, while our graying and inpainting experiments show that reducing visual distractors further improves performance.

\textbf{\textit{4. Iterative visual access is a promising direction}}, as naive relevant-region extraction consistently improves performance across models, while Chain-of-Thought prompting does not reliably help.

\section{Related Work} \label{sec: related_work}

\noindent\textbf{Vision-centric multimodal benchmarks.}
Recent MLLM families such as LLaVA~\citep{liu2023visual}, InternVL~\citep{Chen_2024_CVPR}, QwenVL~\citep{bai2023qwen}, GPT-5~\citep{openai2025gpt5}, and Gemini-2.5-Pro~\citep{comanici2025gemini} perform strongly on general vision--language tasks, as tracked by vision-centric benchmarks where answers must be inferred from the image rather than from dataset or language priors (e.g., MMStar~\citep{NEURIPS2024_2f8ee6a3}, BLINK~\citep{fu2024blink}, RealWorldQA~\citep{xai2024realworldqa}, CV-Bench~\citep{tong2024cambrian1}, NaturalBench~\citep{NEURIPS2024_1e69ff56}). Others isolate specific difficulties: V\textsuperscript{*}Bench~\citep{Wu_2024_CVPR} targets visual search for small objects, while HR-Bench~\citep{wang2025divide} and MME-RealWorld~\citep{zhang2024mme} stress high-resolution perception. MIMO-Bench adds a new axis: dense same-category scenes that raise distractor count and force referring expressions to take on richer compositions to disambiguate the target.

\noindent\textbf{Referring expressions.}
A referring expression is classically studied in natural language generation as a \emph{distinguishing description}: a phrase singling out an intended referent from competing distractors while respecting the Gricean maxims~\citep{grice1975,dale1995}. Disambiguation against similar candidates is thus intrinsic to its purpose, while its form ranges from a bare category label to a long relational description. Vision--language work brought referring expressions into real-image grounding through ReferItGame~\citep{kazemzadeh-etal-2014-referitgame} and the MS-COCO RefCOCO/RefCOCO+/RefCOCOg suite~\citep{yu2016,Mao_2016_CVPR}. {Referring Expression Comprehension~\citep{yu2016,Chen_2025_CVPR} and related tasks, including phrase grounding~\citep{plummer2015flickr30k}, referring segmentation~\citep{hu2016segmentation,liu2023gres}, Described Object Detection~\citep{xie2023DOD}, and Referring Expression Instance Retrieval~\citep{hao2025reir}, treat reference resolution or localization as the final output. Closest to our setting, Ref-Adv~\citep{dong2026refadv} also pairs compositional expressions with same-category distractors, but averages $4.01$ distractors per case and evaluates localization itself. MIMO-Bench instead places reference resolution \emph{upstream} of attribute binding and hierarchical counting, in scenes averaging $47.7$ same-category distractors per target,\footnote{Not a like-for-like measure: Ref-Adv's $4.01$ counts its authors' curated hard distractors~\citep{dong2026refadv}, whereas ours counts every same-category instance SAM3~\citep{carion2025sam3segmentconcepts} detects per target prompt, including non-confusable ones and detector errors. Read it as a difference in scene density, not distractor difficulty.} so that resolving the reference is a prerequisite rather than the task.} To construct expressions that disambiguate a target among these distractors, we adopt the attribute--object--relation framing used to characterize fine-grained vision--language structure~\citep{aro,Ma_2023_CVPR,sugarcrepe}, composing attributes, actions, and inter-object relations into expressions that uniquely identify the referent.

\noindent\textbf{Cognitive foundations.}
MIMO-Bench's difficulties are rooted in human visual cognition. Locating a target among similar items is visual search: without pop-out, it is slow and serial~\citep{TREISMAN198097,wolfe1989guided}. A candidate's features must then be correctly integrated (attribute binding), yet under clutter humans misbind color, material, or shape, producing illusory conjunctions~\citep{treisman1982,Roskies1999Binding}. Enumeration is similarly constrained: small sets are subitized, but counting degrades as items grow and feature variability drops~\citep{kaufman1949discrimination}. VLMs inherit these failures: multi-object performance depends on the probability of feature interference among co-occurring objects~\citep{NEURIPS2024_cdcc6d47}, and counting worsens with more instances and less diversity~\citep{fsc147,countbench,tallyqa}. MIMO-Bench brings search, binding, and enumeration into dense natural scenes where same-category instances act as similar-feature distractors.

{\noindent\textbf{Internal grounding mechanisms.}
A complementary line of work studies how MLLMs internally allocate attention to visual evidence and how grounding failures emerge. Layer-wise analyses show attention progressively concentrating on task-relevant regions, while cluttered scenes can fragment attention and cause smaller targets to be under-attended~\citep{lyu2025vdc,liu2025seeingbutnot,lyu2026pade}. Other work localizes grounding failures to internal image-specific features and improves grounding through representation steering~\citep{lyu2026cns}. These approaches intervene inside the model; our interventions are behavioural, modifying the visual input through marking, graying, or cropping and measuring the resulting change in MIMO performance without identifying the internal mechanism responsible.}

\DeclareRobustCommand{\cmark}{\textcolor{green!60!black}{\ding{51}}}
\DeclareRobustCommand{\xmark}{\textcolor{red!70!black}{\ding{55}}}

\begin{table}[!ht]
\centering
\caption{\textbf{Positioning MIMO-Bench.}
Each row places a benchmark group along axes characterizing its construction:
image source (real/synthetic), template-based question generation, compositional
referring expressions (REs), image resolution (Hi/Lo), and if same-category
distractor density is a controlled design axis (\cmark), incidental ($\sim$), or
not considered (\xmark). \# subtasks is the mean number of visual
subtasks per query. {Mean Distr.\ is the mean number of
same-category instances per target, excluding the target itself; unless noted
otherwise, values are computed using our SAM3-based protocol.
$^\dagger$For Ref-Adv, we use the average same-category distractor count
reported by its authors.}}
\label{tab:benchmark_positioning}

\renewcommand{\arraystretch}{1.0}
\setlength{\tabcolsep}{3pt}

\resizebox{\columnwidth}{!}{%
\begin{tabular}{l l c c c c c c c}
\toprule
\textbf{Lineage} &
\textbf{Benchmark(s)} &
\textbf{Real/Syn} &
\textbf{Template} &
\textbf{Comp.\ REs} &
\textbf{Image Res.} &
\textbf{Distractor} &
{\textbf{Mean Distr.}} &
\textbf{\# subtasks} \\
\midrule

\multirow{2}{*}{Compositional}
& CLEVR~\citep{Johnson_2017_CVPR}
& \xmark/\cmark
& \cmark
& \cmark
& \xmark/\cmark
& \xmark
& {1.74}
& 7.2 \\

& GQA~\citep{hudson2019gqa}
& \cmark/\xmark
& \cmark
& \cmark
& \xmark/\cmark
& \xmark
& {1.67}
& 3.1 \\

\midrule

Vision-centric (general)
& \begin{tabular}[c]{@{}l@{}}
    MMStar, NaturalBench,\\
    VQAv2~\citep{goyal2017vqav2}
  \end{tabular}
& \cmark/\xmark
& \xmark
& \xmark
& \xmark/\cmark
& \xmark
& {2.49/2.11/1.46}
& 3.9/2.7/2.6 \\

\midrule

\multirow{2}{*}{Visual difficulty}
& V\textsuperscript{*}Bench
& \cmark/\xmark
& \xmark
& \xmark
& \cmark/\xmark
& $\sim$
& {0.12}
& 2.4 \\

& HR-Bench, MME-RealWorld
& \cmark/\xmark
& \xmark
& \xmark
& \cmark/\xmark
& $\sim$
& {2.01/6.54}
& 4.7/3.9 \\

\midrule

{Referring expression}
& {Ref-Adv~\citep{dong2026refadv}}
& {\cmark/\xmark}
& {\xmark}
& {\cmark}
& {\xmark/\cmark}
& {\cmark}
& {4.01$^\dagger$}
& {6.6} \\

\midrule

\textbf{Ours}
& \textbf{MIMO-Bench}
& \cmark/\cmark
& \cmark
& \cmark
& \cmark/\cmark
& \cmark
& {\textbf{47.7}}
& \textbf{9.6} \\

\bottomrule
\end{tabular}%
}

\end{table}

\section{MIMO-Bench}

    \subsection{Overview}
    
    We introduce MIMO-Bench, a benchmark to evaluate reasoning capabilities of MLLMs in scenes with multiple instances of the same category. These \emph{dense same-category instances} act as natural visual distractors and necessitate \emph{compositional referring expressions (REs)} to disambiguate them, raising their subtask count. MIMO-Bench spans two reasoning tasks,
    \textit{attribute binding} and \textit{numerical counting}, constructed from real and synthetic images.  Table~\ref{tab:benchmark_positioning} positions MIMO-Bench amongst existing benchmarks, showing MIMO-Bench treats distractor-driven ambiguity as a controlled design axis and has a high mean subtask count of 9.6.

    \subsection{Tasks}
    \label{sec:tasks_ref}

    We instantiate the MIMO setting through \textbf{attribute binding} and \textbf{numerical counting}, as defined below.

\begin{figure}[!ht]
    \vspace{-0.3cm}
    \centerline{
    \includegraphics[width=0.9\linewidth]{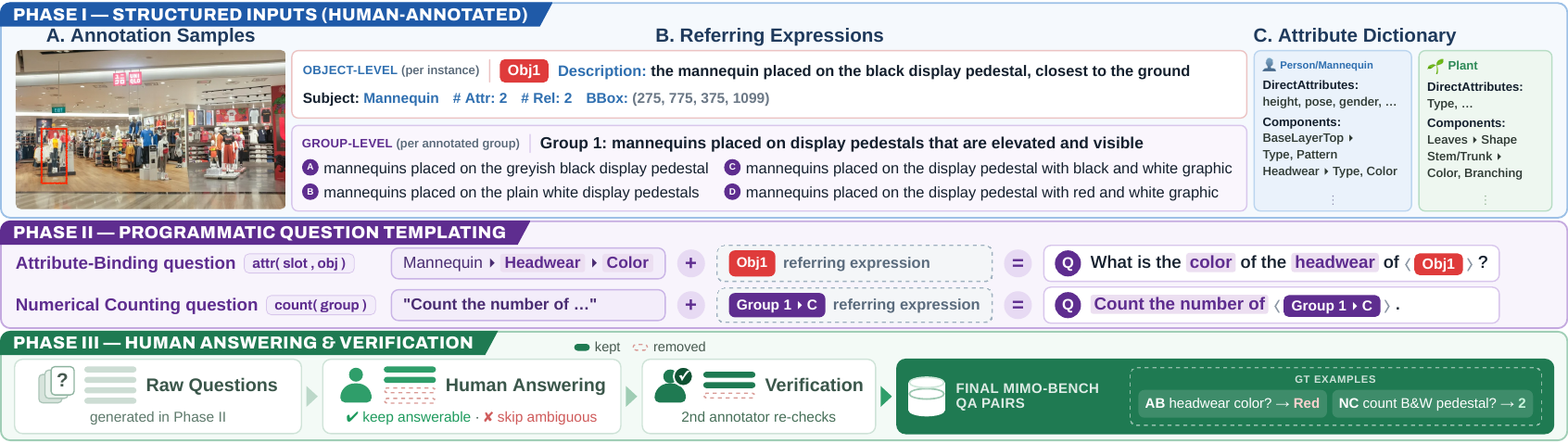}}
    \caption{\textbf{MIMO-Bench Creation Pipeline}: \textbf{AB} questions are generated by mapping instance annotations to the attribute dictionary, producing raw questions. \textbf{NC} questions are generated from REs describing groups. Human answering and verification retain only visually grounded and unambiguous QA pairs.}
    \vspace{-0.5cm}
    \label{fig:annotation_pipeline}
    \end{figure}

    \textbf{Attribute Binding (AB)} is posed as a multiple-choice task. It evaluates a model's ability to correctly associate a target object with its specified attributes and the attributes of its structured components, where the task requires attaching {\emph{each}} specified attribute to the \emph{same} target instance{\footnote{$\mathcal{G}_{\mathrm{ACC}}$ (Eq.~\ref{eq:ab_metrics}) scores each target group by the fraction of its valid attribute questions answered correctly and weights every group equally; it is not an all-attributes-correct criterion. The qualifier \emph{group} denotes the unit of aggregation, not the strictness of the criterion.}} (see Fig.~\ref{fig:annotation_pipeline}). In cognitive science, \textit{binding} refers to integrating multiple perceptual features such as color, shape, and material into a coherent object representation. Both humans and MLLMs are susceptible to the \textit{binding problem}, where the presence of visually similar distractors leads to incorrect attribution of attributes across nearby objects, motivating attribute binding as a task for our benchmark.
    
    \textbf{Numerical Counting (NC)} is posed as an open-ended counting task, where the model predicts the number of target instances matching a specified RE. Each count is defined over the region delimited by its RE, and counts are organized hierarchically so that region-bounded subgroup counts must sum to their parent group, a consistency check enforced during annotation. Prior studies~\citep{NEURIPS2024_cdcc6d47} show that low visual diversity among objects reduces counting accuracy, and human perception itself struggles when the number of objects exceeds four~\citep{kaufman1949discrimination}. By combining high feature similarity and object counts, MIMO-Bench evaluates how MLLMs perform numerical counting under multi-instance ambiguity.
    
    \noindent\textbf{Referring Expressions (REs): Necessity.}
    Both tasks are \emph{referential}: solving either requires comprehending an RE to localize the target(s) and performing the task-specific operations. A category-level reference, e.g., ``the plant'', is insufficient and ambiguous in MIMO scenes. Each target must be specified by an RE that is \emph{descriptive}, \emph{informative} and \emph{unambiguous}~\citep{grice1975}. Long expressions often become necessary and natural owing to the dense scene structure. In attribute binding, the RE identifies a single target instance, whereas in numerical counting, it specifies the group of instances to be counted.
    
\begin{figure*}[!ht]
\centering

\begin{minipage}[c]{0.53\textwidth}
    \centering
    \includegraphics[width=\linewidth]{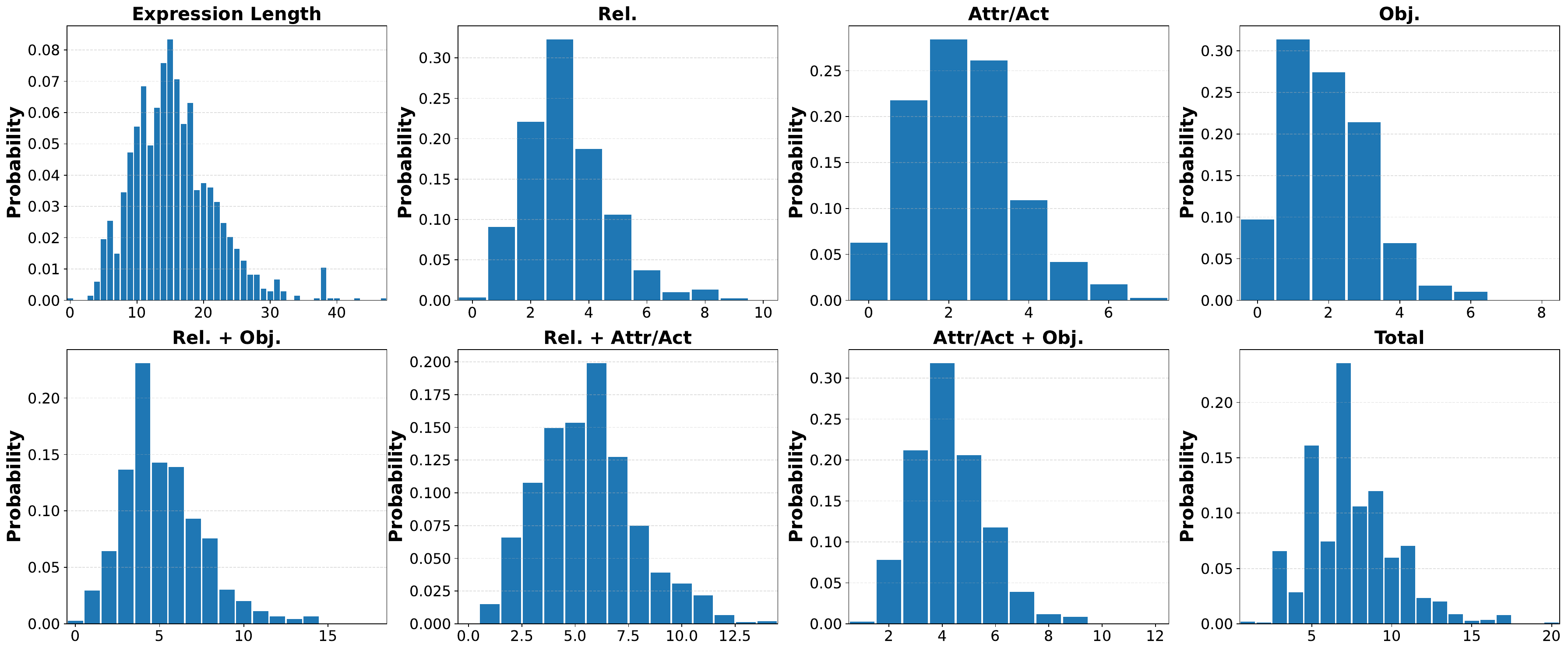}
\end{minipage}%
\hfill
\begin{minipage}[c]{0.42\textwidth}
    \centering
    \includegraphics[width=\linewidth]{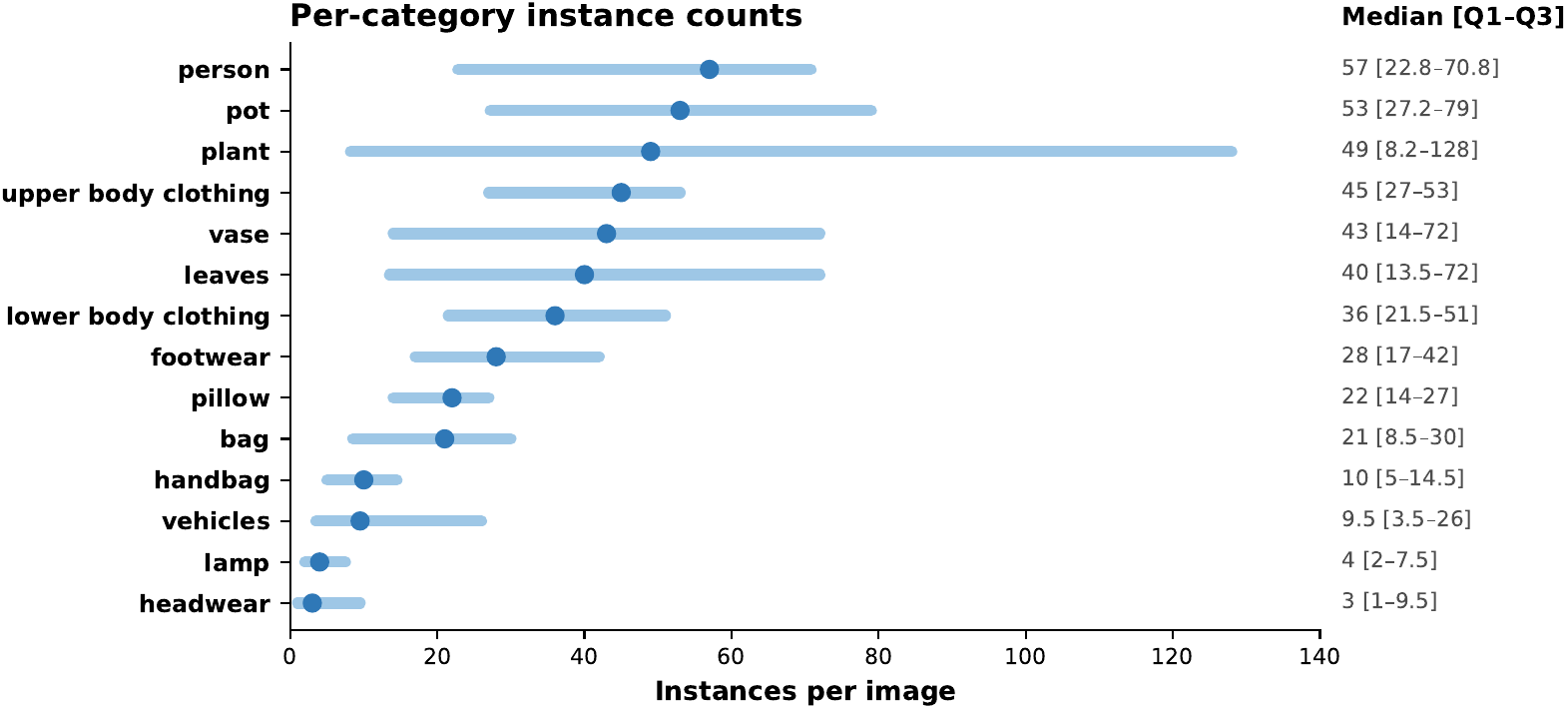}
\end{minipage}

\vspace{-0.3cm}
\caption{\textbf{Dataset statistics for MIMO-Bench.}
Distributions of RE length and composition (left), and per-category instance
counts over images containing each category (right).}
\label{fig:dataset_statistics}
\vspace{-0.2cm}
\end{figure*}

    \noindent\textbf{REs: Structure.}
    An RE may combine a target's intrinsic attributes, e.g., color, material, or size; actions or states, e.g., walking or sitting; and relations with neighboring objects. For example, \textit{``the plant whose pot is vertically striped with yellow and white''} or \textit{``the woman second closest to the person second closest to the woman near the trolley with the pink poster''}. Counting targets are specified analogously, e.g., \textit{vertically striped pillows}, \textit{plants on the topmost shelf}, or \textit{transparent utensils with purple lids placed between the gas stove and the blue kettle}. The difficulty of REs arises from their composition, a model must comprehend the attributes, actions or states, and inter-object relations it contains, and ground them to the correct object or group of objects in the scene. These subtasks correspond to components of a visual scene graph~\citep{Johnson_2017_CVPR,Ma_2023_CVPR}; they are often trivial in single-object settings but become non-trivial under MIMO conditions as REs in MIMO-Bench require solving approximately 9.6 such visual subtasks on average, twice as many as in simple references in existing vision-centric benchmarks, as shown in Tab.~\ref{tab:benchmark_positioning}. Fig.~\ref{fig:dataset_statistics} summarizes \textbf{RE length and composition} in terms of relations, attributes/actions, objects and their combinations.

    \subsection{Benchmark Construction}
    \label{sec:bench_construction}
    
    We develop a semi-automatic benchmark construction pipeline to ensure high-quality, linguistically precise, and visually grounded QA pairs, as shown in Fig.~\ref{fig:annotation_pipeline}.
    
    \noindent \textbf{Data Collection:} Our benchmark requires scenes containing multiple instances per object category. We sample real images from SA-1B~\citep{kirillov2023segment}, selecting images based on the number of annotated bounding boxes and scene depth. As SA-1B is predominantly outdoors, we supplement it with synthetic images to broaden coverage of indoor scenes, object categories, and attribute configurations that are hard to obtain at scale from real-image sources. Synthetic images are generated using Gemini-2.5-Pro~\citep{comanici2025gemini} with carefully designed prompts to emulate MIMO conditions. We also experimented with Grok-4 and Flux for image generation; however, these models produced lower realism, feature diversity, and prompt adherence compared to Gemini-2.5-Pro. A pilot across several open and closed models on Gemini-generated images showed no self-preference advantage for Gemini-2.5-Pro, with a non-Gemini model scoring highest. Data sampling, generation protocol and qualitative comparison, and self preference-bias are detailed in the appendix. Fig.~\ref{fig:dataset_statistics} reports per-category instance-count distributions over images containing that category. Categories such as person, vase, and pot contain several tens of instances per image on average, showing that targets are typically surrounded by many same-category distractors. Distractor count was calculated using SAM3~\citep{carion2025sam3segmentconcepts}, with protocol detailed in the appendix.

    \noindent \textbf{Data Annotation:} For each image, objects are selectively annotated using polygonal bounding masks{\footnote{We release these target masks and the corresponding bounding-box coordinates with MIMO-Bench, so that alternative grounding and localization interfaces can be evaluated without reconstructing the annotations.}}, with each object described by one or more REs following Sec.~\ref{sec:tasks_ref}. We take two precautions to keep these expressions operationally unambiguous and comprehensible, treating an RE as valid only when annotators can identify its intended target without disagreement and flagged cases are revised or removed. First, we use models only as a lightweight linguistic sanity check. We prompt several models to state what they understand each RE to mean, on the hypothesis that an expression which is not overly convoluted will be interpreted consistently across models.  Models agreed on RE interpretation 85\% of the time, indicating the expressions are interpreted consistently rather than being linguistically convoluted. Second, annotators who answer and verify questions flag any expression they find ambiguous or in disagreement with its referenced object. At the answering stage, $6.91\%$ of all REs were flagged; these were re-annotated to remove ambiguity or simplify the language and then returned to the verification round together with the previously-agreed expressions, after which the overall disagreement rate across the full set dropped to $2\%$, with the remaining cases discarded. The complete annotation protocol is detailed in the appendix material.

    \noindent \textbf{Question--Answer Generation:} From the prior stage, we obtain images with annotated objects, where each object is annotated with polygon masks and one or more REs. For each annotated object category, GPT-5~\citep{openai2025gpt5} is used as an auxiliary tool to propose candidate attribute keys for a hierarchical attribute dictionary describing both object-level attributes and attributes of its components, as shown in Fig.~\ref{fig:annotation_pipeline}. The dictionary is manually refined to ensure broad coverage and mutual exclusivity among attributes.
    
    Each object--RE pair is matched to the corresponding object category key in the attribute dictionary, and this match is passed through a template-based question generator, generating raw unanswered multiple-choice questions pertaining to dictionary attributes. The options are hard negatives to avoid ambiguity. For numerical counting, we append \textit{``Count the number of''} to annotated sets to generate questions. These generated unanswered questions are passed through an iterative human answering and verification pipeline. In the first answering round, only those candidate questions are answered whose queried attribute or object is physically present, clearly visible, and unambiguous in the image, and candidate questions over absent or ambiguous attributes are discarded before verification. Annotators are instructed to discard only on these grounds rather than arbitrarily, so that skipping acts as the principal filter from the exhaustive raw pool to visually grounded QA pairs. {The criterion is whether an answer is visually recoverable, not whether the question is easy: candidates were never discarded for being difficult or for being answered incorrectly, so dense scenes and complex REs remain in the benchmark. Thus, the human baseline simply measures human capability on MIMO-Bench, rather than reflecting a benchmark made easier through difficulty-based filtering.} All remaining QA pairs undergo an additional round of human verification to ensure correctness and minimize ambiguity, resulting in a high-quality benchmark. We discuss the human answering in more detail in the appendix, along with MIMO-Bench scalability.

\textbf{MIMO-Bench} contains \textit{3,402 attribute-binding} questions and \textit{400 numerical-counting} questions. Each attribute-binding QA pair corresponds to an image--object--RE--attribute tuple, and each numerical-counting QA pair to an image--target-group--RE tuple.

\section{Evaluation on MIMO-Bench} \label{sec:eval}

\subsection{Experiment Setup}

We evaluate API-accessed models Gemini-3.5-Flash~\citep{google2026gemini35flash}, Gemini-3.1-Pro-Preview~\citep{google2026gemini31pro}, Gemini-2.5-Pro/Flash~\citep{comanici2025gemini}, GPT-5~\citep{openai2025gpt5}, GLM-4.5V~\citep{v2025glm}, IVL3.5-241B-A28B~\citep{wang2025internvl3}, and IVL3-78B~\citep{zhu2025internvl3}, and locally evaluated open-weight models Qwen3-VL (Q3)~\citep{bai2025qwen3vltechnicalreport}, Qwen2.5-VL (Q2.5)~\citep{bai2025qwen2}, the remaining InternVL3/3.5 (IVL3/IVL3.5) variants~\citep{zhu2025internvl3,wang2025internvl3}, and GLM-4.1V-9B-Thinking~\citep{v2025glm}. We append size and variant to family abbreviations, e.g.\ Q3-32B-Instruct, IVL3.5-8B, G3.5-Flash. All experiments use the same task prompt and answer-format instruction across models. Model-specific components are limited to official HuggingFace/API chat templates, image preprocessing routines, and default model-card settings, unless specified otherwise. We did not observe answer-format violations; transient API failures were rerun. {We also report $95\%$ bootstrap confidence intervals ($20{,}000$ replicates), with the resampling protocol and analyses across different granularity levels detailed in Appendix~\ref{app:uncertainty}.}

\begin{figure}[!ht]
\centering

\includegraphics[width=0.95\linewidth]{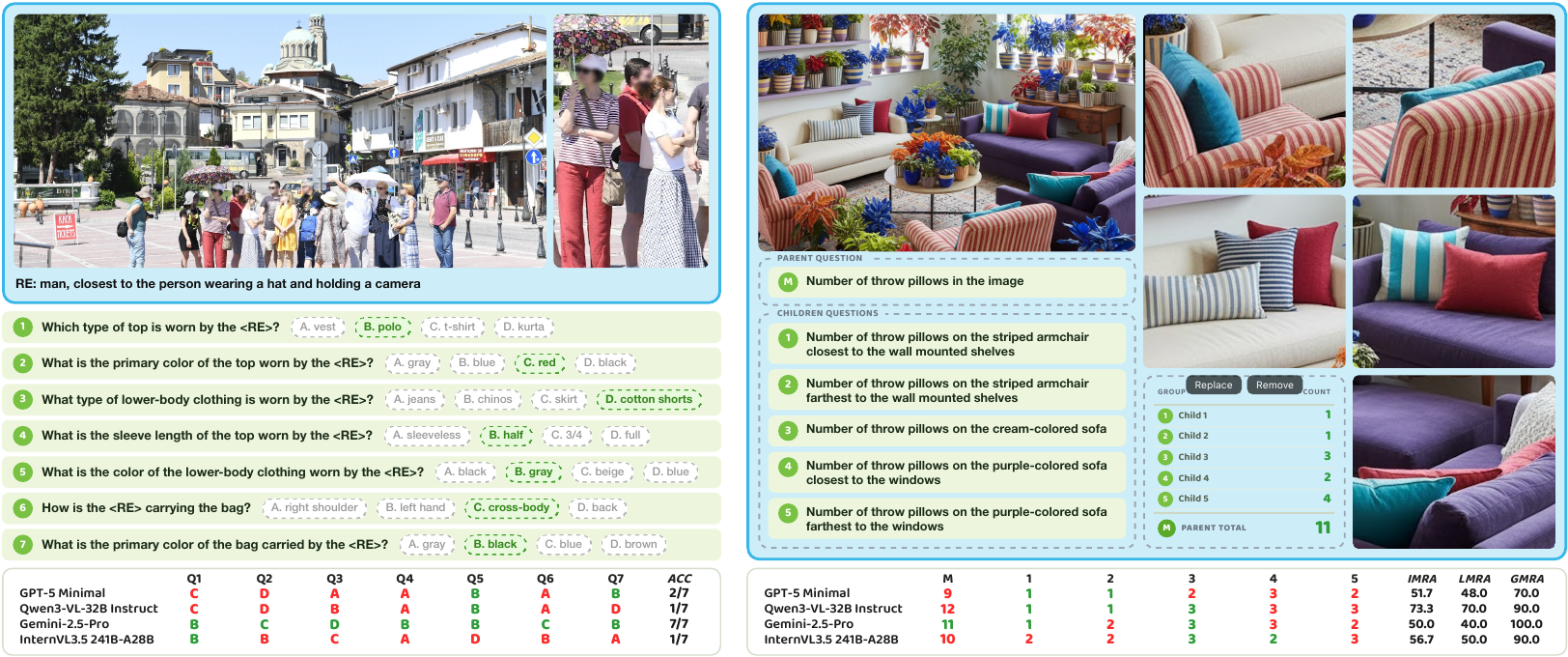}
\hfill

\caption{\textbf{AB} (left) and \textbf{NC} (right) tasks centered on a target instance or target group within an image.}
\label{fig:attribute_counting}
\vspace{-0.5cm}
\end{figure}

\subsection{Evaluation Metrics}\label{sec:eval_metrics}

\noindent\textbf{Attribute Binding.}
We adopt Instance Accuracy ($\mathcal{I}_{\mathrm{ACC}}$) as one of the evaluation metrics
for AB. We refer to it as instance accuracy because each
attribute-binding question corresponds to a single
image--object--referring-expression--attribute quadruplet, which we treat as one instance.
$\mathcal{I}_{\mathrm{ACC}}$ is defined in Eq.~\ref{eq:ab_metrics}, where
$N_A$ denotes the number of correct answers and $N_Q$ represents the total number of
questions.

To evaluate whether MLLMs form a \textit{unified} and \textit{grounded} representation of an object, we extend the evaluation beyond individual attributes. For a given instance in an image, we generate a \textit{set} of AB image-question pairs pertaining to attributes that are visually \textit{present}, \textit{objective}, and define the object \textit{uniquely} in the scene. The binding problem posits that models may fail to associate multiple attributes with a single instance. We introduce a Group Accuracy ($\mathcal{G}_{\mathrm{ACC}}$) metric, similar to~\citep{NEURIPS2024_1e69ff56}, which {aggregates per-attribute correctness within each target group while weighting target groups equally}.
When an instance is characterized by only a few objective attributes, each attribute becomes crucial for identification as even a single misbinding would significantly impair the model's ability to correctly recognize the target object, as compared to objects with multiple identifiable attributes. Group Accuracy imposes a stronger penalty in the case when the objects are defined by few attributes, on account of groups being equally weighted.
Formally:
{\setlength{\abovedisplayskip}{3pt}
\setlength{\belowdisplayskip}{3pt}
\setlength{\abovedisplayshortskip}{3pt}
\setlength{\belowdisplayshortskip}{3pt}
\begin{equation}
\label{eq:ab_metrics}
\mathcal{I}_{\mathrm{ACC}} = \frac{N_A}{N_Q}, \quad
\mathcal{G}_{\mathrm{ACC}} = \frac{1}{N_H}\sum_{h=1}^{N_H} \mathcal{H}_{\mathrm{ACC}}^{(h)}, \quad
\mathcal{H}_{\mathrm{ACC}}^{(h)} = \frac{N_{HA}^{(h)}}{N_{HQ}^{(h)}},
\end{equation}
}
where $\mathcal{H}_{\mathrm{ACC}}^{(h)}$ is the accuracy of an individual group, $N_H$ denotes the total number of groups in the benchmark, $N_{HA}^{(h)}$ and $N_{HQ}^{(h)}$ represent the number of correct answers and total questions within group $h$, respectively. {Additionally, for completeness we also report a strict Exact Group Accuracy, which counts a target as correct only when all of its valid attribute questions are answered correctly, in Appendix~\ref{app:exact_group_accuracy}.}

\noindent\textbf{Numerical Counting} performance is evaluated using the Mean Relative Accuracy ($\mathcal{MRA}$) metric~\citep{Yang_2025_CVPR}. Given a prediction $\hat{y}$, ground truth $y$, and a confidence threshold $\theta$, a prediction is considered correct if its relative error $\frac{|\hat{y} - y|}{y}$ is smaller than $1 - \theta$. Formally,
\begin{equation}
\label{eq:mra}
\resizebox{0.92\linewidth}{!}{$
\mathcal{MRA}(\hat{y},y)=
\frac{1}{|C|}\sum_{\theta \in C}
\mathbb{1}\!\left(\frac{|\hat{y}-y|}{y}<1-\theta\right),
\quad
{\scriptstyle C=\{0.5,0.55,\ldots,0.95\}},
\quad
\scriptstyle \mathcal{MRA}(\hat{y},0)=\mathbb{1}(\hat{y}=0)
$}
\end{equation}
This convention treats zero-count questions as exact-match cases, preventing empty target sets from being discarded while avoiding division by zero. $\mathcal{MRA}$ provides an instance-level assessment of counting accuracy.

We utilize two simple additional measures to understand more about a model's NC capability in the MIMO setting. To do so, we structure the NC questions hierarchically by decomposing counting queries into spatially localized sub-questions. Consider the query: \textit{``Count the number of throw pillows in the image.''}. We decompose this parent query into a set of child queries, each corresponding to a distinct subregion where the target instances may occur, such as \textit{``Count the number of throw pillows on the purple-colored sofa closest to the windows.''} and \textit{``Count the number of throw pillows on the striped armchair farthest to the wall mounted shelves.''}, as shown in Fig.~\ref{fig:attribute_counting}. At the annotation level, the global count satisfies $N = \sum_k N_k$, where $N$ denotes the count over the full image and $N_k$ denotes the count in the $k$-th localized subregion.

Using this structure, we measure an Individual Mean Relative Accuracy ($\mathcal{I}_{\mathrm{MRA}}$), a Global Mean Relative Accuracy ($\mathcal{G}_{\mathrm{MRA}}$), and a Local Mean Relative Accuracy ($\mathcal{L}_{\mathrm{MRA}}$). $\mathcal{I}_{\mathrm{MRA}}$ averages $\mathcal{MRA}$ over all numerical-counting queries, $\mathcal{G}_{\mathrm{MRA}}$ averages $\mathcal{MRA}$ over parent/global queries, and $\mathcal{L}_{\mathrm{MRA}}$ averages $\mathcal{MRA}$ over localized child queries under each parent. They are defined as
\begingroup
\setlength{\abovedisplayskip}{0.35em}
\setlength{\belowdisplayskip}{0.35em}
\setlength{\abovedisplayshortskip}{0.15em}
\setlength{\belowdisplayshortskip}{0.15em}
\begin{equation}
\label{eq:nc_metrics}
\begin{gathered}
\mathcal{I}_{\mathrm{MRA}} =
\frac{1}{|\mathcal{Q}_{\mathrm{NC}}|}
\sum_{q \in \mathcal{Q}_{\mathrm{NC}}} \mathcal{MRA}(q), \\[0.25em]
\mathcal{G}_{\mathrm{MRA}} =
\frac{1}{|\mathcal{P}|}
\sum_{p \in \mathcal{P}} \mathcal{MRA}(q_p)
\qquad
\mathcal{L}_{\mathrm{MRA}} =
\frac{1}{|\mathcal{P}|}
\sum_{p \in \mathcal{P}}
\left(
\frac{1}{|\mathcal{C}(p)|}
\sum_{c \in \mathcal{C}(p)} \mathcal{MRA}(q_c)
\right).
\end{gathered}
\end{equation}
\endgroup
where $\mathcal{Q}_{\mathrm{NC}}$ denotes all numerical-counting queries, $\mathcal{P}$ denotes the set of parent queries, $q_p$ denotes the parent query corresponding to parent $p$, and $\mathcal{C}(p)$ denotes the set of localized child queries associated with $p$. Importantly, $\mathcal{L}_{\mathrm{MRA}}$ does not evaluate whether a model's predicted child counts sum to its predicted parent count; instead, it isolates accuracy on localized child queries. Comparing $\mathcal{G}_{\mathrm{MRA}}$ and $\mathcal{L}_{\mathrm{MRA}}$ therefore reveals whether a model performs better when counting over the full scene or subregion respectively.

\subsection{Result Analysis}
\begin{table*}[!t]
\centering
\tiny
\caption{Evaluation results on MIMO-Bench, with AB and NC reported separately
for real and synthetic splits; scores are percentages. \textbf{Baselines}: random chance is the mean $1/N$ over answer options, and majority guess picks the most common answer per attribute
type. ``Low'', ``medium'' \& ``minimal'' are reasoning efforts for the respective models. Abbreviations: Syn.\ = synthetic; -- = NA}
\label{tab:experiment_real_synth}

\vspace{-0.3cm}
\setlength{\tabcolsep}{2.1pt}
\renewcommand{\arraystretch}{1.02}

\resizebox{\textwidth}{!}{%
\begin{tabular}{@{}l l | cc cc || ccc ccc@{}}
\toprule
\multirow{3}{*}{\textbf{Method}} & \multirow{3}{*}{\textbf{Variant}}
& \multicolumn{4}{c||}{\textbf{Attribute Binding (AB)}}
& \multicolumn{6}{c}{\textbf{Numerical Counting (NC)}} \\
\cmidrule(lr){3-6}\cmidrule(lr){7-12}
& & \multicolumn{2}{c}{\textbf{Real}}
& \multicolumn{2}{c||}{\textbf{Syn.}}
& \multicolumn{3}{c}{\textbf{Real}}
& \multicolumn{3}{c}{\textbf{Syn.}} \\
\cmidrule(lr){3-4}\cmidrule(lr){5-6}\cmidrule(lr){7-9}\cmidrule(lr){10-12}
& & $\mathcal{I}_{\mathrm{ACC}}$
& $\mathcal{G}_{\mathrm{ACC}}$
& $\mathcal{I}_{\mathrm{ACC}}$
& $\mathcal{G}_{\mathrm{ACC}}$
& $\mathcal{I}_{\mathrm{MRA}}$
& $\mathcal{L}_{\mathrm{MRA}}$
& $\mathcal{G}_{\mathrm{MRA}}$
& $\mathcal{I}_{\mathrm{MRA}}$
& $\mathcal{L}_{\mathrm{MRA}}$
& $\mathcal{G}_{\mathrm{MRA}}$ \\
\midrule

\multicolumn{12}{@{}l}{\textbf{Chance \& Human Baselines}} \\[-0.25ex]
\midrule
Chance(Random) &
& 15.8 & 15.6 & 15.8 & 15.3 & \multicolumn{6}{c}{--} \\
Chance(Majority) &
& 34.3 & 35.5 & 34.3 & 33.1 & \multicolumn{6}{c}{--} \\
Human &
& 89.3 & 86.7 & 85.7 & 87.5 & 87.7 & 86.4 & 90.1 & 89.6 & 87.4 & 92.0 \\

\midrule
\multicolumn{12}{@{}l}{\textbf{API-accessed MLLMs}} \\[-0.25ex]
\midrule
\multirow{2}{*}{Gemini-3.x}
& 3.5 Flash Low        & 45.5 & 44.8 & \textbf{73.0} & \textbf{75.6} & \underline{52.5} & 49.4 & 46.7 & \textbf{66.9} & \textbf{66.0} & \textbf{67.2} \\
& 3.1 Pro Preview Low  & 46.5 & 46.3 & \underline{71.8} & \underline{73.5} & \textbf{56.7} & \textbf{55.8} & \underline{63.3} & \underline{64.0} & \underline{62.6} & \underline{64.1} \\
\midrule

\multirow{2}{*}{Gemini-2.5}
& Flash & \underline{51.3} & \underline{51.9} & 48.1 & 42.1 & 30.0 & 28.3 & 38.3 & 47.2 & 41.2 & 51.4 \\
& Pro   & 49.1 & 49.5 & 47.2 & 44.9 & 37.3 & 24.2 & \textbf{71.7} & 56.1 & 53.2 & 63.9 \\
\midrule

\multirow{2}{*}{GPT-5}
& minimal & 32.0 & 31.2 & 28.0 & 24.1 & 45.8 & 47.5 & 36.7 & 43.0 & 40.8 & 47.1 \\
& medium  & \textbf{52.8} & \textbf{53.4} & 49.8 & 48.1 & 51.5 & \underline{55.0} & 56.7 & 47.5 & 42.8 & 54.3 \\
\midrule

GLM
& 4.5V             & 50.3 & 50.0 & 41.7 & 33.8 & 33.9 & 36.2 & 38.3 & 47.3 & 48.5 & 38.6 \\
\midrule

InternVL3.5
& 241B-A28B & 47.6 & 45.2 & 43.3 & 42.9 & 47.8 & 50.0 & 39.5 & 51.2 & 50.6 & 50.1 \\
\midrule

InternVL3
& 78B & 40.2 & 40.3 & 37.6 & 35.0 & 38.9 & 43.3 & 38.3 & 36.6 & 34.2 & 34.3 \\

\midrule
\multicolumn{12}{@{}l}{\textbf{Locally evaluated open-weight MLLMs}} \\[-0.25ex]
\midrule
\multirow{3}{*}{Qwen3-VL}
& 32B-Instruct      & 47.7 & 48.5 & 47.0 & 45.0 & 38.5 & 35.4 & 55.0 & 53.2 & 54.8 & 49.8 \\
& 8B-Instruct       & 42.4 & 42.0 & 36.2 & 32.5 & 34.2 & 28.3 & 43.3 & 47.6 & 48.9 & 35.5 \\
& 30B-A3B-Instruct  & 36.6 & 33.2 & 28.2 & 27.1 & 25.0 & 35.4 & 20.0 & 41.0 & 44.5 & 38.8 \\
\midrule

\multirow{3}{*}{Qwen2.5-VL}
& 72B-Instruct & 38.6 & 38.7 & 50.6 & 45.1 & 30.8 & 35.8 & 23.3 & 46.2 & 47.0 & 43.1 \\
& 32B-Instruct & 38.4 & 36.9 & 41.2 & 41.0 & 28.9 & 28.3 & 28.3 & 42.4 & 40.0 & 47.2 \\
& 7B-Instruct  & 35.5 & 35.2 & 36.4 & 34.3 & 27.3 & 30.0 & 31.7 & 33.9 & 33.5 & 27.2 \\
\midrule

\multirow{3}{*}{InternVL3.5}
& 38B       & 40.9 & 40.3 & 46.5 & 44.6 & 27.3 & 35.4 & 20.0 & 42.5 & 44.3 & 42.0 \\
& 14B       & 32.1 & 31.7 & 39.0 & 40.1 & 31.1 & 37.5 & 16.7 & 42.5 & 38.9 & 47.5 \\
& 8B        & 38.6 & 38.1 & 44.2 & 43.7 & 36.1 & 45.8 & 13.3 & 44.2 & 41.1 & 45.1 \\
\midrule

\multirow{3}{*}{InternVL3}
& 38B & 42.3 & 42.5 & 48.3 & 44.7 & 33.2 & 49.0 & 20.0 & 45.6 & 45.3 & 39.6 \\
& 14B & 35.9 & 35.1 & 39.7 & 38.9 & 26.8 & 32.5 & 6.7 & 37.1 & 34.8 & 40.7 \\
& 8B  & 34.4 & 35.5 & 38.9 & 37.2 & 26.7 & 36.7 & 5.7 & 37.3 & 41.3 & 24.3 \\
\midrule

GLM
& 4.1V-9B-Thinking & 34.6 & 36.2 & 38.7 & 35.3 & 27.3 & 25.0 & 50.0 & 24.9 & 32.6 & 20.0 \\
\bottomrule
\end{tabular}%
}

\vspace{-0.8cm}
\end{table*}

\noindent\textbf{Baselines and the human--model gap.}
We anchor all model scores against three references in Table~\ref{tab:experiment_real_synth}:
a random-chance baseline (mean $1/N$ over answer options),
a majority-guess baseline (most common answer per question type), and a human baseline\footnote{Performed by six humans who did not annotate MIMO-Bench, each  solving 25 groups for AB and 15 for NC}. Humans operate near ceiling on both
tasks and both splits, suggesting that the
benchmark is far from saturated. The gap to the best model is large everywhere:
on real AB the strongest model ($52.8$) trails humans by roughly
$36$ points, on real NC the strongest one-shot global counter
($\mathcal{G}_{\mathrm{MRA}}=71.7$) remains about $18$ points below humans, and
several open-weight models sit at or barely above the majority baseline. The only
scores approaching the human range are the two G3.x models on synthetic
AB ($73.0$, $71.8$); we treat these as an anomaly, since the same two models are mid-pack on real attribute
binding ($45.5$, $46.5$) and no other model shows the effect.
{The self-preference check in Appendix~\ref{app:self_preference} establishes only that the generating model, G2.5-Pro, is not advantaged on its own images; neither G3.x model generated the synthetic split, so that control does not account for their gap. Across alternative option sets, shuffled answer options and deliberately biased option configurations, their synthetic performance moves by at most $3$ points, so the effect is robust to the tested perturbations but remains an observed anomaly rather than evidence of genuine capability. We discuss more in Appendix~\ref{app:self_preference}.}

\noindent\textbf{Attribute binding is far from being solved, and the weakest models barely beat chance predictors.}
On real images the ceiling is low: no model exceeds $52.8\%$
$\mathcal{I}_{\mathrm{ACC}}$ (GPT-5-medium), with G2.5-Flash ($51.3$) and
GLM-4.5V ($50.3$) close behind---all far below the human baseline
(Table~\ref{tab:experiment_real_synth}). The only scores that rise much higher are
the two G3.x models on synthetic ($73.0$, $71.8$), which we flag as anomalous
since those same models are mid-pack on real (discussed below). Against a
random-choice baseline of $15.8\%$ and a majority-answer baseline of $34.3\%$
(on both the real and synthetic splits), the gains are modest: several models
(Q3-30B-A3B-Instruct on both splits, GPT-5-minimal on both splits) fall at or near the majority baseline, adding little over always
predicting the most frequent answer.

\noindent \textbf{Numerical Counting} requires forming the correct target set from an RE, excluding same-category distractors, and
enumerating valid instances. Table~\ref{tab:experiment_real_synth}
shows models differ in whether they favor global ($\mathcal{G}_{\mathrm{MRA}}$)
or localized ($\mathcal{L}_{\mathrm{MRA}}$) counting. G2.5-Pro attains the highest real-split
$\mathcal{G}_{\mathrm{MRA}}$ ($71.7$), while on the synthetic split the
G3.x models lead ($67.2$ for G3.5-Flash-Low, $64.1$ for
G3.1-Pro-Preview-Low). Several locally evaluated open-weight models benefit from localized decomposition,
e.g.\ IVL3-38B ($\mathcal{L}_{\mathrm{MRA}}{=}49.0$ vs.\
$\mathcal{G}_{\mathrm{MRA}}{=}20.0$ real) and IVL3.5-8B ($45.8$ vs.\
$13.3$ real), suggesting that reducing the search space aids target-set
formation when full-image counting is hard. A parent-level gap analysis
(Appendix~\ref{app:additional_results}) confirms this is not an aggregation artifact, and balanced
success across both remains rare. Real images are harder mainly for
instance-level and localized counting: G2.5-Pro drops from
$\mathcal{I}_{\mathrm{MRA}}{=}56.1$/$\mathcal{L}_{\mathrm{MRA}}{=}53.2$
(synthetic) to $37.3$/$24.2$ (real), with Q3-32B-Instruct showing a
similar decline. NC trends are not explained by scale, as gains are
non-monotonic across Qwen and InternVL variants.

\begin{tcolorbox}[
  colback=blue!5,
  colframe=blue!55,
  boxrule=0.4pt,
  arc=2pt,
  left=4pt,
  right=4pt,
  top=3pt,
  bottom=3pt,
]
Both AB and NC in dense same-category scenes remain far from solved. The gap to humans is largest on real images, where the strongest models trail substantially on both tasks.
\end{tcolorbox}

\begin{wraptable}{r}{0.35\textwidth}
\vspace{-0.5cm}
\centering
\caption{
$\mathcal{I}_{\mathrm{ACC}}$ on MIMO-Bench and MIMO-Crop 512/768,
and standard low-res benchmarks. $^{\dagger}$Mean of MMStar,
RealWorldQA, and MM-Bench. I: Instruct variant.}
\label{tab:mimocrop}
\vspace{-0.3cm}

{\scriptsize
\setlength{\tabcolsep}{2.2pt}
\renewcommand{\arraystretch}{1.02}
\begin{tabular}{@{}l|ccc|c@{}}
\toprule
Model &
\shortstack{MIMO\\Bench} &
\shortstack{Crop\\768} &
\shortstack{Crop\\512} &
\shortstack{Existing\\Avg.$^{\dagger}$} \\
\midrule
Q3-32B-I        & 47.5 & 55.0 & 48.9 & 81.4 \\
Q3-30B-A3B-I    & 33.8 & 42.2 & 44.9 & 77.3 \\
Q3-8B-I         & 40.3 & 48.9 & 44.4 & 75.6 \\
Q2.5-72B-I      & 42.6 & 45.7   & 48.3 & 78.3 \\
Q2.5-32B-I      & 39.3 & 48.5 & 45.0 & 74.7 \\
Q2.5-7B-I       & 35.8 & 42.8 & 47.6 & 71.7 \\
IVL3.5-38B    & 42.8 & 51.2 & 55.5 & 79.5 \\
IVL3.5-14B    & 34.4 & 46.9 & 46.4 & 74.1 \\
IVL3.5-8B     & 40.5 & 48.4 & 50.7 & 72.1 \\
IVL3-78B      & 39.3 & 47.9 & 49.5 & 79.4 \\
IVL3-38B      & 44.3 & 52.8 & 56.7 & 78.0 \\
IVL3-14B      & 37.2 & 51.1 & 50.7 & 74.3 \\
IVL3-8B       & 35.9 & 44.7 & 46.8 & 73.6 \\
\bottomrule
\end{tabular}
}

\vspace{-0.5cm}
\end{wraptable}

\noindent\textbf{MIMO-Crop.}
Most existing benchmarks~\citep{tong2024cambrian1,NEURIPS2024_2f8ee6a3,fu2024blink,liu2024mmbench,NEURIPS2024_1e69ff56} use low/standard-resolution images with sparsely distributed objects. MIMO-Bench uses comparatively high-resolution real ($2500\times1500$) and synthetic ($1024\times1024$) images, because high resolution allows a single scene to hold many same-category instances as distractors. We additionally evaluate models on MIMO-Crop, obtained by cropping a window around each target from the original images (mechanism in the appendix), in two variants: $512\times512$ and $768\times768$. Because each crop is a window rather than a downsample, the target keeps full fidelity and is enlarged relative to the frame, while fewer same-category distractors fall inside it. MIMO-Crop thus approximates the low/standard-resolution regime of existing benchmarks and lets us check whether the MIMO difficulty persists once resolution, object scale, and distractor burden are partially controlled.

Table~\ref{tab:mimocrop} evaluates whether MIMO-Crop accounts for the gap between MIMO-Bench and existing low/standard-resolution benchmarks. Models improve on MIMO-Crop, confirming that full-image resolution, object scale, and distractor burden contribute to the difficulty. However, the recovery is incomplete: models improve from $39.5$ on full MIMO-Bench to $48.2$ on MIMO-Crop768 and $48.9$ on MIMO-Crop512 on average, but remain far below their average performance on standard low/standard-resolution benchmarks ($76.2$). This shows that MIMO-Bench difficulty is not explained by resolution and object scale alone, as models struggle even when the target region is enlarged and the number of same-category distractors is reduced.

\noindent\textbf{Synthetic gains are model-dependent}
The advantage is largest by far for the two G3.x models ($+27.5$ and $+25.3$ points
on synthetic AB), with a more moderate gain for Q2.5-72B-Instruct ($+12.0$).
Several other models, however, are flat or real-favored, including Q3-32B-Instruct ($-0.7$),
GLM-4.5V ($-8.6$), G2.5, and GPT-5. This spread suggests synthetic scenes are not
uniformly easier; rather, the synthetic-real gap reflects model-specific strengths rather
than a general property.

\noindent \textbf{Counting as a Function of Target-Set Size.}
We analyze numerical-counting performance as a function of the ground-truth (GT) target-set size. For each prediction $\hat{y}_i$ with GT count $y_i$, we compute signed error $e_i=\hat{y}_i-y_i$, bias $\frac{1}{n}\sum_i e_i$, MAE $\frac{1}{n}\sum_i |e_i|$, exact match $(\hat{y}_i=y_i)$, off-by-one accuracy $(|\hat{y}_i-y_i|\leq1)$, and off-by-two-or-more error $(|\hat{y}_i-y_i|\geq2)$. +ve bias indicates overcounting, while -ve bias indicates undercounting. Full count-conditioned statistics are reported in the appendix.

The count-conditioned profile aggregated across models reveals a structure that is hidden by aggregate counting metrics. Sparse target sets are prone to false positives: for GT count $0$, models achieve only $0.3333$ exact match and exhibit a positive bias of $+1.0833$, indicating that they often count visually similar distractors as valid targets. Small-to-moderate target sets remain the most reliable regime overall: mean $\mathcal{MRA}$ reaches $0.4968$ at count $3$ and $0.5234$ at count $5$, while exact match in these bins is still only $0.3541$ and $0.2888$, respectively, indicating that many predictions are near misses rather than exact counts. This behavior is analogous to the small-numerosity regime in cognitive science, where small sets can be rapidly individuated, while larger sets require slower serial enumeration or approximate numerosity estimation~\citep{revkin2008subitizing,piazza2011subitizing}. As the target set becomes dense, performance degrades sharply and the error direction flips. At GT count $4$, the average bias becomes negative ($-0.5406$), and undercounting grows with target-set size: the bias reaches $-2.2774$ at count $7$, $-3.1172$ at count $10$, $-4.5490$ at count $11$, and $-9.5556$ at count $14$. Exact match also collapses, dropping to $0.0292$ at count $7$ and $0.0000$ at count $14$. This mirrors cognitive evidence that spatially grouped, connected, or visually similar items can be underestimated because they are encoded as fewer perceptual units~\citep{chakravarthi2023fundamental,anobile2017connecting}. Taken together, these results suggest that counting errors are not uniform across target-set sizes: sparse scenes are vulnerable to false inclusion, whereas dense scenes are more often compressed into underestimates.

\noindent\textbf{In InternVL3, language-model scaling alone does not explain AB gains.}
InternVL3 improves from the 8B/14B variants to 38B (real
$34.4/35.9\to42.3$; synthetic $38.9/39.7\to48.3$), coinciding with an upgrade
from the $\sim$300M-parameter InternViT encoder to InternViT-6B at $\geq$38B
(alongside a larger language model). Beyond 38B the trend does not
continue: IVL3-78B regresses to $40.2$ (real) / $37.6$ (synthetic) despite
retaining the 6B encoder and using a far larger language model, suggesting that
language-model scaling alone does not explain the observed gains.

\noindent \textbf{Attribute-Level Failure Modes.}
Most object categories fail within a narrow band of $55$--$68\%$ (person $63.3\%$,
plant $64.1\%$, utensil $68.0\%$), with two-wheelers the exception at $36.8\%$.
Coarse attributes ask what an object is or whether a part is present (type,
presence), while fine-grained attributes ask for a finer property such as color,
pattern, sleeve length, shape, material, or texture. Fine-grained attributes fail
more often than coarse ones ($65.5\%$ vs.\ $55.7\%$). The gap is clearest when the
same object is asked at both levels: a pillow's type fails $26.2\%$ but its color
$78.0\%$ and pattern $75.3\%$; a bag's type fails $35.1\%$ but its pattern $79.2\%$.
Difficulty is uneven within the fine-grained group: pattern and color are
consistently the hardest (pattern fails $70.9\%$ across clothing, bags, and
pillows), whereas shape, material, and sleeve length are more moderate and at times
approach coarse-attribute rates (e.g., pillow shape $32.1\%$, plant-pot material $42.6\%$).

Note: Subsequent experiments in this and the next sections are done on an 800-sample MIMO-Bench Attribute Binding subset.

\begin{wraptable}{r}{0.48\columnwidth}
\vspace{-0.45cm}
\centering
\scriptsize
\caption{\textbf{Effect of thinking/reasoning.}
(``w/o'',``w/'') denotes (Instruct/Thinking) variant for Qwen and (Low/High) reasoning effort for Gemini.}
\vspace{-0.3cm}
\setlength{\tabcolsep}{2.2pt}
\renewcommand{\arraystretch}{1.02}
\begin{tabular}{@{}l l c c c c c@{}}
\toprule
\textbf{Model} & \textbf{Thk.}
& $\mathcal{I}_{\mathrm{ACC}}$ & $\mathcal{G}_{\mathrm{ACC}}$
& $\mathcal{I}_{\mathrm{MRA}}$ & $\mathcal{L}_{\mathrm{MRA}}$ & $\mathcal{G}_{\mathrm{MRA}}$ \\
\midrule
\multirow{2}{*}{Q3-8B}
& w/o & 39.2 & 40.1 & 43.9 & 40.0 & 37.3 \\
& w/  & 44.4 & 44.8 & 47.3 & 62.3 & 43.2 \\
\multirow{2}{*}{Q3-30B-A3B}
& w/o & 35.2 & 32.3 & 34.0 & 39.7 & 30.5 \\
& w/  & 37.0 & 37.3 & 35.4 & 40.1 & 31.3 \\
\multirow{2}{*}{Q3-32B}
& w/o & 46.3 & 46.0 & 46.3 & 46.1 & 50.4 \\
& w/  & 46.2 & 43.5 & 48.8 & 48.9 & 44.2 \\
\multirow{2}{*}{G3.1-Pro-Preview}
& w/o & 53.8 & 54.2 & 62.5 & 58.7 & 62.3 \\
& w/  & 53.2 & 54.3 & 57.7 & 52.8 & 63.5 \\
\multirow{2}{*}{G3.5-Flash}
& w/o & 53.3 & 56.5 & 60.3 & 58.3 & 58.3 \\
& w/  & 56.9 & 54.6 & 66.7 & 71.6 & 63.3 \\
\bottomrule
\end{tabular}
\label{tab:qwen_gemini_thinking}
\vspace{-0.3cm}
\end{wraptable}

\noindent\textbf{Thinking vs.\ Non-Thinking.}
Test-time reasoning does not uniformly improve MIMO performance
(Table~\ref{tab:qwen_gemini_thinking}, as tested on a subset). Its clearest benefit appears in
localized, decomposable counting. For Q3-8B, Thinking improves all
reported metrics, with $\mathcal{I}_{\mathrm{ACC}}$ increasing
$39.2\!\to\!44.4$ and $\mathcal{L}_{\mathrm{MRA}}$ increasing sharply
$40.0\!\to\!62.3$. Q3-30B-A3B also improves on both attribute-binding
metrics ($\mathcal{I}_{\mathrm{ACC}}$ $35.2\!\to\!37.0$,
$\mathcal{G}_{\mathrm{ACC}}$ $32.3\!\to\!37.3$). In contrast, Q3-32B shows a mixed effect:
Thinking slightly reduces AB ($\mathcal{I}_{\mathrm{ACC}}$
$46.3\!\to\!46.2$) and global counting ($\mathcal{G}_{\mathrm{MRA}}$
$50.4\!\to\!44.2$), while still improving localized counting
($\mathcal{L}_{\mathrm{MRA}}$ $46.1\!\to\!48.9$).
A similar lack of uniformity appears in proprietary models: higher reasoning
effort largely hurts G3.1-Pro-Preview on counting
($\mathcal{I}_{\mathrm{MRA}}$ $62.5\!\to\!57.7$, $\mathcal{L}_{\mathrm{MRA}}$
$58.7\!\to\!52.8$) but helps G3.5-Flash across
numerical-counting metrics ($\mathcal{I}_{\mathrm{MRA}}$ $60.3\!\to\!66.7$,
$\mathcal{G}_{\mathrm{MRA}}$ $58.3\!\to\!63.3$, $\mathcal{L}_{\mathrm{MRA}}$
$58.3\!\to\!71.6$). Overall, test-time reasoning does not
reliably improve performance on MIMO-Bench.

\section{Complexity Breakdown} \label{sec:analysis}

\begin{figure}[!ht]
\centering
\vspace{-0.7cm}
\begin{subfigure}[!t]{0.4\columnwidth}
  \centering
  \includegraphics[width=\linewidth]{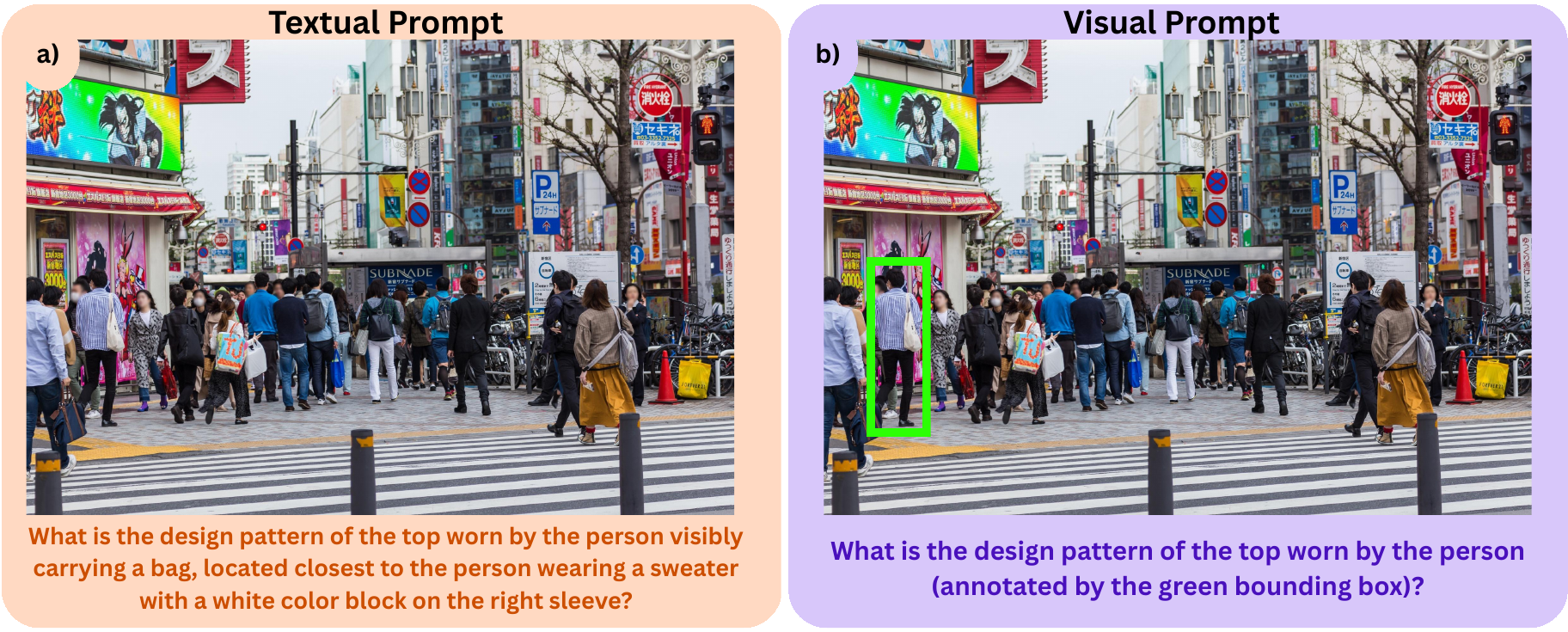}
  \label{fig:visual_prompting_a}
\end{subfigure}
\hfill
\begin{subfigure}[!t]{0.56\columnwidth}
  \centering
  \includegraphics[width=\linewidth]{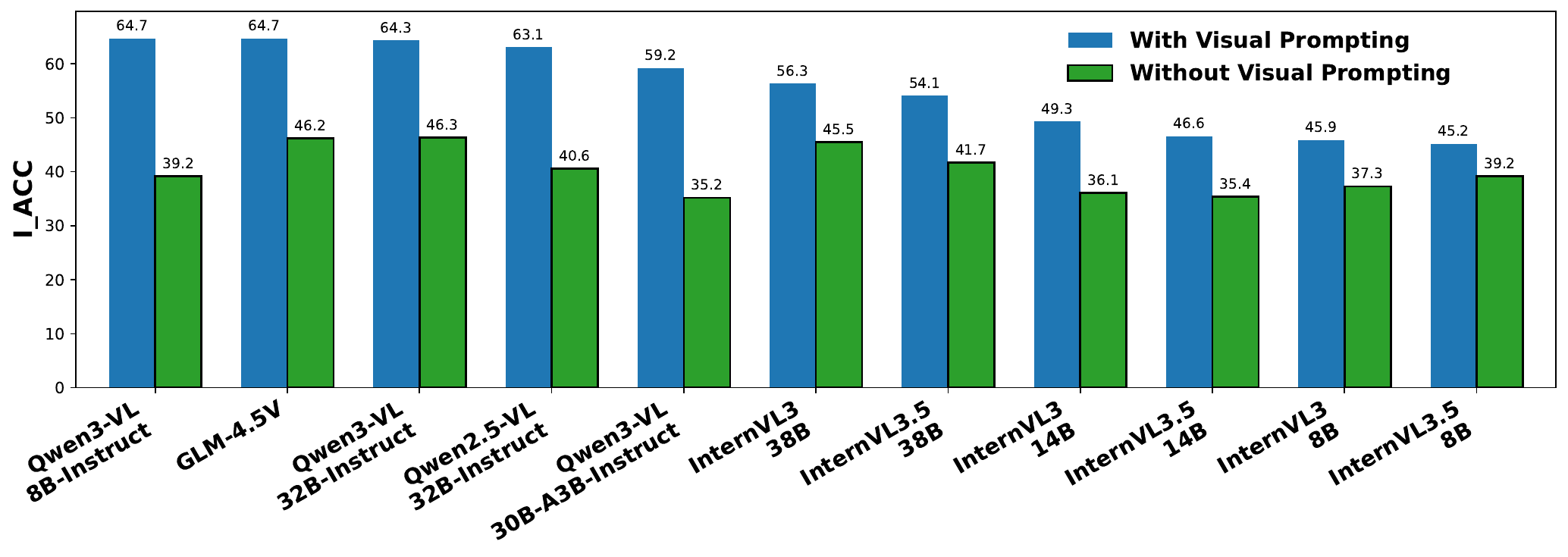}
  \label{fig:visual_prompting_b}
\end{subfigure}
\vspace{-0.6cm}
\caption{Visual prompting setup. Highlighting the target with a bounding box (a) instead of an RE consistently improves $\mathcal{I}_{\text{ACC}}$ (b).}
\label{fig:visual_prompting_combined}
\vspace{-0.3cm}
\end{figure}

Model's error on attribute binding arises from two major distinct stages: the difficulty of comprehending the referring expression (RE) in order to reach the target, and the difficulty of binding attributes once the target is found. Both are affected by the presence of similar-feature distractors. Hence, we ask:
\begin{itemize}[noitemsep, topsep=0pt]
    \item How do MLLMs perform when the complexity of comprehending the referring expression is removed?
    \item How do MLLMs perform when the presence of distractors is reduced?
\end{itemize}

\begin{wrapfigure}{r}{0.42\columnwidth}
\centering
\vspace{-0.5cm}
\includegraphics[width=\linewidth]{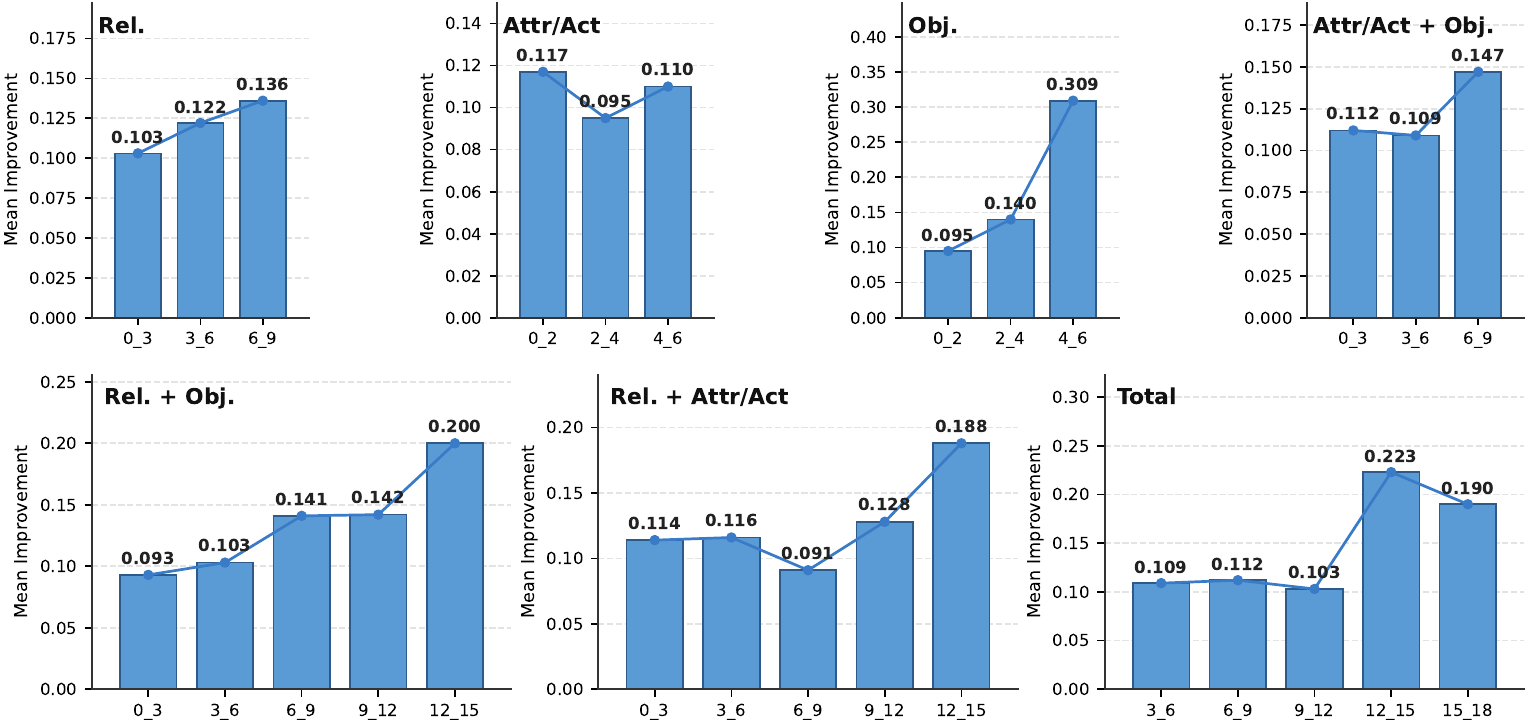}
\vspace{-0.7cm}
\caption{Effect of visual prompting on $\mathcal{I}_{\mathrm{ACC}}$.}
\label{fig:visual_prompting_graph}
\vspace{-0.5cm}
\end{wrapfigure}

\noindent\textbf{Visual Prompting.} We test how models perform when the referring expression no longer has to be comprehended. In the standard setting the model must parse the textual RE and ground it to the correct instance; in the visual-prompting setting we remove this step by replacing the RE with a bounding box/polygon drawn directly on the target. Fig.~\ref{fig:visual_prompting_combined} illustrates how an image--question pair transforms under visual prompting, and reports the effect: removing the RE improves $\mathcal{I}_{\mathrm{ACC}}$ for every evaluated model. The benefit also grows with RE complexity. As shown in Fig.~\ref{fig:visual_prompting_graph}, the gain from visual prompting increases as the number of compositional elements in the expression rises, indicating that the RE is a larger obstacle precisely when it is most compositional. Even with the target marked directly, \textbf{no model exceeds $\sim$65\%}{, though the marker alters the image, so the
residual error cannot be attributed to attribute binding alone; we attempt to disentangle grounding from attribute binding in Appendix~\ref{app:grounding}.}

\begin{tcolorbox}[
  colback=blue!5,
  colframe=blue!55,
  boxrule=0.4pt,
  arc=2pt,
  left=4pt,
  right=4pt,
  top=3pt,
  bottom=3pt,
]
Replacing the referring expression with a visual prompt raises $\mathcal{I}_{\mathrm{ACC}}$ by $+6.0$ to $+25.5$ points (mean $+15.5$) for every evaluated model, yet none exceeds $\sim$65\%.
\end{tcolorbox}

\vspace{-0.3cm}

\begin{figure}[!ht]
\centering
\begin{subfigure}[t]{0.38\columnwidth}
  \centering
  \includegraphics[width=\linewidth]{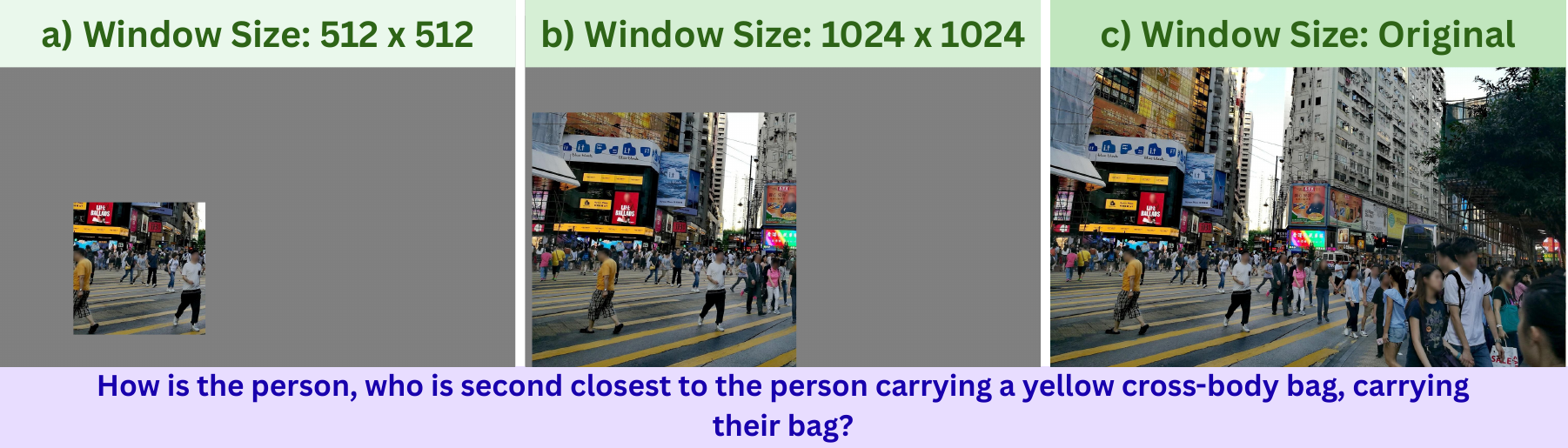}
  \label{fig:gray_a}
\end{subfigure}
\hfill
\begin{subfigure}[t]{0.60\columnwidth}
  \centering
  \includegraphics[width=\linewidth]{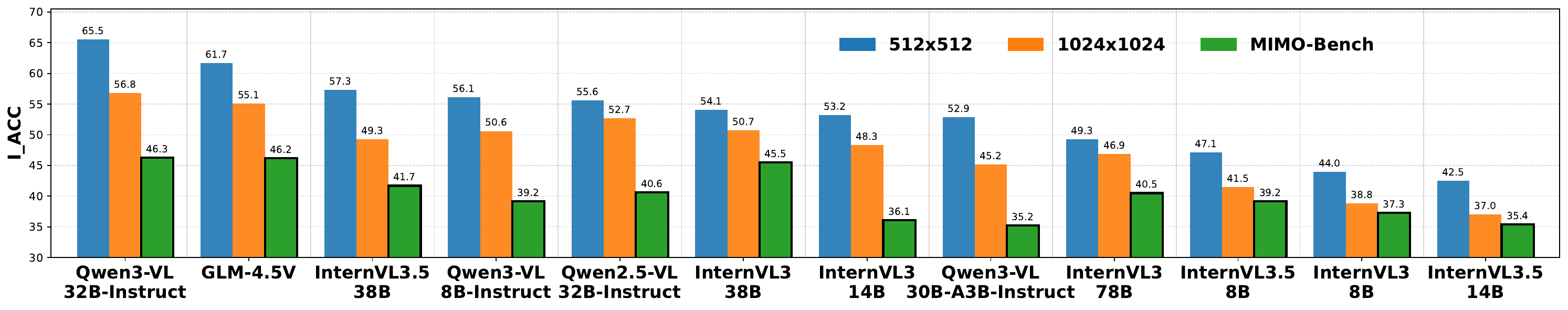}
  \label{fig:gray_b}
\end{subfigure}
\vspace{-0.5cm}
\caption{Visual Distractors Effect: (left) Graying setup \& (right) Performance trends as distractors increase.}
\label{fig:gray_combined}
\end{figure}

\noindent \textbf{Performance as a Function of Visual Distractor Reduction.}
We study how model performance changes under interventions that reduce visual distractor burden. We first analyze this through graying: we gray out the entire image except for a visible window centered on the queried object, using window sizes of 512 and 1024, where smaller windows leave fewer same-category distractors visually active (Fig.~\ref{fig:gray_combined}). Throughout all configurations, the target object's scale, image resolution, and textual prompt remain fixed. Fig.~\ref{fig:gray_combined} shows that $\mathcal{I}_{\mathrm{ACC}}$ generally rises as the window shrinks and fewer distractors remain in view, across the majority of evaluated models. The effect is directional but not strictly proportional: as detailed in the appendix, the accuracy gained per distractor removed varies substantially across images and even between window sizes of the same image, so \textbf{distractor count alone does not predict the size of the change}.

{Graying is a valid distractor-reduction intervention, but not a clean causal isolation of distractor count alone: it reduces visual competition from surrounding objects by suppressing the surrounding region, while also suppressing surrounding context and increasing the relative salience of the retained region. We therefore compare it against blur and inpainting controls in Appendix~\ref{app:distractor_controls}. Across these interventions, performance generally improves as distractor burden is reduced, showing that the directional effect is not specific to graying. Inpainting provides the closest practical approximation to directly removing selected same-category distractors while preserving the rest of the image, although segmentation and generation artifacts prevent treating it as an exact counterfactual. \textbf{The similar behavior observed under inpainting strengthens the interpretation that reducing visual distractor burden benefits MIMO performance, while we do not claim a clean causal effect of distractor count alone.}}

\noindent \textit{The two manipulations help different models.} From Figs.~\ref{fig:visual_prompting_combined} and~\ref{fig:gray_combined}, models such as Q3-8B-Instruct, Q2.5-32B-Instruct, Q3-30B-A3B-Instruct, and GLM-4.5V gain more from removing the referring expression, whereas others (e.g., IVL3.5-38B, IVL3-14B, and Q3-32B-Instruct) gain more from reducing distractors.

\section{What Enhances MLLM Performance in MIMO?}\label{sec:what_enhances}

\noindent \textbf{What helps in performing well in MIMO? A human analogy.}
In MIMO settings, objects rarely exhibit a clear visual ``pop-out'' effect,
requiring humans to actively search the scene to locate targets. This search is
typically guided by semantic and syntactic cues from referring expressions or
internal reasoning, combined with pre-attentive visual features
\citep{wolfe1989guided, TREISMAN198097}. When such guidance is weak, humans
often resort to a serial search strategy, scanning objects sequentially across
the scene. Effective human perception therefore involves examining the image \emph{dynamically} at multiple spatial scales while iteratively refining attention rather than fixed at
input, as in a static forward pass.

We hypothesize that this dynamic
visual access is what the MIMO setting rewards. Notably, several top-performing MLLMs on MIMO-Bench
already incorporate such mechanism: Qwen3-VL's
DeepStack \citep{bai2025qwen3vltechnicalreport} re-injects visual features across
multiple layers of the language model, enabling access to visual representations at different levels of abstraction, while GPT-style interleaved image--text
reasoning \citep{openai2024thinkingimages} repeatedly re-attends to relevant image
regions across intermediate reasoning steps. We
cannot directly verify that these mechanisms are responsible for the observed
performance. To measure the effect of dynamism, we perform a small exploratory study.

\begin{wraptable}{r}{0.22\columnwidth}
\vspace{-0.4cm}
\centering
\setlength{\tabcolsep}{2pt}
\renewcommand{\arraystretch}{1.05}
\caption{Effect of zooming on model performance. $\Delta$ denotes the
change in $\mathcal{I}_{\text{ACC}}$ relative to MIMO-Bench.}
\vspace{-0.3cm}
\label{tab:zooming_delta}
\resizebox{\linewidth}{!}{%
\begin{tabular}{@{}l c c@{}}
\toprule
\textbf{Model} & \textbf{Zoom.} & $\boldsymbol{\Delta}$ \\
\midrule
Q3-8B-Instruct       & Naive & $+0.89$ \\
Q2.5-7B-Instruct     & Naive & $+1.50$ \\
IVL3-8B              & Naive & $+5.46$ \\
IVL3-14B             & Naive & $+16.90$ \\
IVL3.5-8B            & Naive & $+2.14$ \\
IVL3.5-14B           & Naive & $+11.75$ \\
{Q3-32B-Instruct} & {Naive} & {$+3.00$} \\
\bottomrule
\end{tabular}}
\vspace{-0.5cm}
\end{wraptable}

We perform a multi-turn inference in which the model is posed with the
original question and must propose a region -- specified as bounding-box
coordinates -- that it deems sufficient to answer the question. This region is
cropped from the image, and the cropped image is re-queried together with the
original question to obtain the final answer. We refer to this as \emph{naive}
relevant-region extraction as the model is not explicitly trained to propose a relevant crop. As shown in
Table~\ref{tab:zooming_delta}, even this naive single-pass narrowing yields a
consistent improvement across all evaluated models, supporting the hypothesis
that dynamic narrowing of the search space benefits performance on MIMO-Bench, akin to the human mechanism.
{We further include fixed/random crop controls, a two-pass no-crop control, and IoU-conditioned accuracy to disentangle localization quality from cropping, rescaling, and the additional inference pass; see Appendix~\ref{app:region_controls}.}

\begin{tcolorbox}[
  colback=blue!5,
  colframe=blue!55,
  boxrule=0.4pt,
  arc=2pt,
  left=4pt,
  right=4pt,
  top=3pt,
  bottom=3pt,
]
Letting a model narrow the search space before answering consistently improves accuracy, indicating that MIMO rewards dynamic visual access rather than a single static forward pass.
\end{tcolorbox}

\begin{wrapfigure}{r}{0.3\columnwidth}
\vspace{-0.5cm}
\centering
\includegraphics[width=\linewidth]{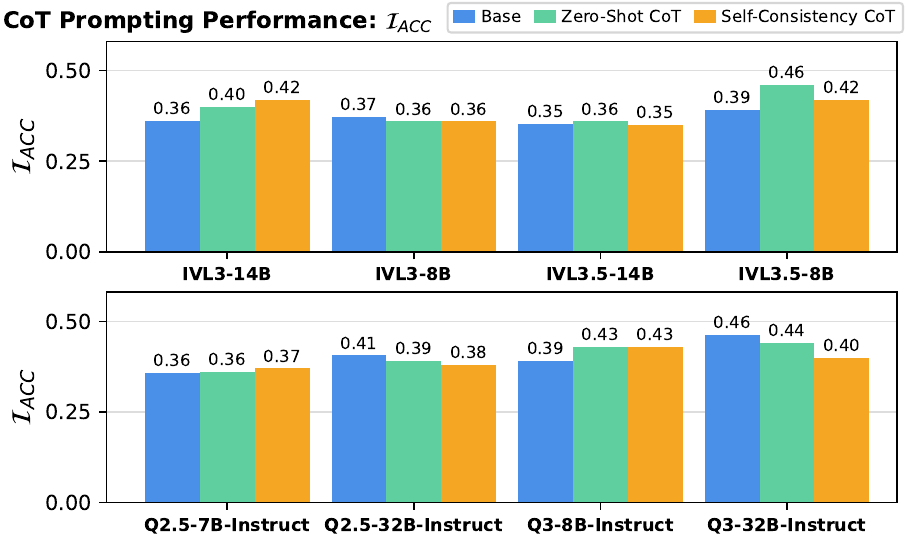}
\vspace{-0.6cm}
\caption{CoT on MIMO-Bench.}
\label{fig:cot}
\vspace{-0.3cm}
\end{wrapfigure}

\noindent\textbf{Does CoT Enhance MIMO Performance?} We evaluate whether
text-based Chain-of-Thought (CoT) prompting helps performance on
MIMO-Bench. Following prior prompting setups~\citep{Yang_2025_CVPR}, we
test Zero-Shot CoT (ZS-CoT) and Self-Consistency CoT (SC-CoT). For
SC-CoT, we run five CoT generations and use majority voting over the
predicted answers. As shown in Fig.~\ref{fig:cot}, CoT yields
model-dependent gains rather than consistently improving performance.
Several models benefit from at least one CoT variant: IVL3.5-8B improves
from $0.39$ to $0.46$ under ZS-CoT, IVL3-14B from $0.36$ to $0.42$ under
SC-CoT, and Q3-8B from $0.39$ to $0.43$ under both variants. Q2.5-7B
improves marginally from $0.36$ to $0.37$ under SC-CoT, while IVL3.5-14B
remains nearly unchanged. In contrast, several models degrade under CoT:
IVL3-8B slightly decreases, Q2.5-32B drops from $0.41$ to $0.39$ with
ZS-CoT and $0.38$ with SC-CoT, and Q3-32B drops from $0.46$ to $0.44$
with ZS-CoT and $0.40$ with SC-CoT. These results suggest that CoT can
provide modest gains for some models, but is not uniformly beneficial
and may degrade performance for others.

\section{Conclusion}\label{sec:conclusion}
We introduce MIMO-Bench, a benchmark for evaluating multimodal LLMs in settings with multiple instances across multiple object categories. MIMO-Bench is constructed around the complexity introduced by distractors, and our experiments show how strongly distractors hinder performance. We further show that removing the visual subtask posed by referring expressions yields a large performance gain, while our distractor-reduction interventions show that reducing visual distractor burden generally improves model performance. Based on these findings, we believe current MLLMs should first master the elementary tasks of attribute binding and numerical counting in MIMO before tackling harder problems. We also argue that, in settings such as MIMO, an ideal evaluation should couple reasoning with localization, requiring models to localize the referred object. We explored this through two interfaces: selecting the correct target among visually marked candidate boxes, and directly predicting its coordinates. {These interfaces impose different demands and yield substantially different localization behavior (Appendix~\ref{app:grounding}), preventing us from treating localization as a stable mandatory intermediate benchmark stage.} Such localization can be viewed as an additional feature alongside semantic features, making the binding task more complete. {We release the target masks and boxes to facilitate future evaluation along this direction.} As models become more robust, we hope MIMO-Bench serves as a realistic benchmark and further inspires research on improving reasoning in multimodal LLMs.

\bibliography{main}
\bibliographystyle{tmlr}

\appendix
\clearpage
\section*{Appendix}
\noindent In this appendix, we discuss

\newcommand{\appTallFigHeight}{0.82\textheight}
\newcommand{\appSubFigHeight}{0.365\textheight}
\lstset{%
  basicstyle=\scriptsize\ttfamily,
  breaklines=true,
  breakatwhitespace=false,
  columns=fullflexible,
  keepspaces=true
}

\begin{itemize}

\item Benchmark construction including image curation, the synthetic-image self-preference check, the SAM3 distractor-count protocol, QA generation, attribute dictionary construction, referring expression parsing, the human annotation protocol, and scalability (Appendix~\ref{app:benchmark_construction}).

\item Creation of MIMO-Crop (Appendix~\ref{app:mimo_crop}).

\item Additional experimental details and results, including evaluation setup, {strict Exact Group Accuracy,} the parent-level global--local counting comparison {and hierarchical consistency}, full count-conditioned counting statistics {and per-count support}, the visual prompting {and grounding} analyses, {the graying and relevant-region controls,} and {uncertainty and significance analysis} (Appendix~\ref{app:additional_results}).

\item Inference mechanisms for naive relevant-region extraction (Appendix~\ref{app:naive_zooming}) and for zero-shot and self-consistency CoT (Appendix~\ref{app:cot_mechanism}).

\item Limitation of the current work (Appendix~\ref{app:limitations}).

\end{itemize}

\section{Benchmark Construction}\label{app:benchmark_construction}

In this section, we describe the complete dataset and the benchmark generation process. Our benchmark comprises two tasks: \textit{attribute binding} and \textit{numerical counting}. Attribute binding tasks are defined with respect to a specific target instance. We consider VQA questions that consist of a primary task pertaining to a target object. Consequently, each target object is referenced by a referring expression that must uniquely localize the object in the image. For example, an attribute binding question may be of the form: ``What is the color of the \textit{car}?" where the italicized phrase serves as the referring expression.

For numerical counting, the question similarly relies on a referring expression to specify the set of instances to be counted. For example, ``Count the number of \textit{glass jars} in the image?" where the referring expression identifies the exact group of objects to be counted.

\subsection{Image Curation}

\begin{figure}[tbp]
\centering
\includegraphics[width=0.98\linewidth]{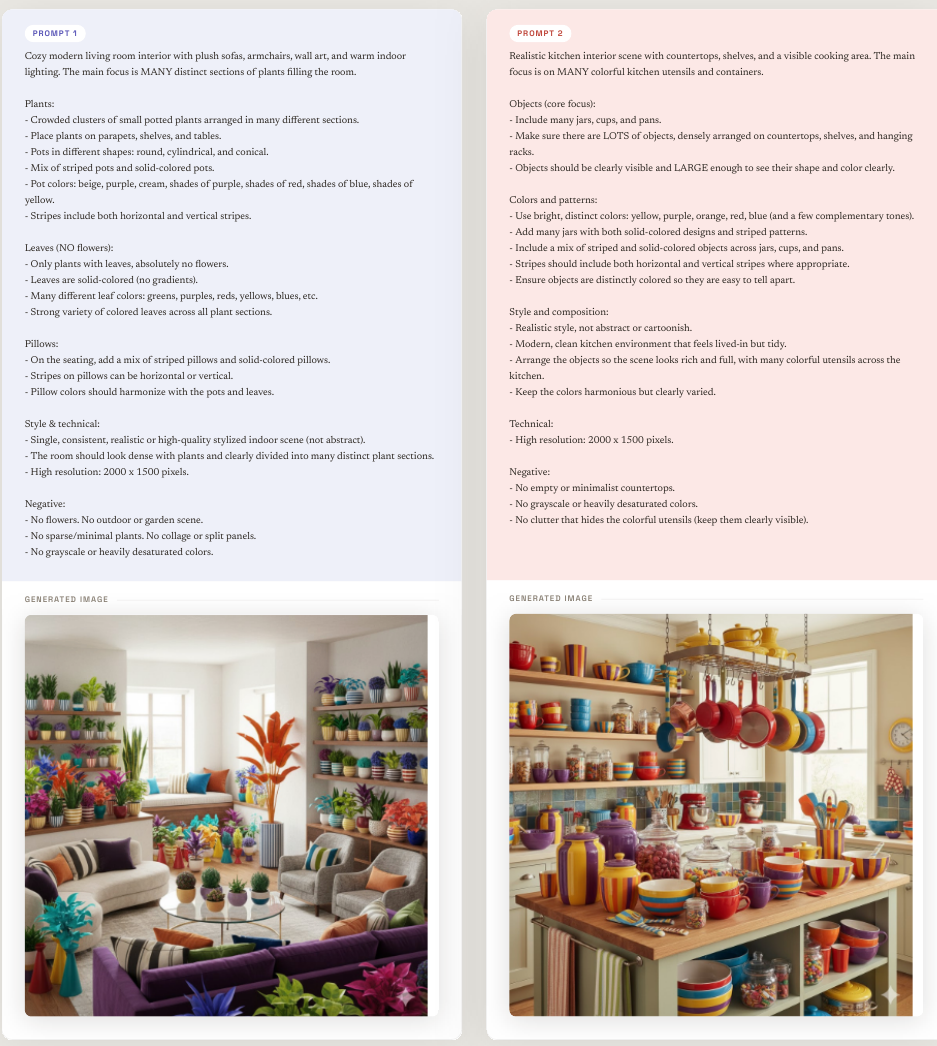}
\caption{Example prompts and corresponding generated synthetic images for MIMO-Bench. We consistently swap aesthetic details to gather varying images characterized by the MIMO setting.}
\label{fig:synthetic_prompt_images}
\end{figure}

For synthetic images, we use G2.5-Pro~\citep{comanici2025gemini} for image generation. Example prompts and their corresponding generated images are shown in Fig.~\ref{fig:synthetic_prompt_images}. Although the prompts requested a resolution of $2000\times1500$ pixels, G2.5-Pro returned images at its default resolution of $1024\times1024$ pixels. We reproduce the prompts verbatim, including the requested resolution, and retain all synthetic images at the returned $1024\times1024$ resolution. As each image is generated, we inspect the physical content of the image. We ensure that the image does not have inconsistencies such as color smudges (color from an object spreading into the background), physically disconnected objects (e.g., leaves from a plant are disconnected and seem to be hanging in the air), objects with impossible interactions (the leg of a table passing through the chair seat), etc.

We sample natural images from the SA-1B~\citep{kirillov2023segment} dataset using two filters. First, since MIMO-Bench requires object-dense scenes with many candidate distractors, we filter images using the number of available SA-1B mask/box annotations and discard images with insufficient annotated objects. Second, among the remaining object-dense images, we apply a depth-based filter to avoid close-up or visually flat scenes. Specifically, we estimate metric depth using the outdoor variant of Depth Anything V2~\citep{yang2024depth} and prioritize images with larger estimated scene depth. Images are retained only if they pass both the object-density filter and the depth filter. This depth score is used only as a sampling heuristic; final inclusion is determined by manual inspection, retaining images that contain multiple instances of at least one object category and support visually grounded, unambiguous referring expressions.

To quantify the same-category distractor density of the curated images, we compute per-category instance counts using SAM3~\citep{carion2025sam3segmentconcepts}, a promptable concept segmentation model. For each image and each annotated object category, we supply the category name (e.g., ``person'', ``plant'', ``vase'') as a text concept prompt; SAM3 returns one instance-level segmentation mask per detected instance, and the distractor count is the total number of returned masks. These counts underlie the per-category instance-count distributions reported in Fig.~\ref{fig:dataset_statistics} of the main paper.

\noindent\textbf{Data availability.} We will release all MIMO-Bench annotations and evaluation code. For the real-image split, we will release the corresponding SA-1B image identifiers and reconstruction instructions; users must obtain the underlying images directly from Meta under the SA-1B Dataset Research License, and we will not redistribute those images. For the synthetic split, we will release the Gemini-2.5-Pro-generated images together with their annotations under the terms specified in the accompanying repository.

\subsection{Synthetic Image Self-Preference Check}\label{app:self_preference}

\begin{wraptable}{r}{0.4\columnwidth}
\vspace{-0.35cm}
\centering
\small
\setlength{\tabcolsep}{5.5pt}
\renewcommand{\arraystretch}{1.05}
\begin{tabular}{@{}l c@{}}
\hline
\textbf{Model} & \textbf{Syn. AB $\mathcal{I}_{\mathrm{ACC}}$} \\
\hline
G2.5-Pro & 47.7 \\
GPT-5-medium & 49.7 \\
IVL3.5-241B-A28B & 45.1 \\
Q3-32B-Instruct & 47.9 \\
\hline
\end{tabular}
\vspace{-0.3cm}
\caption{\textbf{Self-preference check on Gemini-generated images.}
Attribute-binding instance accuracy on the synthetic split. G2.5-Pro
does not exhibit a clear self-preference advantage despite being used to
generate the synthetic images.}
\label{tab:synthetic_self_preference}
\vspace{-0.35cm}
\end{wraptable}

Since the synthetic images in MIMO-Bench are generated using G2.5-Pro~\citep{comanici2025gemini}, we additionally check whether G2.5-Pro obtains an artificial advantage on this split. We compare G2.5-Pro with proprietary and open-source models on a small synthetic attribute-binding pilot using the same prompt, answer-format constraints, and decoding setup as the main benchmark evaluation. As shown in Table~\ref{tab:synthetic_self_preference}, G2.5-Pro does not show a clear self-preference advantage on Gemini-generated images: its performance is comparable to strong non-Gemini models rather than being uniquely dominant. This suggests that the synthetic split does not merely favor the image generator, and that performance on this split still reflects the ability to ground compositional referring expressions and bind attributes under dense same-category distractors. {\label{app:synthetic_robustness}This control covers only G2.5-Pro: neither G3.5-Flash nor G3.1-Pro-Preview generated the synthetic split, so applying the same check to them would test family-level affinity rather than generator self-preference. We therefore probed the G3.x synthetic result directly for evaluation artifacts, re-running the synthetic attribute-binding split under alternative option sets, shuffled answer options and deliberately biased option configurations; synthetic performance remained within $3$ points in every condition. These checks rule out the tested option-ordering and answer-set artifacts, but do not rule out evaluation artifacts in general and do not explain the gap, which we therefore report as an observed anomaly rather than genuine capability or a generator effect.}

\begin{figure}[tbp]
\centering
\includegraphics[width=\linewidth]{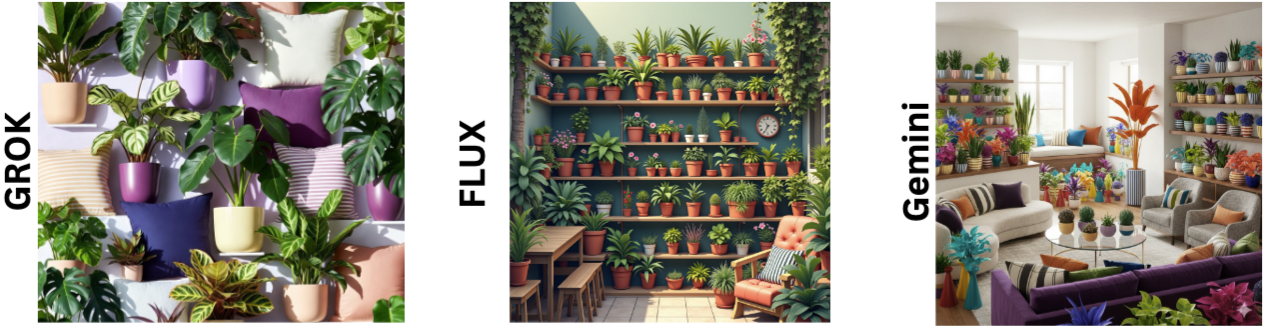}
\caption{\textbf{Qualitative comparison of synthetic images generated by Grok-4, Flux, and G2.5-Pro} under the same MIMO-style prompts. G2.5-Pro produces more realistic, feature-diverse scenes with stronger prompt adherence. Grok-4 and Flux exhibit reduced realism, weaker feature diversity, and object-level artifacts, making them less suitable for constructing unambiguous MIMO-style QA pairs.}
\label{fig:syn_image_comparison}
\end{figure}

We also qualitatively compared G2.5-Pro with Grok-4 and Flux for synthetic image generation. Although all three models could generate dense multi-instance scenes, G2.5-Pro was selected because it more reliably produced visually coherent scenes with stronger prompt adherence. In comparison, Grok-4 and Flux more frequently produced images with reduced realism, weaker feature diversity, or object-level artifacts that made them less suitable for constructing unambiguous MIMO-style QA pairs. Fig.~\ref{fig:syn_image_comparison} provides a qualitative side-by-side illustration.

\subsection{QA Generation}

Each object in an image is annotated by referring expression(s) and bounding polygons. Alongside this, images for numerical counting are annotated on a group-wise basis, as detailed in Sec.~\ref{sec:eval_metrics} and illustrated in Fig.~\ref{fig:attribute_counting} of the main paper.

For attribute binding, each object in an image is mapped with its attribute dictionary to generate an exhaustive set of raw \textit{multiple-choice} image-question pairs with single-choice answers. For numerical counting, the annotations are prefixed by the phrase ``Count the number of \dots", and the expected answer is a number. These raw image-question pairs are then answered and verified by humans.

\noindent \textbf{Considerations:}

Few attributes, such as color, are continuous. We construct multiple-choice options using carefully selected negative categories. For example, when the ground-truth color is ``blue'', we exclude candidate colors that are perceptually too similar to blue to avoid overlap or ambiguity (see id 4 in Fig.~\ref{fig:question_dict_sample}). Instead, we select visually distinct and commonly used color terms as answer options.

\subsection{Attribute Dictionaries}\label{attribute_dictionary}

\begin{figure}[h]
    \centering
    \includegraphics[width=\linewidth,height=15cm,keepaspectratio]{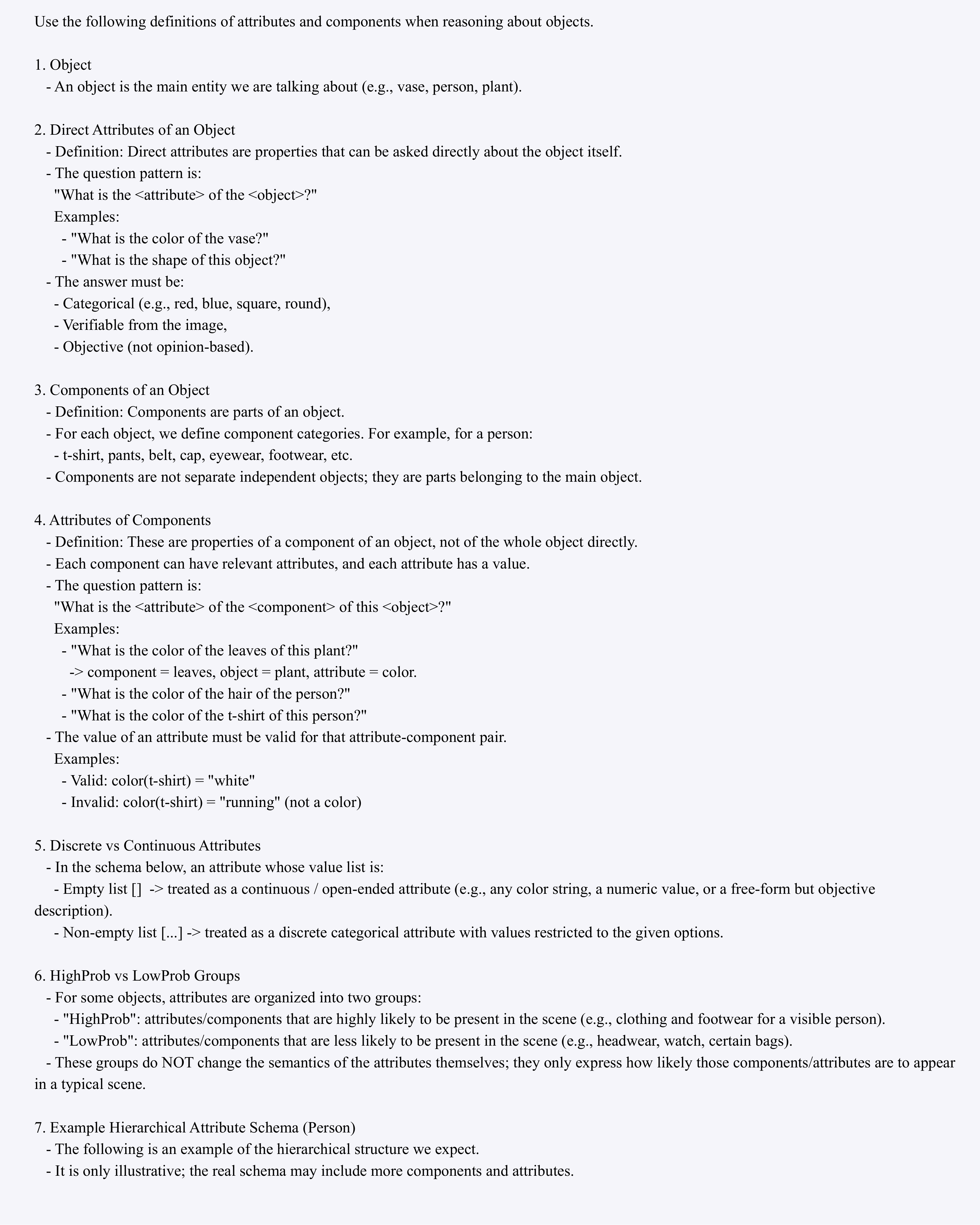}
    \caption{Pre-prompt used for generating attribute-binding dictionaries for different object categories (1/2).}
    \label{fig:dict_prompt1}
\end{figure}

\begin{figure}[p]
    \centering
    \includegraphics[width=\linewidth,height=\appTallFigHeight,keepaspectratio]{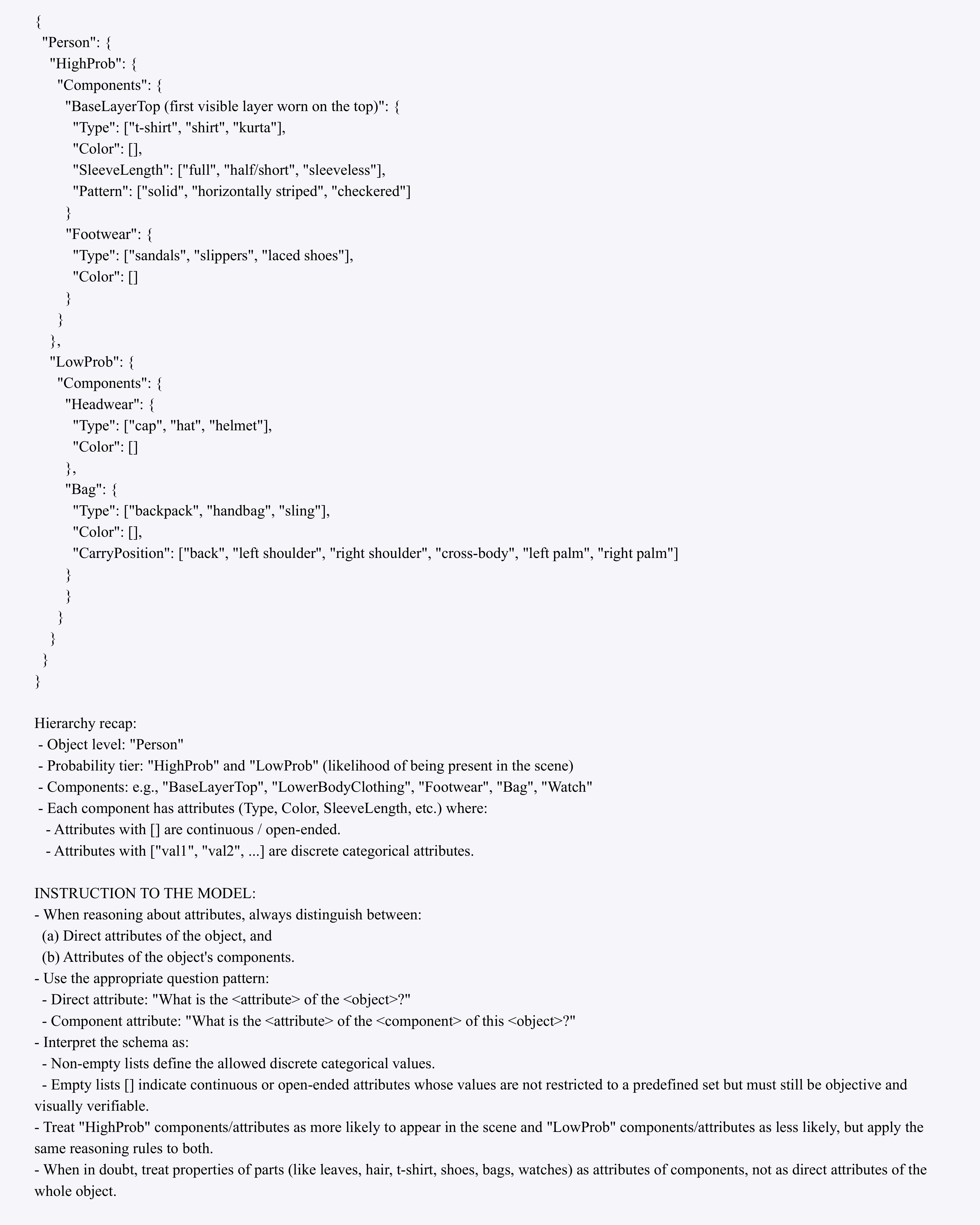}
    \caption{Pre-prompt used for generating attribute-binding dictionaries for different object categories (2/2).}
    \label{fig:dict_prompt2}
\end{figure}

Attribute Dictionaries are hierarchical dictionaries that contain the possible attributes of an object category or its components could possess. The discrete attributes usually have a set of plausible answers which are exclusive and exhaustive, whereas continuous attribute values have a soft boundary and are left to the human answering annotator to decide what the value should be. These dictionaries are mapped with target objects from an image and passed through a template-based question generation script that generates a raw list of multiple-choice questions. We join the prompts from Figs.~\ref{fig:dict_prompt1},~\ref{fig:dict_prompt2}(represented in two figures due to spacing limitations) in order to generate the dictionaries for different object categories Figs.~\ref{fig:person_json},~\ref{fig:plant_json},~\ref{fig:lamp_json}, and~\ref{fig:utensil_json} present attribute dictionaries corresponding to ``person", ``plant", ``lamp", and ``utensil", respectively.


\begin{figure}[p]
\centering
\begin{tcolorbox}[colback=blue!10, colframe=black, title=Human]
{\tiny
\begin{lstlisting}
{
  "Person": {
    "HighProb": {
      "Components": {
        "BaseLayerTop (first visible layer worn on the top)": {
          "Type": ["t-shirt", "shirt", "polo", "kurta", "sweatshirt", "vest", "dress/frock", "top"],
          "Color": [],
          "SleeveLength": ["full", "half/short", "sleeveless", "3/4"],
          "Pattern": ["solid", "horizontally striped", "checkered", "printed", "vertically striped"]
        },
        "UpperBodyClothing (visible layers over the first/base visible layer of top)": {
          "Type": ["jacket", "sweater", "hoodie", "blazer", "cardigan", "overshirt"],
          "Color": [],
          "SleeveLength": ["full", "half/short", "sleeveless", "3/4"],
          "Pattern": ["solid", "horizontally striped", "checkered", "printed", "vertically striped"]
        },
        "LowerBodyClothing": {
          "Type": ["full length jeans", "trousers", "cotton shorts/capris", "skirt", "leggings", "denim shorts/capris", "chinos", "linen pants", "sweatpants"],
          "Color": []
        },
        "Footwear": {
          "Type": ["sandals", "slippers", "barefoot", "boots(ankle or calf)", "knee boots", "laced shoes(includes sports shoes, sneakers)", "loafers(includes moccasins, slip-ons, loafers)", "heels"],
          "Color": []
        }
      }
    },
    "LowProb": {
      "Components": {
        "Headwear": {
          "Type": ["cap", "hat", "beanie", "turban", "helmet"],
          "Color": []
        },
        "Bag": {
          "Type": ["backpack", "handbag", "tote", "sling", "briefcase", "purse", "duffel bag"],
          "Color": [],
          "CarryPosition": ["back", "front", "left shoulder", "right shoulder", "left forearm", "right forearm", "cross-body", "left palm", "right palm", "around the waist"]
        },
        "Watch": {
          "Wrist": ["left", "right", "unknown"],
          "Color": []
        }
      }
    }
  }
}
\end{lstlisting}
}
\end{tcolorbox}
\caption{Person attribute dictionary in JSON format.}
\label{fig:person_json}
\end{figure}

\begin{figure}[p]
\centering
\begin{tcolorbox}[colback=blue!10, colframe=black, title=Plant]
{\tiny
\begin{lstlisting}
{
  "Plant": {
    "HighProb": {
      "DirectAttributes": {
        "Type": ["tree", "shrub", "herb", "creeper", "climber"]
      },
      "Components": {
        "Leaves": {
          "Color": [],
          "Shape": ["oval", "lanceolate", "heart-shaped", "needle-like", "lobed", "round"]
        },
        "StemOrTrunk": {
          "Color": ["green", "brown", "grey"],
          "Branching": ["unbranched", "branched"]
        }
      }
    },
    "LowProb": {
      "DirectAttributes": {
        "FlowerPresence": ["yes", "no"],
        "FruitPresence": ["yes", "no"]
      },
      "Components": {
        "Flowers": {
          "Color": []
        },
        "FruitsOrPods": {
          "Color": [],
          "Presence": ["present", "absent"]
        },
        "Roots": {
          "Visibility": ["exposed", "underground"],
          "Color": []
        },
        "PotOrContainer": {
          "Material": ["clay", "ceramic", "plastic", "metal", "cement"],
          "PrimaryColor": [],
          "Shape": ["round", "cylindrical", "rectangular", "conical/bowl"],
          "Texture": ["rough", "matte", "glossy", "ridged/groovy"],
          "Pattern": ["plain", "horizontally striped", "vertically striped", "dotted", "floral", "geometric", "checked/chequered", "gradient"],
          "StripedColors": {
            "Stripe1Color": [],
            "Stripe2Color": []
          }
        }
      }
    }
  }
}
\end{lstlisting}
}
\end{tcolorbox}
\caption{Plant attribute dictionary in JSON format.}
\label{fig:plant_json}
\end{figure}

\begin{figure}[p]
\centering
\begin{tcolorbox}[colback=blue!10, colframe=black, title=Lamp]
{\tiny
\begin{lstlisting}
{
  "Lamp": {
    "HighProb": {
      "DirectAttributes": {
        "Type": ["table lamp", "floor lamp", "wall-mounted lamp", "desk lamp", "hanging lamp"]
      },
      "Components": {
        "Lampshade": {
          "Shape": ["conical", "cylindrical", "drum", "bell", "square"],
          "Color": [],
          "Transparency": ["transparent", "translucent", "opaque"],
          "Pattern": ["plain", "striped", "floral", "painted(other than floral)"]
        },
        "Base": {
          "Shape": ["round", "rectangular", "square", "tripod", "ornate"],
          "Material": ["metal", "wood", "ceramic", "glass"],
          "Color": []
        },
        "Leg": {
          "Color": []
        }
      }
    },
    "LowProb": {
      "DirectAttributes": {
      },
      "Components": {
      }
    }
  }
}
\end{lstlisting}
}
\end{tcolorbox}
\vspace{-5pt}
\captionof{figure}{Lamp attribute dictionary in JSON format.}
\label{fig:lamp_json}

\begin{tcolorbox}[colback=blue!10, colframe=black, title=Utensil]
{\tiny
\begin{lstlisting}
{
  "Utensil": {
    "HighProb": {
      "DirectAttributes": {
        "Type": ["spoon", "fork", "knife", "ladle", "spatula", "tongs", "bowl", "plate", "cup", "pan", "pot", "tray", "jar", "jug"]
      },
      "Components": {
        "OuterSurface": {
          "PrimaryColor": [],
          "Transparency": ["transparent", "translucent", "opaque"],
          "Pattern": ["plain", "horizontally striped", "vertically striped", "dotted", "floral", "geometric", "checked/chequered", "gradient"],
          "StripedColors": {"Stripe1Color": [], "Stripe2Color": []}
        },
        "InnerSurface": {
          "Color": []
        }
      }
    },
    "LowProb": {
      "DirectAttributes": {
      },
      "Components": {
        "Lid": {
          "Presence": ["yes", "no"],
          "Material": ["steel", "glass", "plastic", "ceramic"],
          "Color": [],
          "Shape": ["flat", "domed"]
        }
      }
    }
  }
}
\end{lstlisting}
}
\end{tcolorbox}
\vspace{-5pt}
\captionof{figure}{Utensil attribute dictionary in JSON format.}
\label{fig:utensil_json}
\end{figure}

\subsection{Referring Expression Parsing}\label{app:referring_expression_parsing}

\begin{figure}[h]
\centering
\includegraphics[width=\linewidth,height=18cm, keepaspectratio]{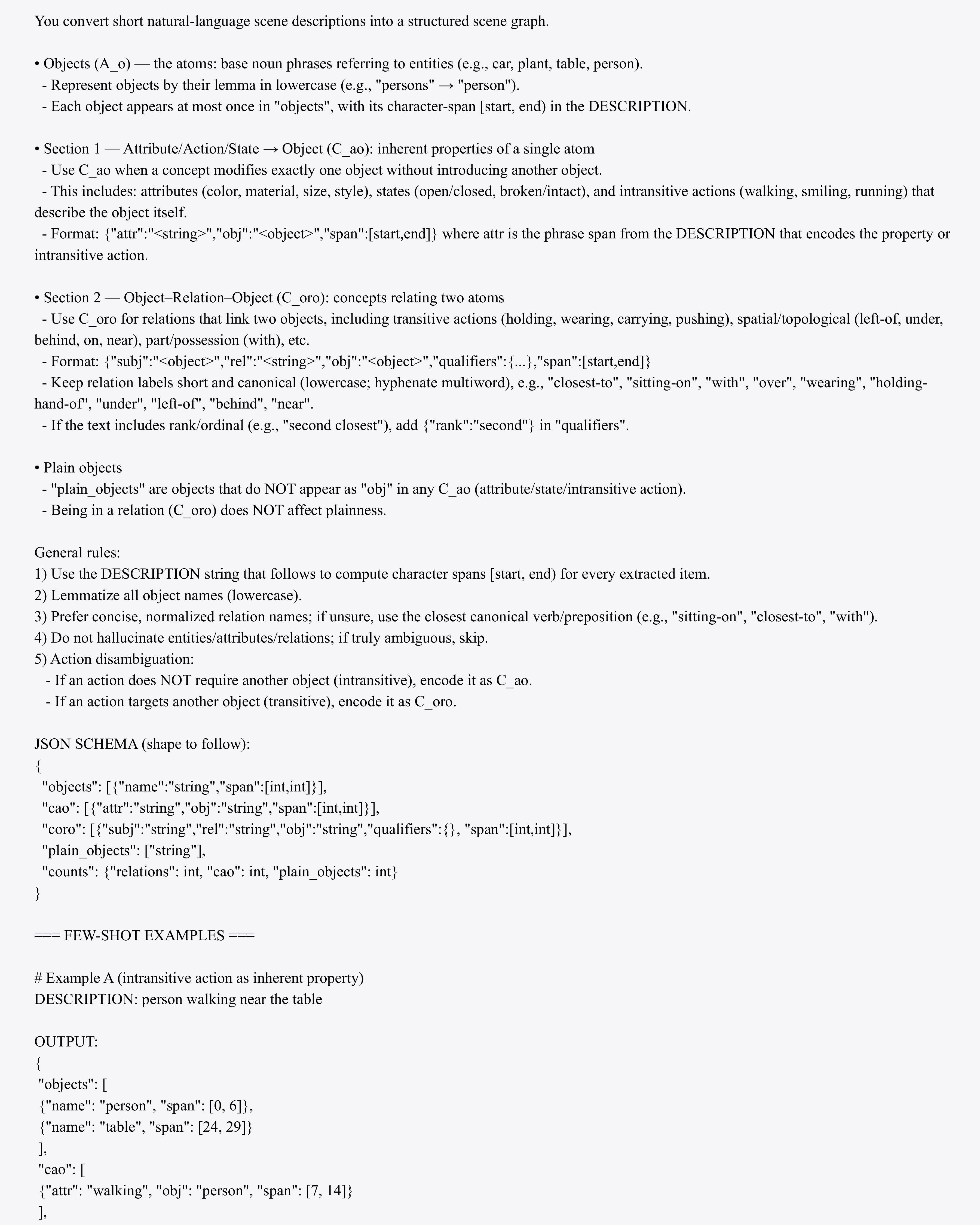}
\caption{Pre-prompt for parsing referring expressions into attributes/actions, relations, and plain objects (1/3).}
\label{fig:exp1}
\end{figure}

\begin{figure}[p]
    \centering
    \begin{subfigure}{0.48\linewidth}
        \centering
        \includegraphics[width=\linewidth,height=\appTallFigHeight,keepaspectratio]{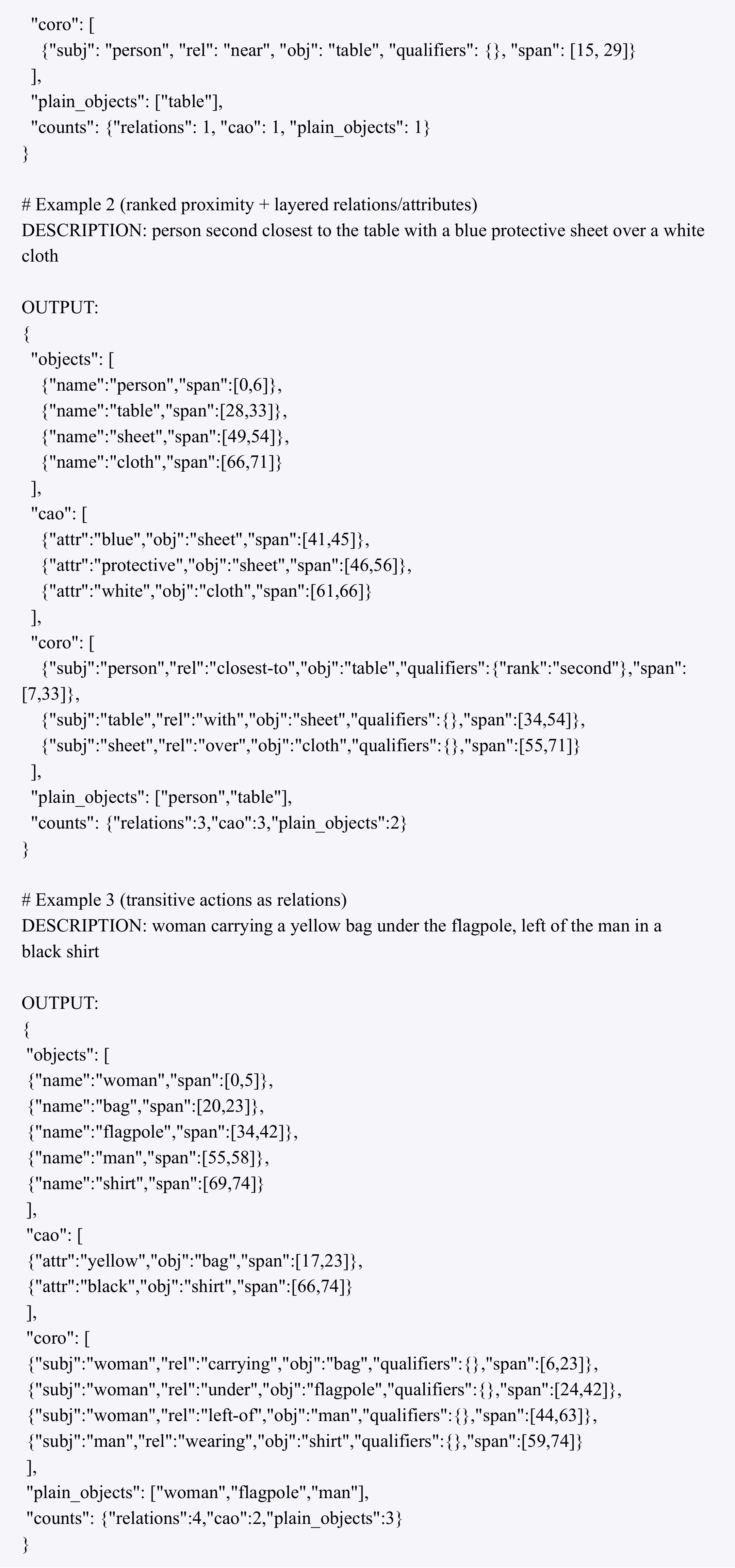}
        \caption{Pre-prompt for parsing referring expressions into attributes/actions, relations, and plain objects (2/3).}
        \label{fig:exp2}
    \end{subfigure}
    \hfill
    \begin{subfigure}{0.48\linewidth}
        \centering
        \includegraphics[width=\linewidth,height=\appTallFigHeight,keepaspectratio]{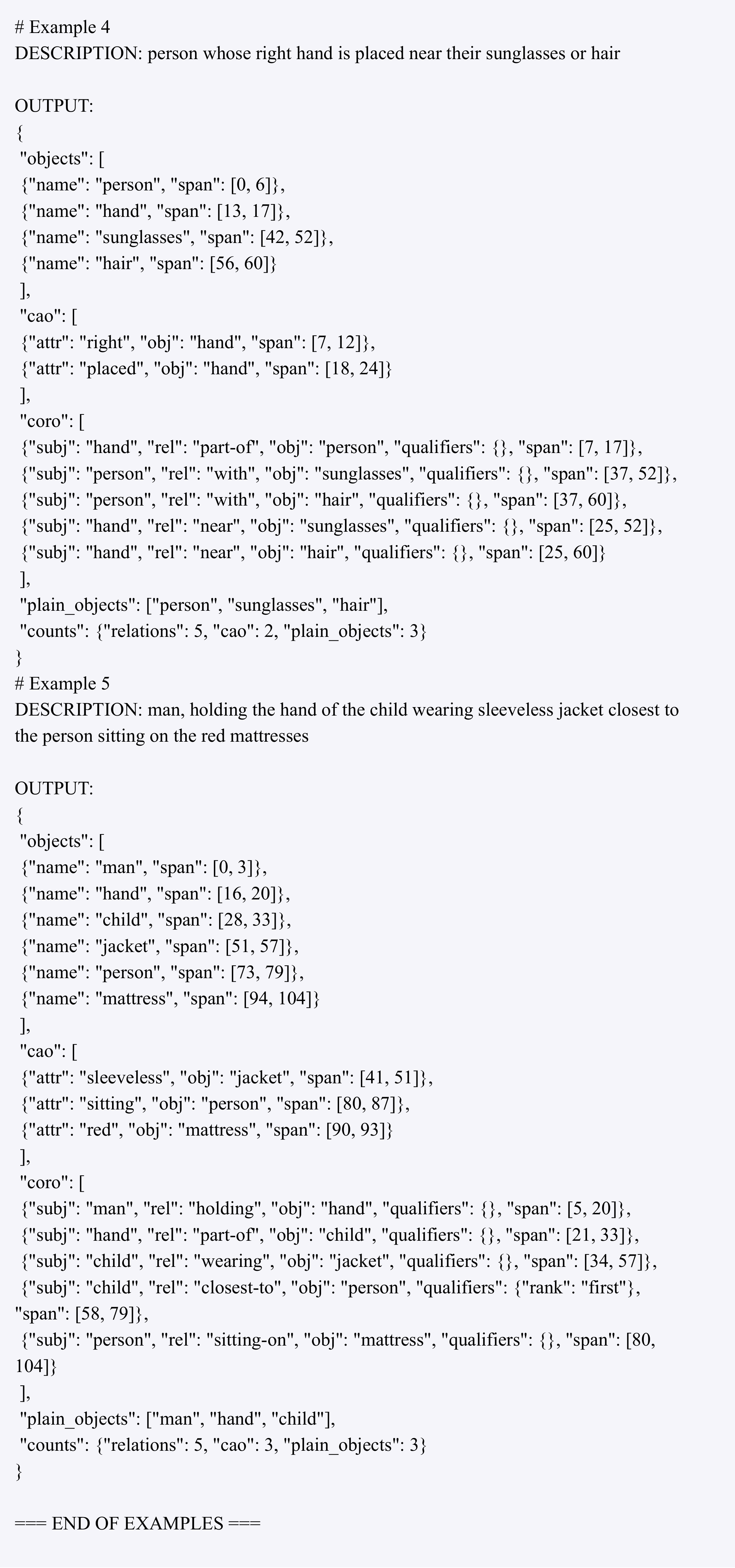}
        \caption{Pre-prompt for parsing referring expressions into attributes/actions, relations, and plain objects (3/3).}
        \label{fig:exp3}
    \end{subfigure}
\end{figure}

As discussed in the main paper, each referring expression is made up of attributes/actions, relations, and neighboring objects. To parse it, we use G2.5-Pro~\citep{comanici2025gemini}. We use the prompts as shown in Figs.~\ref{fig:exp1},~\ref{fig:exp2}, and~\ref{fig:exp3} and append them with the expression we want to parse. Each output from the model is stored as shown in the structure in the prompting example. Parsed expressions help us to disentangle the causes of certain results.

The parser converts a referring expression into a structured scene graph consisting of: (i) \textbf{objects}, base noun phrases referenced in the expression (lemmatized, lowercase); (ii) \textbf{Attr} (the \texttt{cao} field in the parser output), attribute/action/state bindings that modify a single object (e.g., color, material, or an intransitive action such as ``walking''); (iii) \textbf{coro}, object--relation--object triples linking two objects (transitive actions, spatial/topological relations, possession); and (iv) \textbf{plain\_objects}, objects that do not appear as the target of any Attr entry. Each parse also reports a \texttt{counts} field with the number of relations, Attr bindings, and plain objects.

We define the number of visual subtasks posed by a referring expression as the sum
\begin{equation}
\label{eq:subtask_count}
\#\text{subtasks} = \#\text{relations} + \#\text{Attr} + \#\text{plain\_objects},
\end{equation}
i.e., the count of \texttt{coro} triples plus the count of Attr bindings plus the count of plain objects returned by the parser for that expression. The mean subtask count reported for a benchmark (e.g., $9.6$ for MIMO-Bench in Table~\ref{tab:benchmark_positioning}) is the average of Eq.~\ref{eq:subtask_count} over all parsed referring expressions in that benchmark. The parser and this counting rule are benchmark-agnostic: the same prompt and formula, applied to plain natural-language referring expressions with no benchmark-specific structure required, were used to obtain the subtask counts for MIMO-Bench and for every other benchmark listed in Table~\ref{tab:benchmark_positioning} (CLEVR, GQA, MMStar, NaturalBench, VQAv2, V\textsuperscript{*}Bench, HR-Bench, and MME-RealWorld).

\begingroup
Table~\ref{tab:re_composition} reports these components across benchmarks. Object
count alone does not distinguish MIMO-Bench, and CLEVR contains more Attr
bindings; the clearest separation is in relations, where MIMO-Bench averages $3.23$
against at most $1.34$ elsewhere, roughly $2.4\times$ the next highest. The elevated
subtask count therefore comes mainly from relational structure rather than expression
length.

\begin{table}[!htbp]

\centering
\caption{Referring-expression composition across benchmarks, parsed with the same
benchmark-agnostic parser. Objects, Attr bindings and relations are the three
terms of Eq.~\ref{eq:subtask_count}; same-category distractor density is reported in
Table~\ref{tab:benchmark_positioning}.}
\label{tab:re_composition}
\small
\setlength{\tabcolsep}{6pt}
\renewcommand{\arraystretch}{1.05}
\begin{tabular}{@{}lrrrr@{}}
\toprule
\textbf{Benchmark} & \textbf{Objects} & \textbf{Attr} & \textbf{Relations} & \textbf{\# Subtasks} \\
\midrule
CLEVR          & 2.50 & \textbf{3.50} & 1.20 & 7.20 \\
GQA            & 2.24 & 0.42 & 0.43 & 3.10 \\
MMStar         & 3.36 & 0.08 & 0.46 & 3.90 \\
NaturalBench   & 1.87 & 0.53 & 0.30 & 2.70 \\
VQAv2          & 1.76 & 0.35 & 0.49 & 2.60 \\
V\textsuperscript{*}Bench & 1.92 & 0.07 & 0.40 & 2.40 \\
HR-Bench       & 2.92 & 0.44 & 1.34 & 4.70 \\
MME-RealWorld  & 2.70 & 0.25 & 0.95 & 3.90 \\
\midrule
\textbf{MIMO-Bench} & \textbf{4.01} & 2.33 & \textbf{3.23} & \textbf{9.56} \\
\bottomrule
\end{tabular}%
\end{table}
\endgroup

\clearpage
\subsection{Human Answering and Verification Protocol}

\textbf{Human Answering}

\begin{figure}[!t]
\centering
\captionsetup[subfigure]{font=scriptsize,justification=centering,skip=1pt}
\begin{subfigure}[t]{0.31\linewidth}
    \centering
    \includegraphics[width=\linewidth]{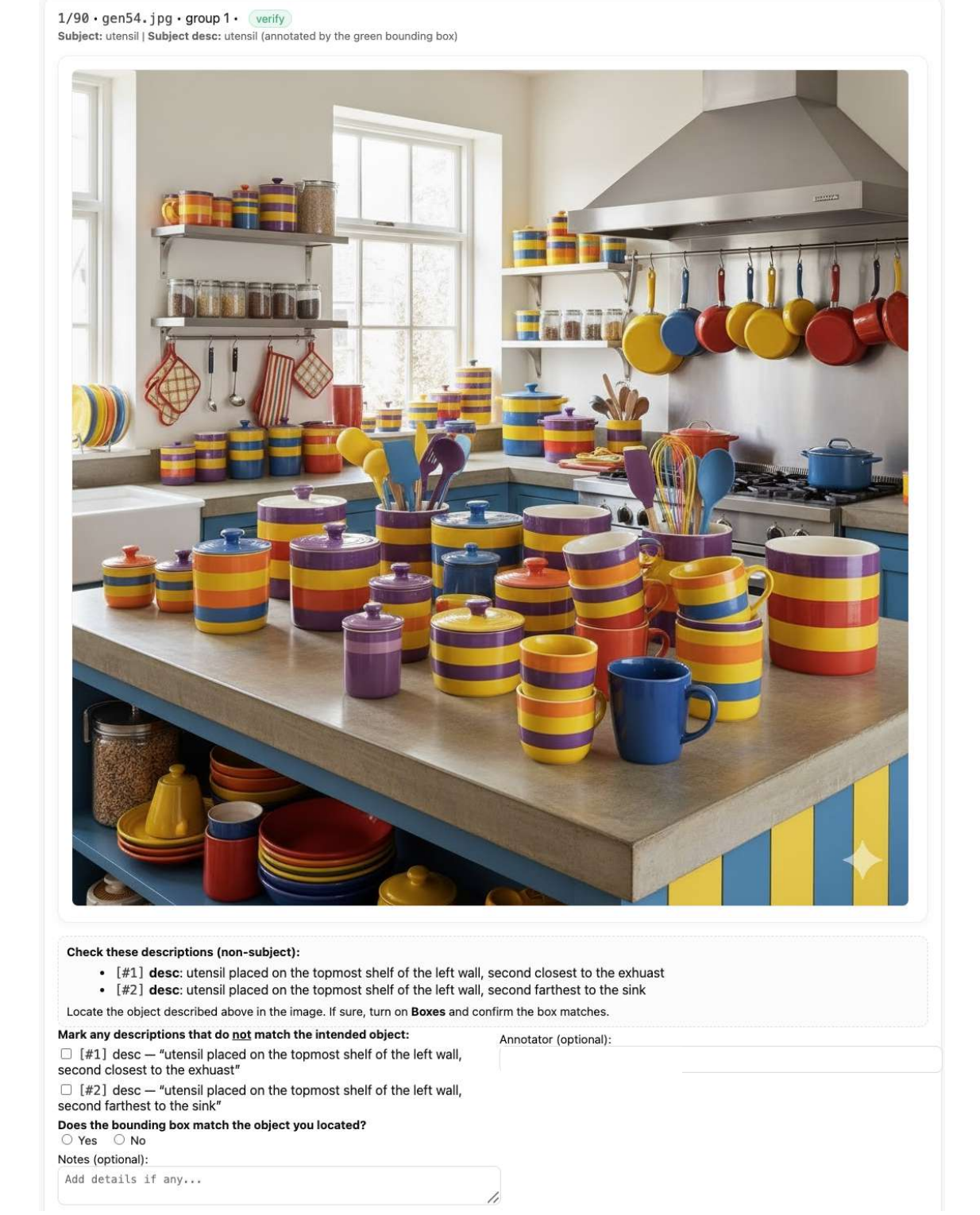}
    \caption{Description without bounding box.}
    \label{fig:desc_no_box}
\end{subfigure}\hfill
\begin{subfigure}[t]{0.31\linewidth}
    \centering
    \includegraphics[width=\linewidth]{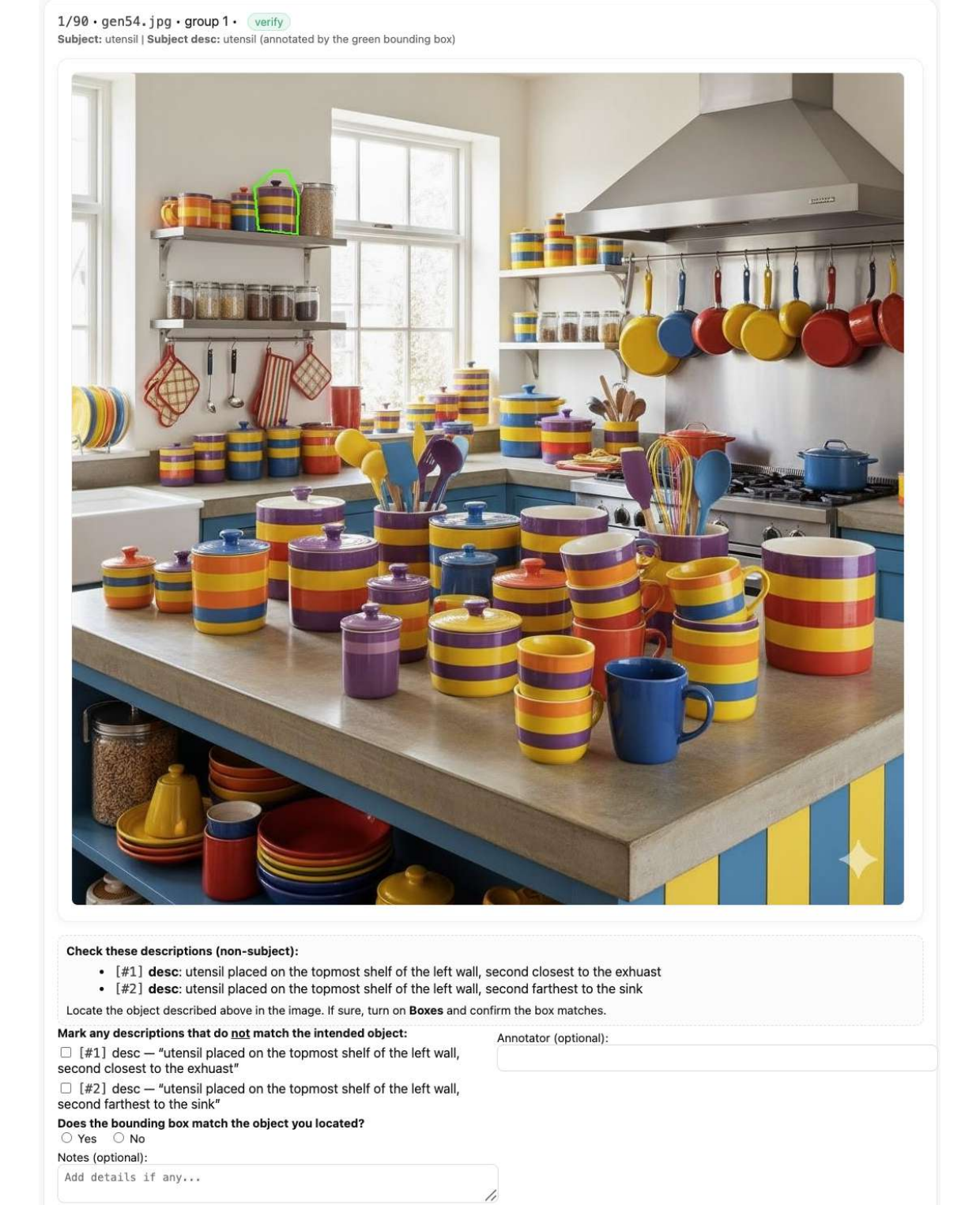}
    \caption{Description with bounding box.}
    \label{fig:desc_box}
\end{subfigure}\hfill
\begin{subfigure}[t]{0.31\linewidth}
    \centering
    \includegraphics[width=\linewidth]{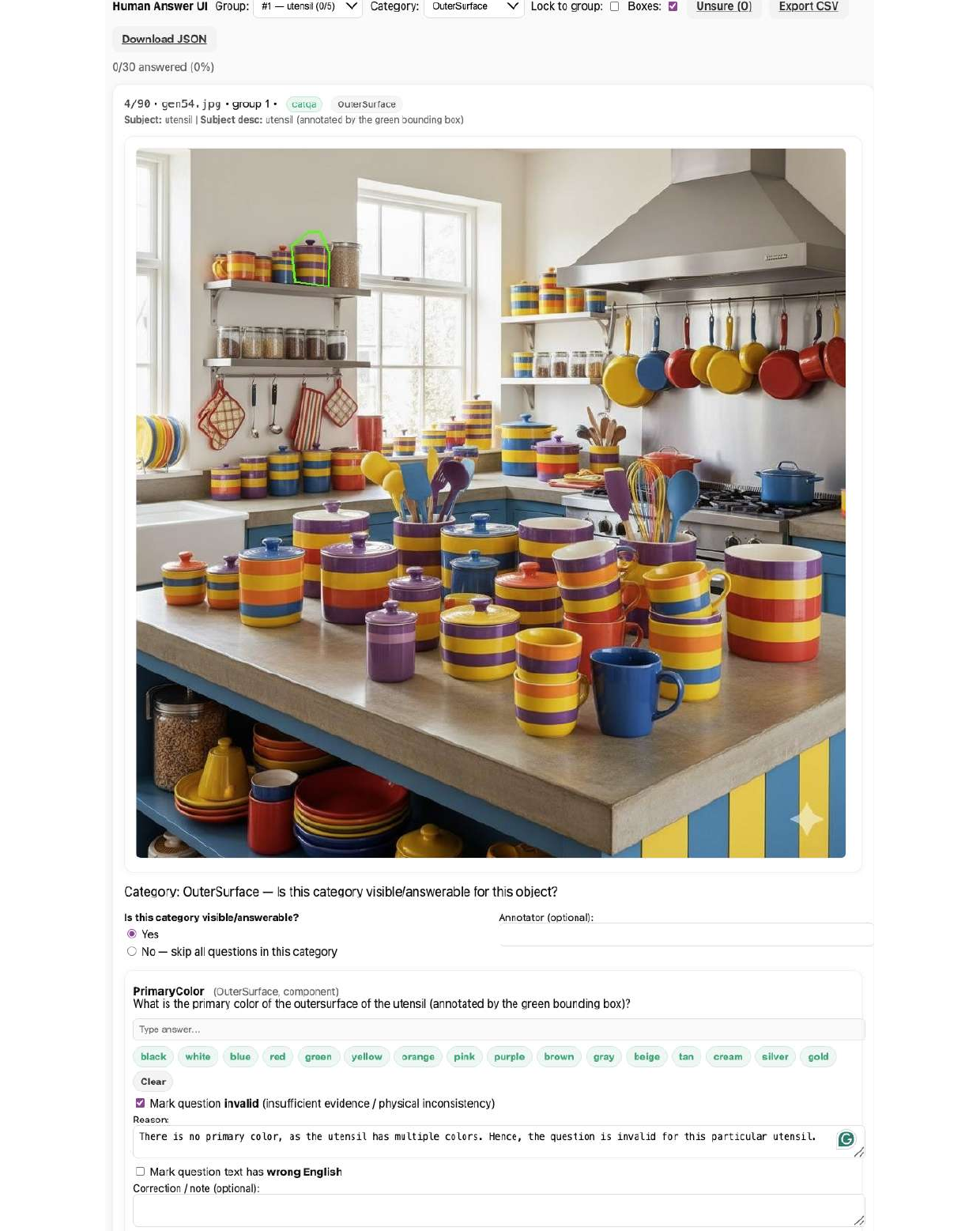}
    \caption{Invalid question example.}
    \label{fig:invalid_question}
\end{subfigure}

\vspace{0.45em}
\begin{subfigure}[t]{0.31\linewidth}
    \centering
    \includegraphics[width=\linewidth]{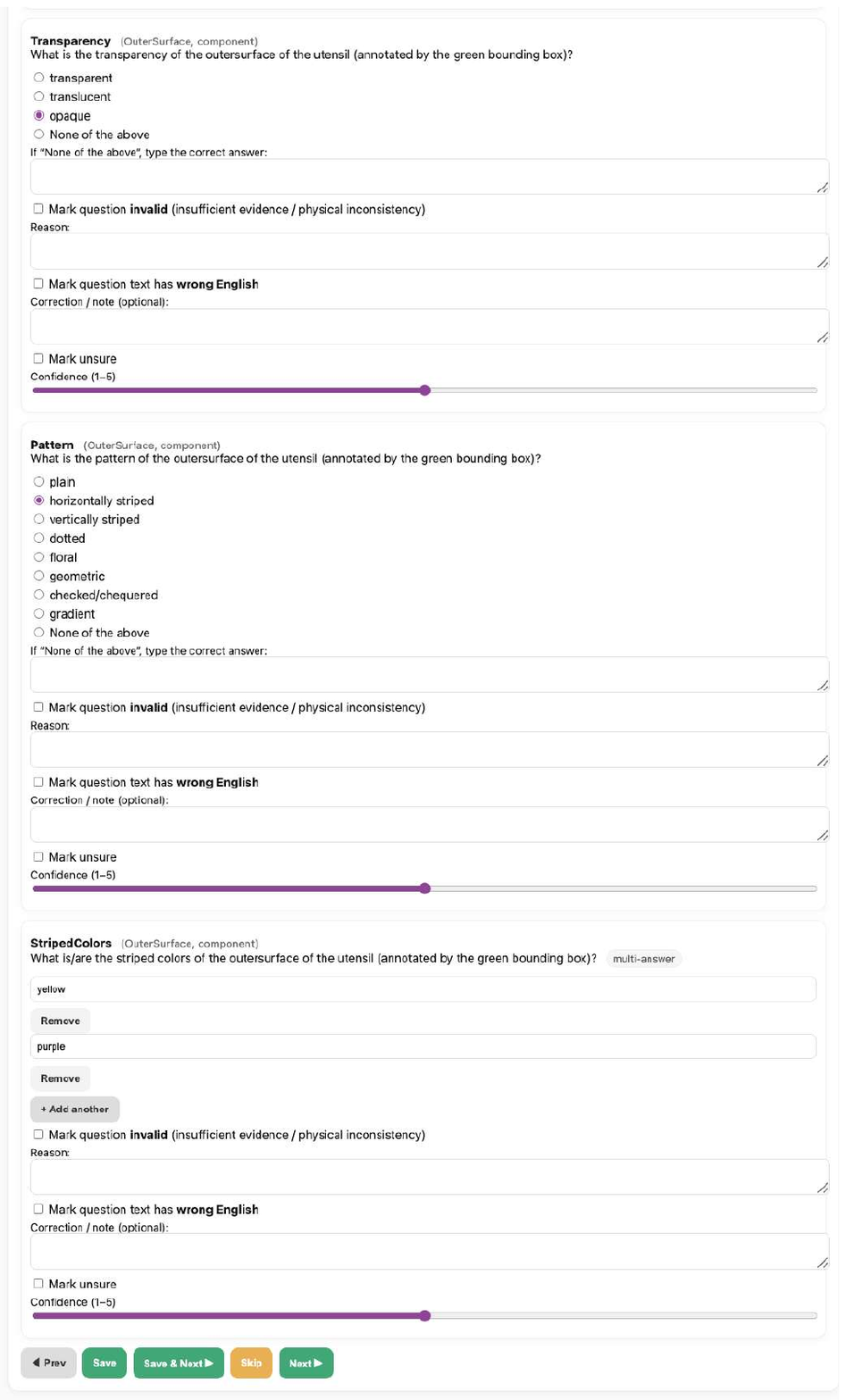}
    \caption{Valid question example.}
    \label{fig:valid_questions}
\end{subfigure}\hfill
\begin{subfigure}[t]{0.31\linewidth}
    \centering
    \includegraphics[width=\linewidth]{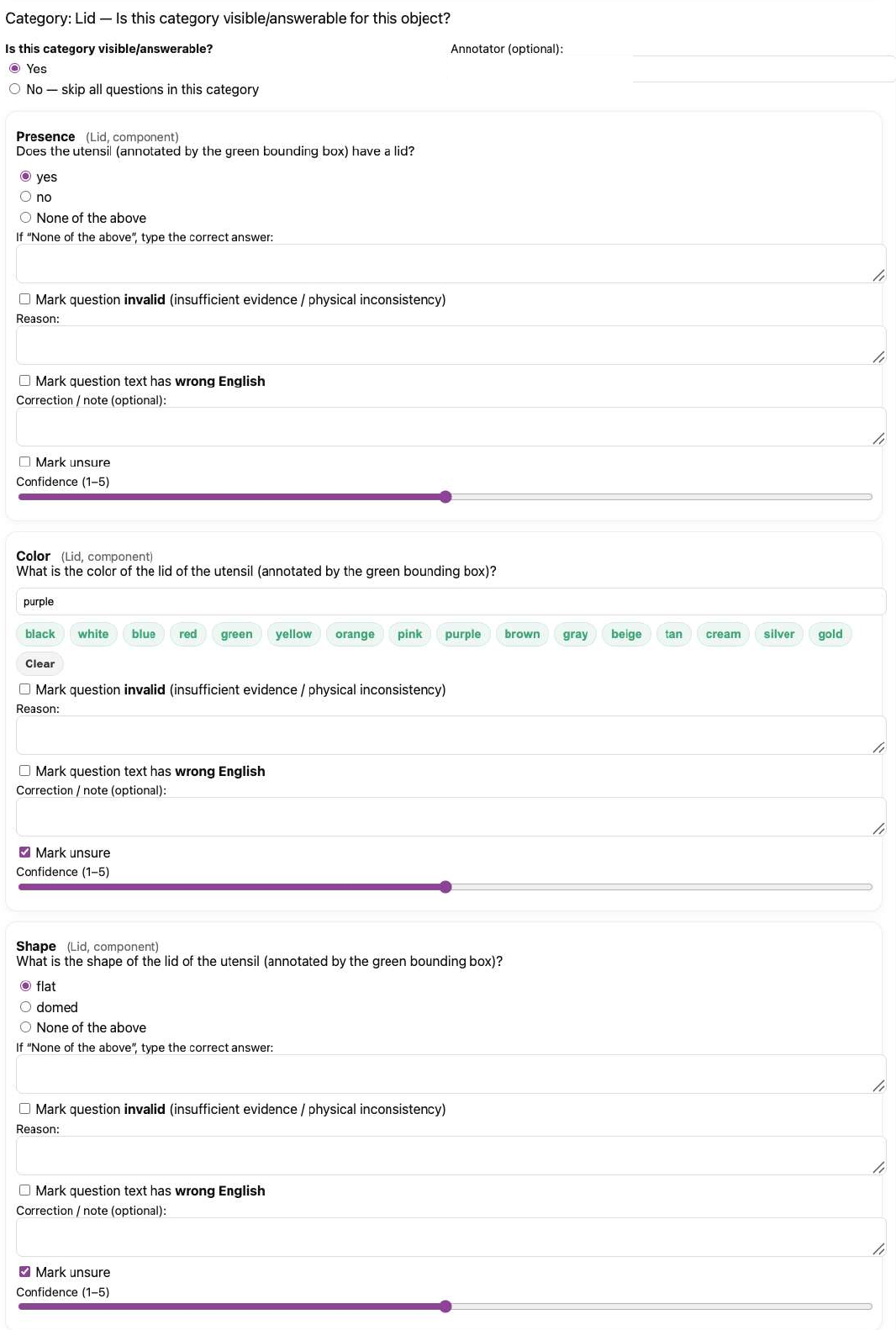}
    \caption{Ambiguous question example.}
    \label{fig:unsure_question}
\end{subfigure}\hfill
\begin{subfigure}[t]{0.31\linewidth}
    \centering
    \includegraphics[width=\linewidth]{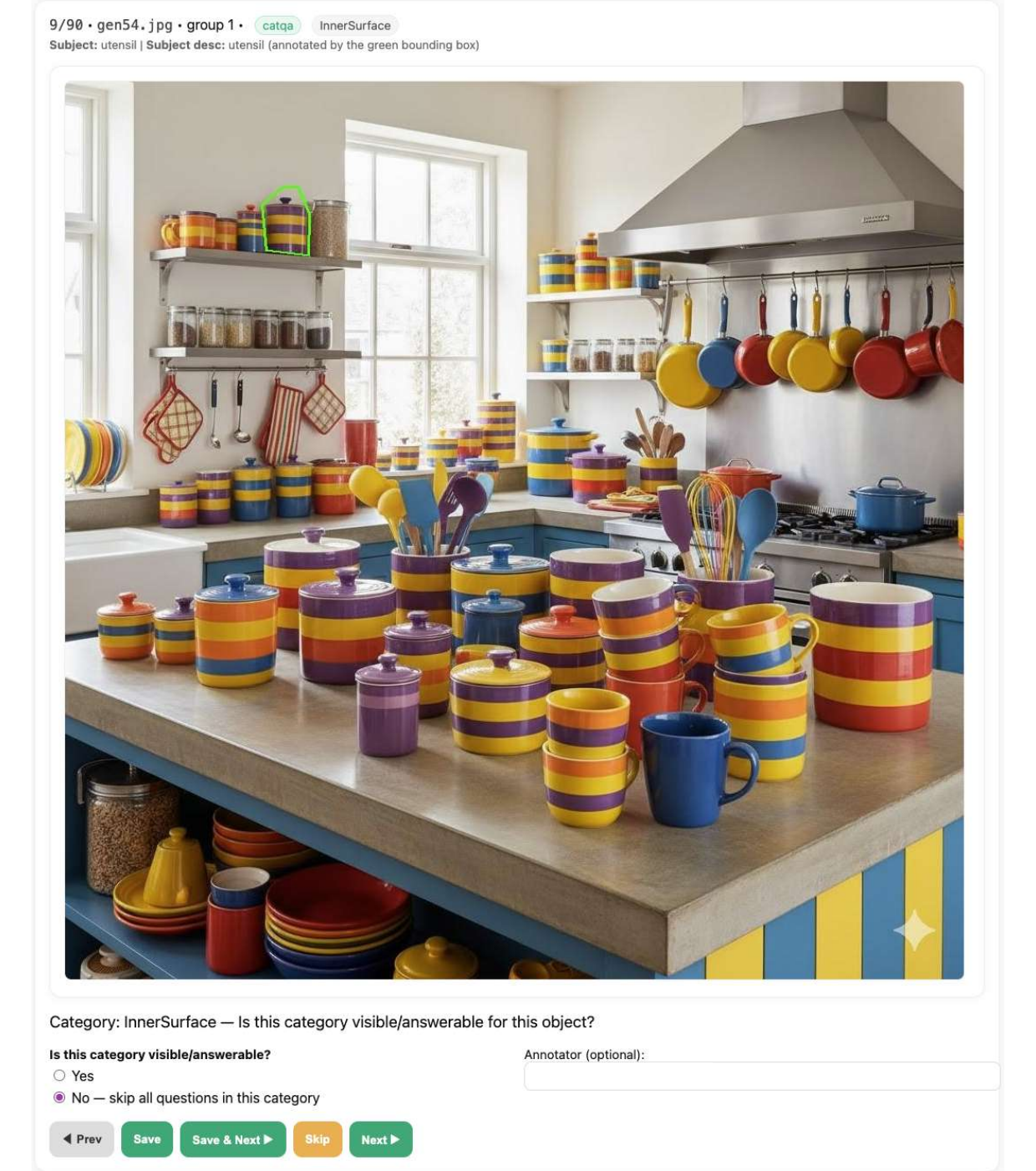}
    \caption{Non-answerable attribute example.}
    \label{fig:non_answerable}
\end{subfigure}

\caption{Human annotation interface examples used for referring-expression verification and QA filtering. Panels show target verification with and without bounding boxes, invalid and valid question cases, ambiguous questions, and non-answerable attributes.}
\label{fig:desc_compare}
\vspace{-10pt}
\end{figure}

\noindent For the first round of human answering, we implemented the following standardized protocol to ensure consistency across annotators:
\begin{itemize}

\item Make sure the referring expression correctly localizes the target object. Annotators should first read the referring expression and identify which object it localizes. Once identified, they can toggle the bounding box to verify that the localized object exactly matches the content of the bounding box. If the expression is ambiguous or does not match the annotated object, the annotator should flag the expression. As shown in Fig.~\ref{fig:desc_compare}, annotators can mark descriptions that do not correspond to the object they initially interpreted. These ambiguous expressions are then refined and sent for a second round of human verification.

\item Once the referring expressions are verified, annotators proceed to answer the questions.

\item While answering, annotators must be confident about their responses; if they are unsure, they can mark the question as \textit{unsure} (as shown in Fig.~\ref{fig:unsure_question}).

\item If the annotators feel that a question is invalid for a particular object, they can mark the question invalid (as shown in Fig.~\ref{fig:invalid_question}), or leave it unanswered.

\item Any component or attribute of an object for which no physically visible answer is present, then it should be marked as unanswerable (as shown in Fig.~\ref{fig:non_answerable}) or left unanswered.

\item If the answer to the question is not included in the options, the annotator is requested to suggest the answer by clicking on `None of the Above'.

\item If any question inadvertently includes the answer itself, human annotators are requested to mark the question as invalid. Such questions are discarded from the benchmark.

\item For numerical counting, annotators are required to report the number of target instances corresponding to each query within a group. They must also verify that the sum of the counts for all child queries matches the count provided for the parent query. If this consistency check fails, the entire group is flagged and returned to the question annotators for refinement.

\end{itemize}

\noindent \textbf{Human Verification}

All answered questions and refined expressions from the human answering stage are forwarded for verification.

\begin{itemize}

\item In this stage, human verifiers confirm that each referring expression uniquely and unambiguously identifies the intended target object. They also ensure that the answer selected in the previous stage objectively and correctly answers the queries. If both conditions are satisfied, the question--answer pair is accepted into the benchmark.

\item When an annotator from the answering stage selected `None of the Above' and provided an alternative answer, the verifier must assess whether this suggestion was necessary and whether it is objectively correct. If the verifier agrees with the suggestion, the question is retained by appending the proposed answer to the option list. The attribute dictionaries are updated accordingly to incorporate this new answer.

\item For numerical counting tasks, annotators check the numerical estimates associated with each query within a group. They additionally confirm that the sum of all child-query counts matches the parent-query count, ensuring hierarchical consistency.

\item For numerical counting, the groups that were revised during the first stage, a joint review by two independent annotators is conducted. This step ensures that all refinements are accurate and that the numerical annotations are fully consistent before inclusion in the benchmark.

\end{itemize}

The entire human annotation process was done by 9 annotators. It was ensured that annotators only verified the questions answered by other annotators.

\noindent \textbf{Human Baseline.} The human baseline reported in Table~\ref{tab:experiment_real_synth} was performed by six individuals who were not involved in the annotation of MIMO-Bench. Each solved 25 groups for attribute binding and 15 for numerical counting.

Refer to Fig.~\ref{fig:question_dict_sample} to see how the attribute binding question for a synthetic image gets narrowed down to a final set of questions. MLLMs are asked the final set of questions for each of the referring expressions. Fig~\ref{fig:num_dict} shows the formation of the numerical counting questions for a synthetic image.

\begin{figure}[!ht]
\centering
\includegraphics[width=\linewidth,height=0.58\textheight,keepaspectratio]{figures/supplementary/question_dict_sample.pdf}
\caption{Sample question dictionary of attribute binding getting modified from raw question answers to the final set of questions, having gone through human answering and verification.}
\label{fig:question_dict_sample}
\end{figure}

\begin{figure}[!ht]
\centering
\includegraphics[width=\linewidth,height=0.45\textheight,keepaspectratio]{figures/supplementary/num_dict.pdf}
\caption{Sample question dictionary for numerical counting task. A parent node will have a non-empty ``children" list as seen for nodes $1$ and $6$. }
\label{fig:num_dict}
\end{figure}

\subsection{Scalability of Benchmark Construction}

MIMO-Bench is designed as a quality-controlled benchmark rather than a fully automatic data-generation pipeline. The most expensive stages are human-dependent: annotators must create or verify referring expressions, ensure that each expression uniquely localizes the intended target, discard ambiguous or non-answerable questions, and verify the final answers. This makes scaling MIMO-Bench challenging compared to benchmarks that rely only on templated generation or automatically mined annotations.

At the same time, the pipeline is partially scalable. Once images, object masks, and referring expressions are available, the attribute dictionary and template-based question generator automatically produce a large pool of candidate attribute-binding questions, while numerical-counting questions are generated from group-level annotations. Human effort is therefore concentrated on filtering and verification rather than writing every question from scratch. This design is intentional: in the MIMO setting, small ambiguity in the referring expression or target object can change the answer, so fully automated scaling would likely introduce noisy QA pairs. We therefore prioritize visually grounded and unambiguous examples over maximizing dataset size. Future extensions can scale the benchmark by improving automatic referring-expression proposal and ambiguity detection, while retaining human verification as the final quality-control step.

\section{MIMO-Crop}\label{app:mimo_crop}

\begin{figure}[tbp]
\centering
\begin{subfigure}{0.47\linewidth}
    \centering
    \includegraphics[width=\linewidth,height=0.55\textheight,keepaspectratio]{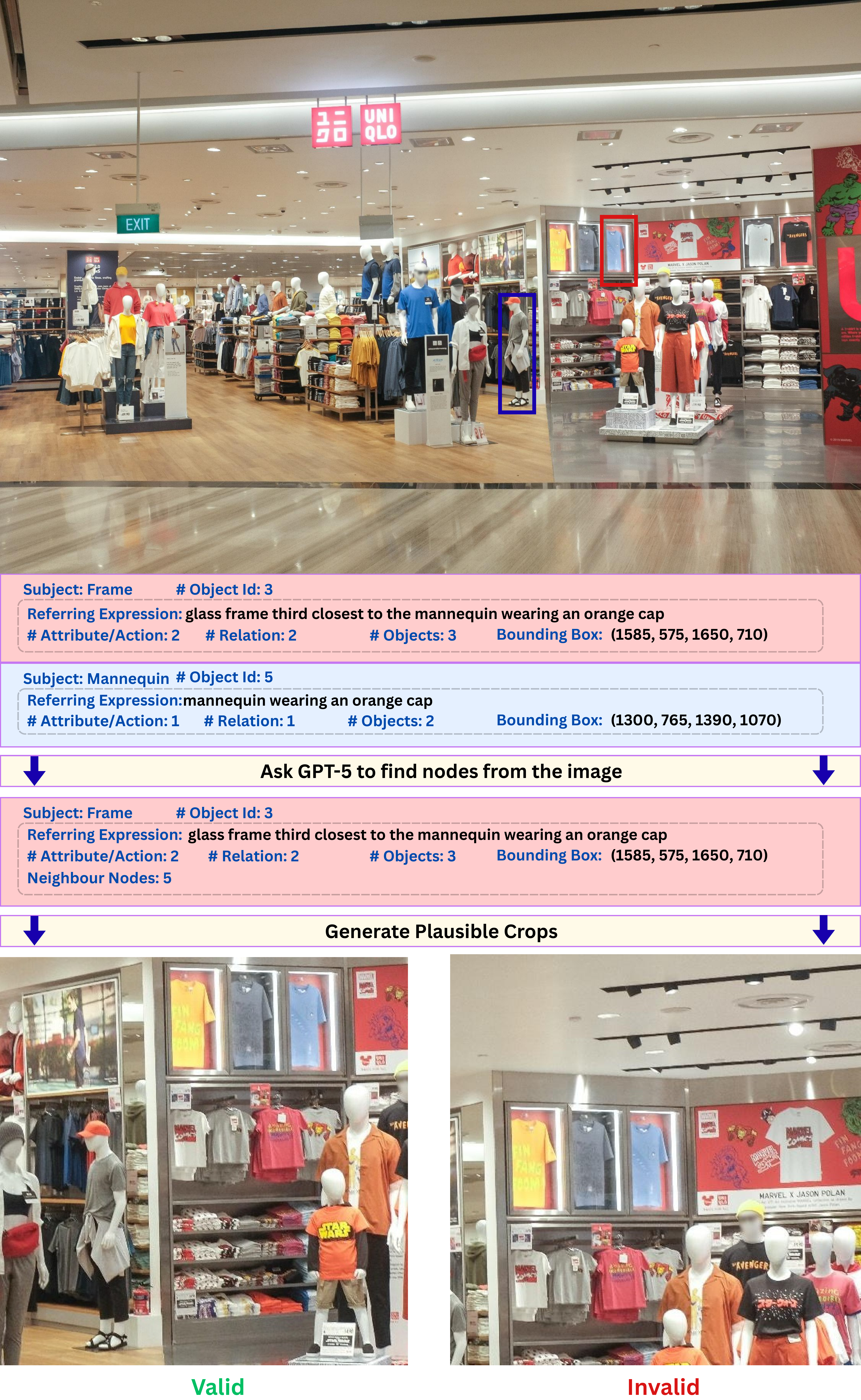}
    \caption{Feasible crop.}
    \label{fig:sg_avail_acc}
\end{subfigure}\hfill
\begin{subfigure}{0.47\linewidth}
    \centering
    \includegraphics[width=\linewidth,height=0.55\textheight,keepaspectratio]{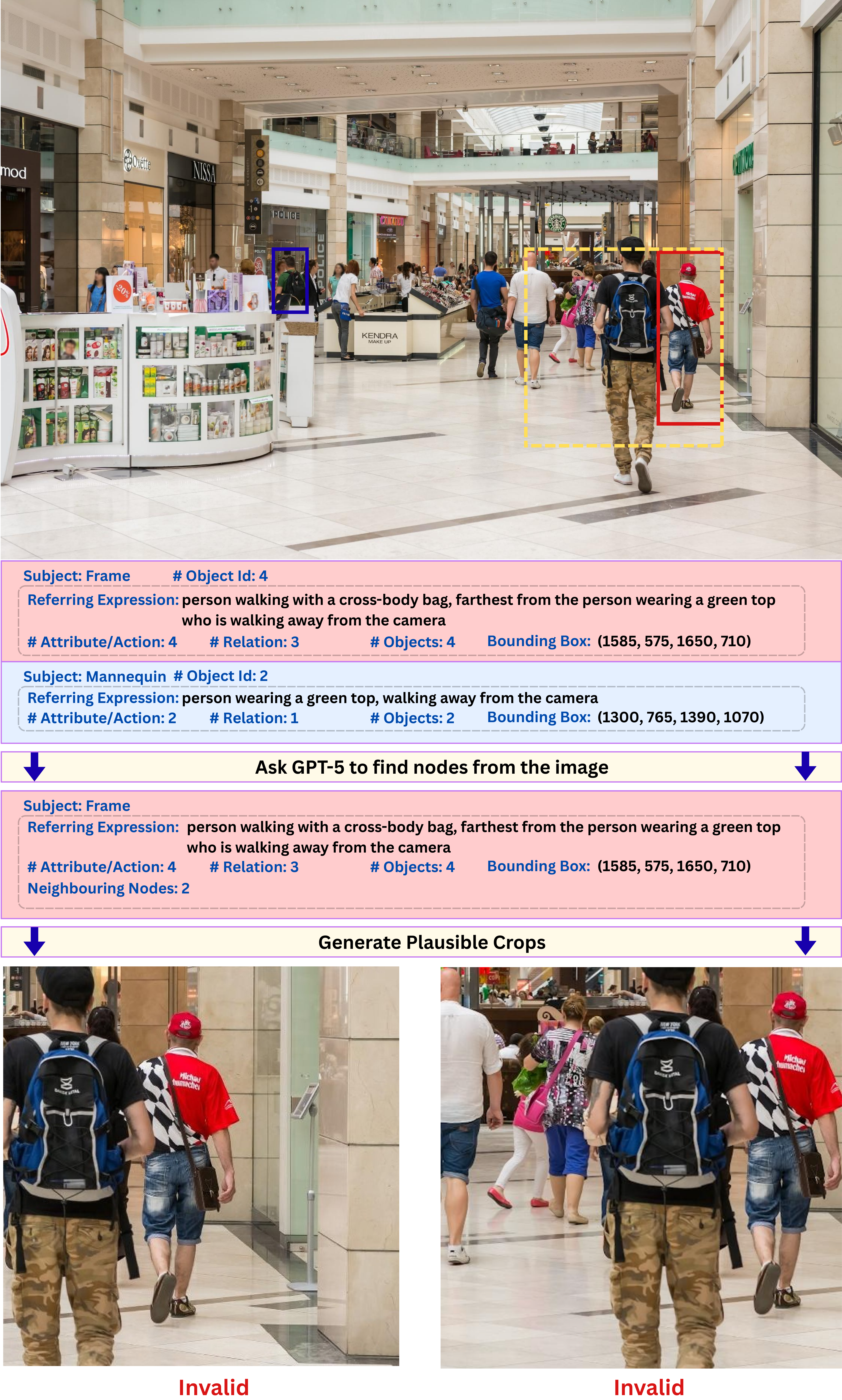}
    \caption{Rejected crop.}
    \label{fig:sg_avail_rej}
\end{subfigure}
\caption{MIMO-Crop examples. Left: a feasible crop where the target and referenced neighboring object are both retained. Right: a rejected case where no crop can preserve all objects mentioned in the referring expression.}
\label{fig:sg_avail_examples}
\end{figure}

We curate MIMO-Crop using a semi-automated pipeline. For each annotated object in an image of MIMO-Bench, we already have a referring expression and a bounding polygon annotation. We want to crop the images from MIMO-Bench in such a manner that the cropped version still uses the same referring expressions. To achieve this, we must ensure that every referenced object in the expression is present in the crop. Otherwise, the crop becomes invalid, as the expression would refer to neighboring objects no longer visible. Therefore, we create a cropped image corresponding to each expression of each annotated object in every image.

For each annotated object in an image, we use GPT-5 to identify how many other nodes (objects) are mentioned in their expression, shown in Fig.~\ref{fig:sg_avail_acc}.
If the target object has annotated neighboring nodes in its expression (Figs.~\ref{fig:sg_avail_acc} and~\ref{fig:sg_avail_rej}), our cropping script aims to create a crop that can contain these neighboring nodes in the crop. As seen in Fig.~\ref{fig:sg_avail_acc}, some object expressions have neighboring nodes' bounding boxes close enough, hence our cropping script automatically includes them in the plausible crops. After generating plausible crops, we send them for human verification. In cases like Fig.~\ref{fig:sg_avail_rej}, no crop can contain the neighboring object referred to in the expression. Consequently, such object-expression pairs are automatically rejected from MIMO-Crop.

In the above cases, GPT-5~\citep{openai2025gpt5} can correctly parse neighboring nodes as corresponding to a single coherent annotated object. However, in some instances, the neighboring nodes do not correspond to a single annotated object, but describe regions or composite structures (``balcony", ``window parapet", and ``street"). For these object--expression pairs, we randomly generate candidate crops around the target object and manually verify whether the referenced components are actually present in the image. If the mentioned nodes cannot be visually confirmed, the corresponding samples are discarded. Throughout the whole process, approximately 7\% of the object-expression pairs were discarded, suggesting that generating random crops for the majority suffices to create a plausible crop.

\begin{figure}[p]
\centering

\begin{tcolorbox}[
    width=0.90\linewidth,
    colback=white,
    colframe=black,
    boxrule=0.4pt,
    arc=0pt,
    left=7pt,
    right=7pt,
    top=7pt,
    bottom=7pt
]
{\scriptsize
\begin{lstlisting}
You are a precise visual question answering assistant.

Answer the question about the provided image.

Question (use EXACTLY as written below; do NOT rewrite or reformat):
{q}

RESPONSE FORMAT (STRICT):
- Write at most one short sentence (optional).
- Then append EXACTLY ONE XML tag on the SAME line at the very end: <answer>SHORT_ANSWER</answer>
- Do NOT add anything after </answer>. Do NOT add extra tags, code fences, or headings.

SHORT_ANSWER RULES:
- If the question text contains labeled options on separate lines (e.g., "A. ...", "B) ...", "1) ..."):
  - Choose the correct option label only (A, B, 1, 2, ...).
  - Output exactly that label inside the tag, e.g., <answer>A</answer>.
- Otherwise (no labeled options in the text):
  - Use a minimal span (e.g., "green", "yes", "no", "3", or an object name).
  - lowercase, no punctuation.
  - If unsure or not visible, use "unknown".

EXAMPLES (follow format precisely):
- The cup appears light green. <answer>green</answer>
- The observation matches choice B. <answer>B</answer>
- Not visible. <answer>unknown</answer>

Now answer:
\end{lstlisting}
}
\end{tcolorbox}

\caption{Prompt used for attribute-binding questions.}
\label{fig:attribute_binding_prompt}
\end{figure}

\begin{figure}[p]
\centering

\begin{tcolorbox}[
    width=0.90\linewidth,
    colback=white,
    colframe=black,
    boxrule=0.4pt,
    arc=0pt,
    left=7pt,
    right=7pt,
    top=7pt,
    bottom=7pt
]
{\scriptsize
\begin{lstlisting}
You are a precise visual counting assistant.

Answer the question about the provided image.

Question (use EXACTLY as written below; do NOT rewrite or reformat):
Count the number of {desc_text}

RESPONSE FORMAT (STRICT):
- Write at most one short sentence (optional).
- Then append EXACTLY ONE XML tag on the SAME line at the very end: <answer>N</answer>
- Do NOT add anything after </answer>. Do NOT add extra tags, code fences, or headings.

COUNTING RULES:
- Count visible instances only; if ambiguous or heavily occluded, count clearly visible ones.
- If none are visible, answer 0.
- Output just the integer N inside <answer>.

EXAMPLES (follow format precisely):
- There are three. <answer>3</answer>
- None present. <answer>0</answer>

Now answer:
\end{lstlisting}
}
\end{tcolorbox}

\caption{Prompt used for numerical-counting questions.}
\label{fig:numerical_counting_prompt}
\end{figure}

\section{Additional Results}\label{app:additional_results}

\subsection{Evaluation Setup}

To ensure consistency across evaluation runs, we use greedy decoding for all models unless specified otherwise. Models such as GPT-5~\citep{openai2025gpt5} (minimal and medium-effort reasoning modes), G3.5-Flash~\citep{google2026gemini35flash} and G3.1-Pro-Preview~\citep{google2026gemini31pro} (low reasoning effort), G2.5 (Pro and Flash)~\citep{comanici2025gemini}, GLM-4.5V~\citep{v2025glm}, IVL3.5-241B-A28B~\citep{wang2025internvl3}, and IVL3-78B~\citep{zhu2025internvl3} are accessed through their respective APIs. Locally evaluated open-weight models, including GLM-4.1V-9B-Thinking, are run from their publicly available HuggingFace weights. Local experiments are conducted on a mix of hardware: a local cluster equipped with two NVIDIA L40S GPUs (48 GB VRAM each) and RunPod instances with an NVIDIA A100 PCIe GPU (80 GB VRAM).

We strictly prompt the models to provide answers within $<$answer$>$ $<$/answer$>$ XML tags. The exact prompts used for attribute binding and numerical counting are shown in Figs.~\ref{fig:attribute_binding_prompt} and~\ref{fig:numerical_counting_prompt}, respectively.

\begingroup
\subsection{Exact Group Accuracy}
\label{app:exact_group_accuracy}

The reported group accuracy $G_{\mathrm{ACC}}$ is group-balanced rather than
all-or-nothing: it averages each target's fraction of correct valid-attribute
answers with equal weight per target. To complement it, Exact Group Accuracy
counts a target as correct only if every valid attribute question is correct.
The word ``group'' names the aggregation unit rather than a strict success
criterion: attributes belonging to one target are scored together, and each
target group receives equal weight regardless of its number of valid attributes.
This prevents richly annotated targets from dominating and is why we retain the
name $G_{\mathrm{ACC}}$ while clarifying the surrounding prose. Table~\ref{tab:exact_group_accuracy} reports the strict Exact Group Accuracy alongside $G_{\mathrm{ACC}}$.

\begin{table}[!htbp]

\centering
\caption{Exact Group Accuracy (\%): the percentage of target groups for which all valid attribute questions are answered correctly.}
\label{tab:exact_group_accuracy}
\scriptsize
\setlength{\tabcolsep}{5pt}
\renewcommand{\arraystretch}{1.04}
\begin{tabular}{@{}lrlr@{}}
\toprule
\textbf{Model} & Exact group (\%) & \textbf{Model} & Exact group (\%) \\
\midrule
Gemini-2.5-Flash & 33.33\% & InternVL3.5-8B & 3.92\% \\
Qwen3-VL-8B-Thinking & 15.83\% & InternVL3.5-38B & 3.61\% \\
InternVL3.5-241B-A28B & 11.75\% & InternVL3-8B & 3.55\% \\
Qwen3-VL-32B-Thinking & 9.78\% & InternVL3-14B & 3.41\% \\
Gemini-2.5-Pro & 9.39\% & Qwen3-VL-8B-Instruct & 3.13\% \\
Qwen2.5-VL-72B-Instruct & 6.94\% & Qwen3-VL-30B-A3B-Thinking & 2.70\% \\
Qwen3-VL-32B-Instruct & 6.14\% & Qwen2.5-VL-7B-Instruct & 2.34\% \\
InternVL3-78B & 5.43\% & Qwen3-VL-30B-A3B-Instruct & 2.08\% \\
GLM-4.1V-9B-Thinking & 5.42\% & InternVL3.5-14B & 1.71\% \\
InternVL3-38B & 5.17\% & GLM-4.5V & 1.15\% \\
Qwen2.5-VL-32B-Instruct & 4.60\% &  &  \\
\bottomrule
\end{tabular}%
\end{table}

Exact Group Accuracy ranges from 1.15\% to 33.33\%, confirming that complete
all-attribute success is substantially rarer than partial within-group accuracy.
The two metrics therefore answer complementary questions.
\endgroup

\subsection{Parent-Level Global--Local Counting Comparison}

\begin{table*}[!t]
\centering
\scriptsize
\setlength{\tabcolsep}{3.0pt}
\renewcommand{\arraystretch}{1.03}
\caption{Parent-level comparison between global and localized counting on the parent groups. ``G$>$L'' and ``L$>$G'' give the percentage of parent groups where the global score is strictly greater or strictly smaller than the localized child average, respectively. sign($\Delta_{GL}$) reports the direction of the gap $\mathcal{G}_{\mathrm{MRA}}-\mathcal{L}_{\mathrm{MRA}}$ over the real and synthetic sets. $\sim$ denotes a near-tie. A negative sign together with L$>$G\,\% exceeding G$>$L\,\% indicates that localized decomposition helps.}
\label{tab:parent_local_gap}
\resizebox{0.62\textwidth}{!}{%
\begin{tabular}{@{}lcrr@{}}
\toprule
\textbf{Model} & \textbf{sign($\Delta_{GL}$)} & \textbf{G$>$L (\%)} & \textbf{L$>$G (\%)} \\
\midrule
G2.5-Flash & $+$ & 42.1 & 43.9 \\
G2.5-Pro & $+$ & 61.4 & 29.8 \\
GPT-5-medium & $+$ & 59.6 & 31.6 \\
Q3-32B-Instruct & $-$ & 31.6 & 54.4 \\
Q3-8B-Instruct & $-$ & 33.3 & 52.6 \\
Q3-30B-A3B-Instruct & $-$ & 29.8 & 56.1 \\
Q2.5-72B-Instruct & $-$ & 40.4 & 49.1 \\
Q2.5-32B-Instruct & $+$ & 49.1 & 31.6 \\
Q2.5-7B-Instruct & $-$ & 28.1 & 47.4 \\
IVL3.5-241B-A28B & $-$ & 0.0 & 17.5 \\
IVL3.5-38B & $-$ & 38.6 & 45.6 \\
IVL3.5-14B & $+$ & 47.4 & 42.1 \\
IVL3.5-8B & $+$ & 47.4 & 47.4 \\
IVL3-78B & $\sim$ & 38.6 & 38.6 \\
IVL3-38B & $-$ & 35.7 & 50.0 \\
IVL3-14B & $+$ & 41.7 & 50.0 \\
IVL3-8B & $-$ & 28.6 & 61.2 \\
GLM-4.5V & $-$ & 40.4 & 49.1 \\
GLM-4.1V-9B-Thinking & $-$ & 0.0 & 66.7 \\
\bottomrule
\end{tabular}%
}
\vspace{-0.4cm}
\end{table*}

\noindent Table~\ref{tab:parent_local_gap} shows that several models benefit more from localized decomposition than from full-scene counting at the parent-group level. This is visible both in the negative sign($\Delta_{GL}$) entries\footnote{For some models the gap on the real set can be larger than on the synthetic set; however, the synthetic set accounts for a larger volume of the numerical-counting parent groups and hence dominates the aggregate direction.} and in the frequency columns: for example, IVL3-8B favors local counting on $61.2\%$ of parent groups versus $28.6\%$ global, and Q3-8B-Instruct on $52.6\%$ versus $33.3\%$. By contrast, G2.5-Pro and GPT-5-medium favor global counting more often than local counting. These parent-level comparisons confirm that the main-text differences between $\mathcal{G}_{\mathrm{MRA}}$ and $\mathcal{L}_{\mathrm{MRA}}$ are not solely artifacts of averaging across groups.

\begingroup
For numerical counting, we additionally test internal hierarchy preservation.
For parent--children group $g$, let
$e_g=\hat{y}^{(g)}_{\mathrm{parent}}-\sum_{c\in\mathcal{C}(g)}\hat{y}^{(g)}_c$.
Exact consistency is the percentage of groups with $e_g=0$; MAE is the mean
$|e_g|$; signed bias is the mean $e_g$. These metrics measure internal
consistency, not correctness against ground truth. Table~\ref{tab:hierarchical_consistency} reports these hierarchical-consistency measures.

\begin{table}[!htbp]

\centering
\caption{Hierarchical consistency of numerical-counting predictions. Negative bias means the predicted parent count is smaller than the sum of predicted child counts.}
\label{tab:hierarchical_consistency}
\scriptsize
\setlength{\tabcolsep}{7pt}
\renewcommand{\arraystretch}{1.04}
\begin{tabular}{@{}lrrr@{}}
\toprule
\textbf{Model} & \textbf{Exact Cons. (\%)} & \textbf{MAE} & \textbf{Bias} \\
\midrule
IVL3.5-241B-A28B & 72.7 & 1.00 & -1.00 \\
Q3-32B-Thinking & 50.0 & 0.75 & -0.35 \\
G3.1-Pro-Preview-Low & 45.3 & 1.55 & -0.19 \\
G3.5-Flash-Low & 45.3 & 1.34 & -0.36 \\
G3.5-Flash-High & 43.4 & 1.58 & -1.02 \\
Q3-8B-Thinking & 42.9 & 0.86 & -0.86 \\
GLM-4.1V-9B-Thinking & 25.0 & 2.00 & -0.75 \\
Q2.5-72B & 22.6 & 3.42 & -1.57 \\
IVL3-78B & 20.8 & 3.62 & -3.36 \\
GPT-5-medium & 19.2 & 2.37 & -0.90 \\
GLM-4.5V & 18.9 & 3.17 & -1.70 \\
Q3-32B-Instruct & 18.9 & 2.87 & -2.34 \\
G2.5-Pro & 17.0 & 2.74 & -1.45 \\
Q3-30B-A3B-Instruct & 17.0 & 4.04 & -3.55 \\
G3.1-Pro-Preview-High & 15.2 & 1.55 & -0.39 \\
Q2.5-32B & 11.3 & 3.53 & -2.70 \\
Q2.5-7B & 11.3 & 3.68 & -2.81 \\
Q3-8B-Instruct & 11.3 & 3.45 & -2.96 \\
G2.5-Flash & 10.2 & 4.84 & -2.76 \\
IVL3-8B & 9.8 & 6.20 & -6.20 \\
IVL3-14B & 9.1 & 5.55 & -4.15 \\
IVL3-38B & 7.7 & 3.71 & -3.25 \\
IVL3.5-38B & 7.5 & 4.42 & -4.00 \\
IVL3.5-14B & 3.8 & 4.72 & -4.42 \\
IVL3.5-8B & 3.8 & 3.98 & -3.79 \\
\bottomrule
\end{tabular}
\end{table}

Consistency varies from 3.8\% to 72.7\%, and signed bias is negative for every
model. Thus, when predictions disagree, the parent estimate is generally below
the sum of localized child estimates. Consistency should not be interpreted as
accuracy: mutually consistent predictions can still differ from ground truth.
\endgroup

\subsection{Full Count-Conditioned Numerical Counting Statistics}

\begin{table*}[!t]
\centering
\scriptsize
\setlength{\tabcolsep}{5.0pt}
\renewcommand{\arraystretch}{1.03}
\caption{Full count-conditioned numerical-counting statistics aggregated across all available model predictions. Rows are grouped by ground-truth (GT) target-set size. We report mean $\mathcal{MRA}$, exact-match rate, signed bias $(\hat{y}-y)$, mean absolute error (MAE), off-by-one rate $(|\hat{y}-y|\leq 1)$, and off-by-two-or-more rate $(|\hat{y}-y|\geq 2)$. Positive bias indicates overcounting and negative bias indicates undercounting. Per-GT sample frequencies are intentionally omitted for compactness.}
\label{tab:count_conditioned_stats}
\resizebox{\textwidth}{!}{%
\begin{tabular}{@{}rrrrrrr@{}}
\toprule
\textbf{GT} & \textbf{$\mathcal{MRA}$} & \textbf{Exact} & \textbf{Bias} & \textbf{MAE} & \textbf{Off-by-1} & \textbf{Off-by-2+} \\
\midrule
0  & 0.3333 & 0.3333 & +1.0833 & 1.0833 & 0.6111 & 0.3889 \\
1  & 0.4069 & 0.4069 & +0.7430 & 0.9272 & 0.8223 & 0.1777 \\
2  & 0.3555 & 0.3555 & +0.4947 & 1.1089 & 0.6838 & 0.3162 \\
3  & 0.4968 & 0.3541 & +0.5889 & 1.4324 & 0.7109 & 0.2891 \\
4  & 0.4263 & 0.2564 & -0.5406 & 1.4124 & 0.5962 & 0.4038 \\
5  & 0.5234 & 0.2888 & -0.6231 & 1.6261 & 0.6231 & 0.3769 \\
6  & 0.4261 & 0.2012 & -1.5045 & 2.2913 & 0.4384 & 0.5616 \\
7  & 0.3270 & 0.0292 & -2.2774 & 3.0949 & 0.2299 & 0.7701 \\
8  & 0.3976 & 0.0714 & -2.0952 & 3.1905 & 0.2619 & 0.7381 \\
9  & 0.4583 & 0.0709 & -2.0000 & 2.9764 & 0.2756 & 0.7244 \\
10 & 0.3734 & 0.0781 & -3.1172 & 3.6953 & 0.1719 & 0.8281 \\
11 & 0.3294 & 0.0196 & -4.5490 & 4.8235 & 0.2549 & 0.7451 \\
12 & 0.3882 & 0.1319 & -3.8889 & 4.3194 & 0.1736 & 0.8264 \\
13 & 0.4941 & 0.0588 & -3.5294 & 3.8824 & 0.1765 & 0.8235 \\
14 & 0.0500 & 0.0000 & -9.5556 & 9.6667 & 0.0556 & 0.9444 \\
16 & 0.4083 & 0.0000 & -5.5278 & 5.6389 & 0.0556 & 0.9444 \\
17 & 0.1750 & 0.0278 & -8.5556 & 11.2222 & 0.0278 & 0.9722 \\
19 & 0.3611 & 0.0000 & -6.0000 & 7.0000 & 0.1111 & 0.8889 \\
\bottomrule
\end{tabular}%
}
\vspace{-0.4cm}
\end{table*}

\noindent Table~\ref{tab:count_conditioned_stats} makes two trends explicit. First, low-count bins are prone to false inclusion: at GT $0$, exact match is only $0.3333$ and the average bias is positive ($+1.0833$). Second, as GT count increases, the dominant failure mode shifts toward undercounting: the bias becomes negative starting at GT $4$ and becomes substantially more negative in high-count bins, while off-by-two-or-more errors dominate. This extends the main-text discussion with the full per-GT error profile while keeping the presentation compact.

\clearpage

Most numerical-counting questions have ground-truth counts between 1 and 12.
Results for individual bins above 12 should therefore be treated as unstable;
aggregate conclusions are based primarily on the well-supported 1--12 range.

\subsection{Visual Prompting Discussion}

\begin{table}[!ht]
\centering
\small
\addtolength{\tabcolsep}{-2.0pt}
\begin{tabular}{l| c |c |c |c} \hline
Method &$512\times 512$ &$1024\times 1024$ &MIMO-Bench & WVP  \\\cline{2-5}
&$\mathcal{I}_{\mathrm{ACC}}$ &$\mathcal{I}_{\mathrm{ACC}}$ &$\mathcal{I}_{\mathrm{ACC}}$ &$\mathcal{I}_{\mathrm{ACC}}$ \\\hline
\multicolumn{5}{c}{\textbf{Evaluated MLLMs}} \\\hline
\multicolumn{5}{l}{\textbf{Qwen3-VL}} \\\hline
32B                 &65.5 &56.8 &46.3 &64.3 \\
30B-A3B             &52.9 &45.2 &35.2 &59.2 \\
8B                  &56.1 &50.6 &39.2 &64.7 \\\hline
\multicolumn{5}{l}{\textbf{Qwen2.5-VL}} \\\hline
32B                 &55.6 &52.7 &40.6 &63.1 \\\hline\hline
\multicolumn{5}{l}{\textbf{InternVL3.5}} \\\hline
38B                 &57.3 &49.3 &41.7 &54.1 \\
14B                 &42.5 &37.0 &35.4 &46.6 \\
8B                  &47.1 &41.5 &39.2 &45.2 \\\hline
\multicolumn{5}{l}{\textbf{InternVL3}} \\\hline
78B                 &49.3 &46.9 &40.5 &--\textsuperscript{\dag}   \\
38B                 &54.1 &50.7 &45.5 &56.3 \\
14B                 &53.2 &48.3 &36.1 &49.3 \\
8B                  &44.0 &38.8 &37.3 &45.9 \\\hline
GLM-4.5V            &61.7 &55.1 &46.2 &64.7 \\\hline
\end{tabular}
\caption{Comparison of MLLM performance on the subset used for visual-prompting and graying ablations. The $512\times512$, $1024\times1024$, and MIMO-Bench columns correspond to the graying setup with progressively more visible distractors, while WVP reports performance with visual prompting. \textsuperscript{\dag}\,We could not obtain a WVP result for IVL3-78B because of API server issues.}
\label{tab:function_distractor}
\end{table}

\begin{figure}[tbp]
\centering
\begin{subfigure}{0.49\linewidth}
    \centering
    \includegraphics[width=\linewidth]{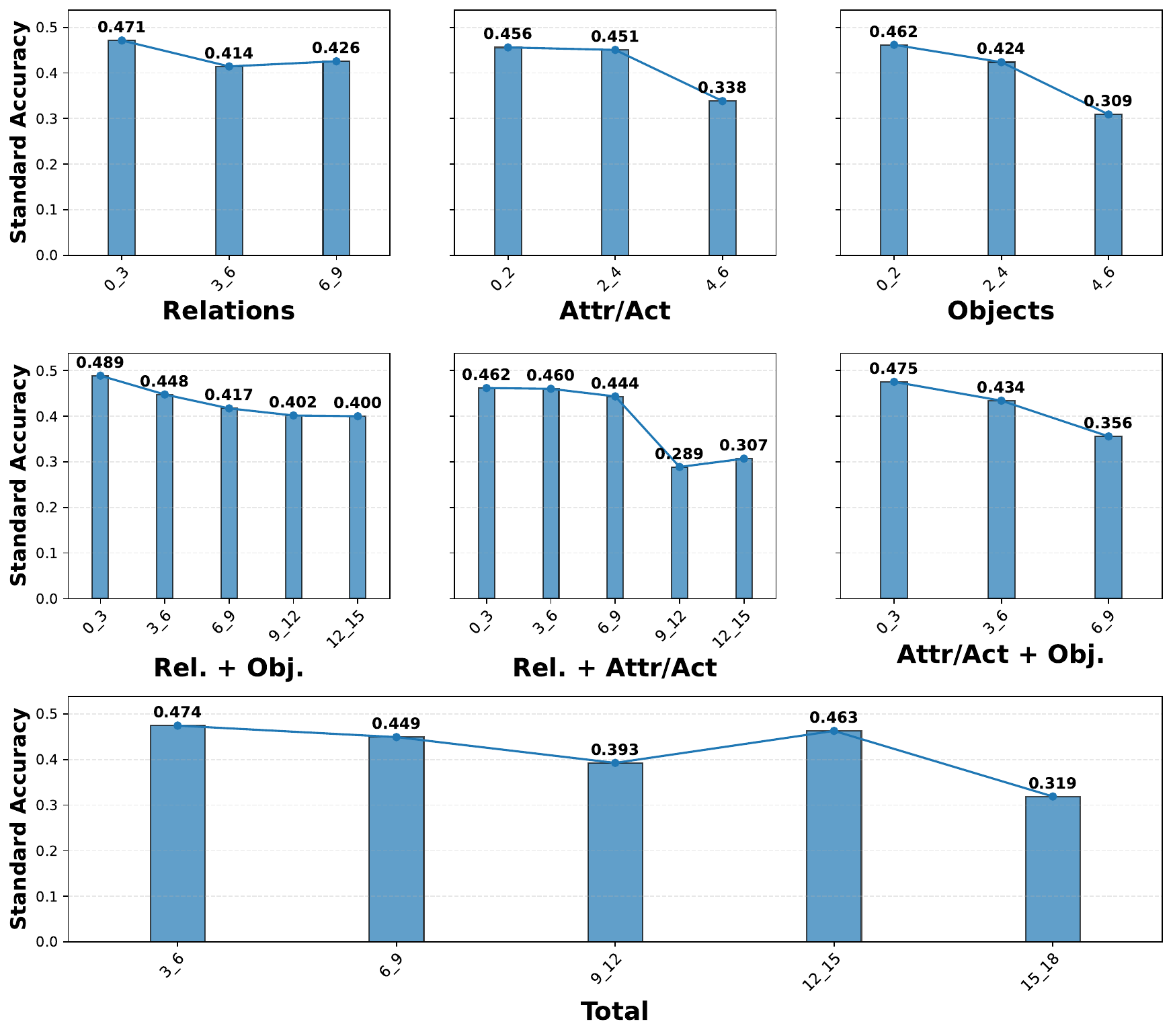}
    \caption{Standard accuracy.}
    \label{fig:prompt_corr_norm}
\end{subfigure}\hfill
\begin{subfigure}{0.49\linewidth}
    \centering
    \includegraphics[width=\linewidth]{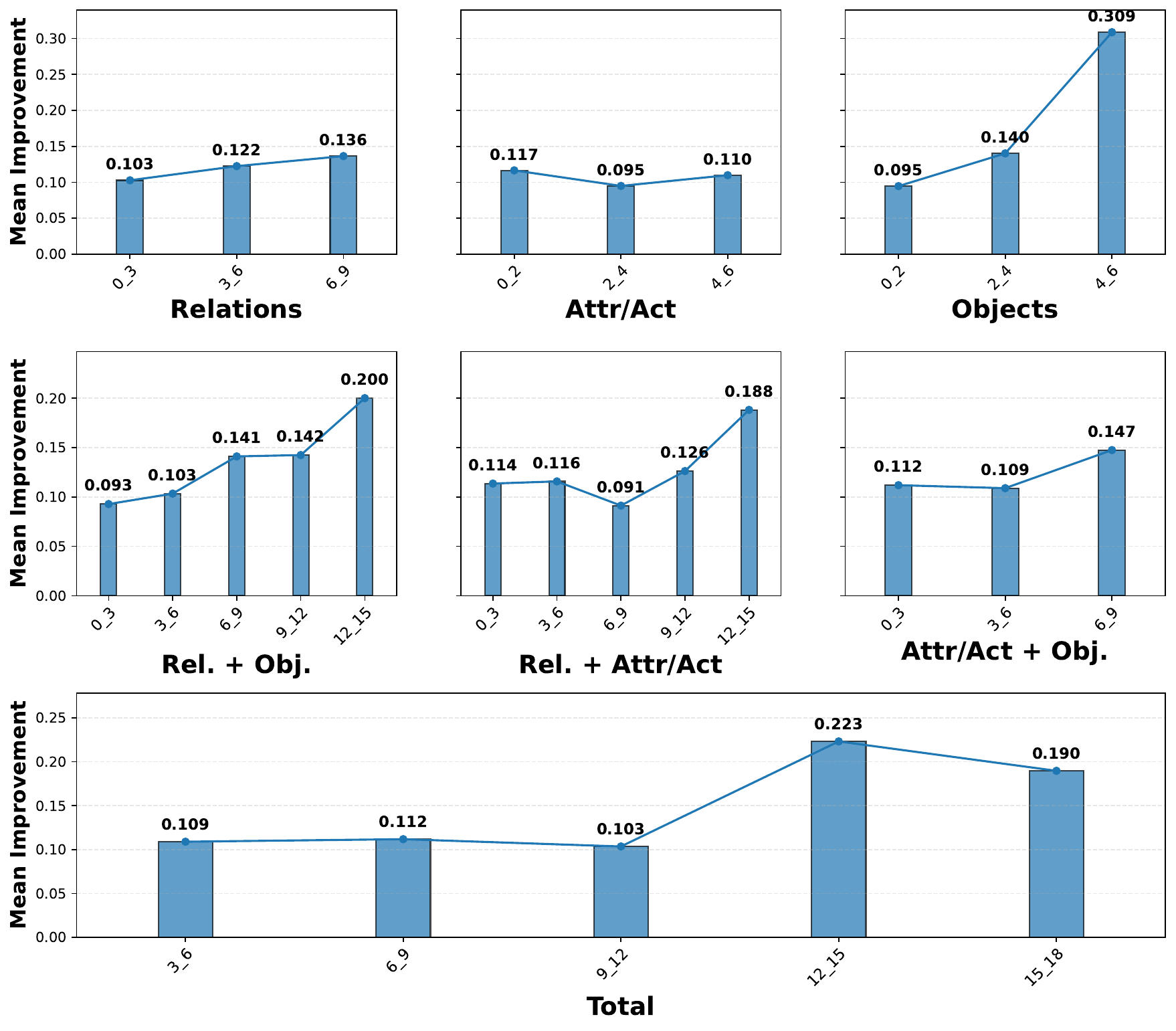}
    \caption{Absolute gain with visual prompting.}
    \label{fig:prompt_corr}
\end{subfigure}
\caption{Compact view of how referring-expression composition relates to standard accuracy and absolute improvement under visual prompting.}
\label{fig:prompt_corr_compact}
\end{figure}

\begin{figure}[tbp]
\centering
\includegraphics[width=0.58\linewidth]{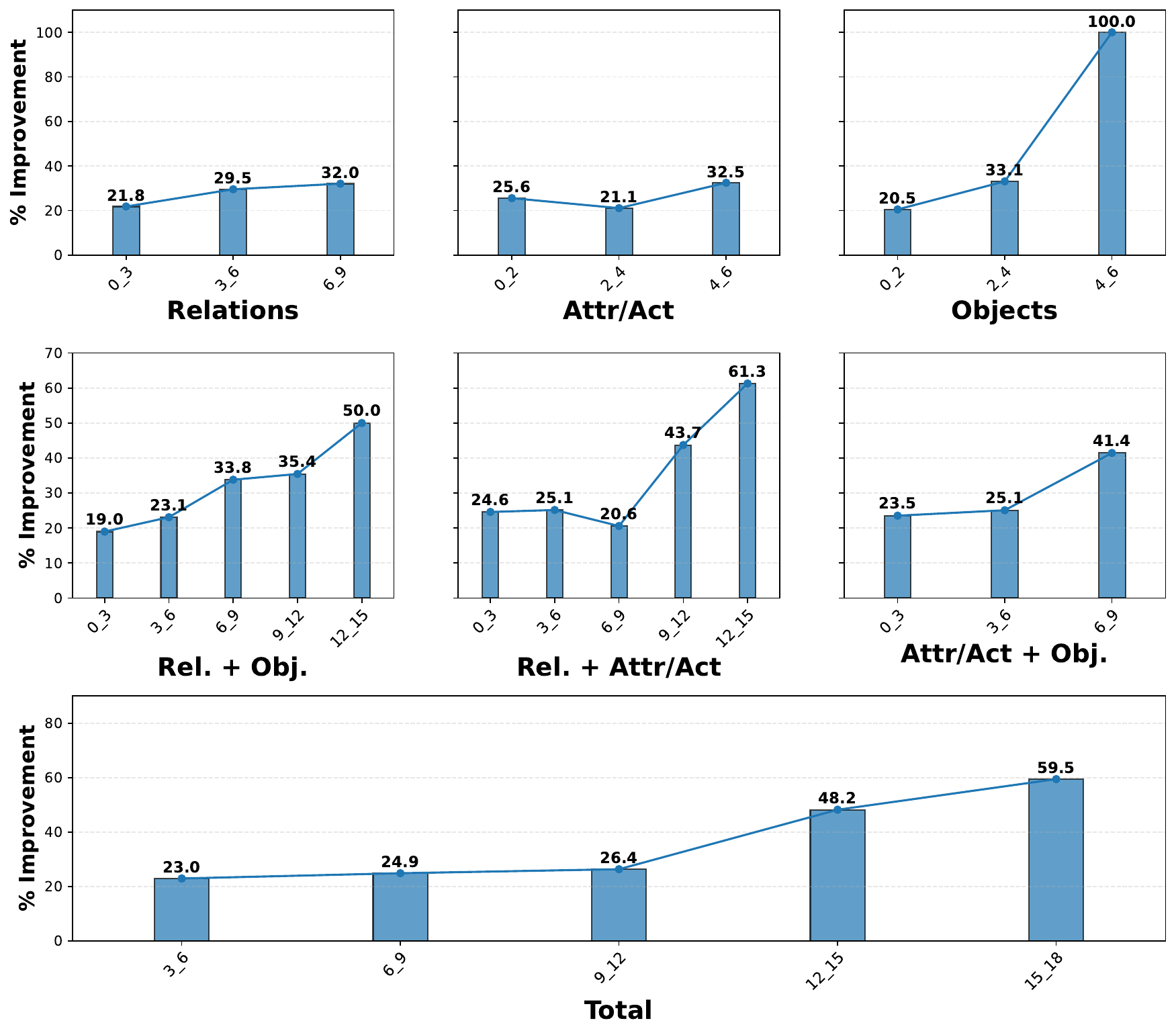}
\caption{Percentage gain with visual prompting.}
\label{fig:prompt_corr_percentage}
\vspace{-10pt}
\end{figure}

To recap, visual prompting in attribute binding helps us in removing the need of referring expressions for isolation of the target object, by replacing them with a visual marker/prompt over the target. Referring expressions, as discussed in Sec.~\ref{sec:tasks_ref} of the main paper, pose multiple subtasks to the model in the quest to localize the target object. Visual prompting essentially reduces the number of visual subtasks posed to MLLMs to 2 (object recognition and bounding box recognition). Kindly refer to  Fig.~\ref{fig:prompt_corr_percentage} to visualize how performance increases as the referring expression gets shorter and a visual marker is used.

Figs.~\ref{fig:prompt_corr_norm},~\ref{fig:prompt_corr}, and~\ref{fig:prompt_corr_percentage} show how questions with varying compositions (attributes/actions, relations, and objects) of  referring expressions behave in different settings. The first row in each figure reports the corresponding performance as the count (binned) of a single component in the expression increases in isolation. The second row reports performance for pairwise combinations of components, whereas the third row reports performance for expressions involving all three components together. The y-axis in all plots shows performance averaged across all models presented in Fig.~\ref{fig:visual_prompting_combined} of the main paper.

Fig.~\ref{fig:prompt_corr_norm} displays how $\mathcal{I}_{\mathrm{ACC}}$ in the standard setting (without visual prompting) varies with the composition of the expression. The results suggest that as the count of the components, such as attributes/actions, relations, and objects, increases, either in isolation or in combination, the accuracy in standard evaluation largely decreases monotonically. The plot essentially tells us that increasing any of the components in the expression leads to a decrease in the average standard accuracy of models, thereby empirically demonstrating the complexity induced by the components of the referring expression. Specifically, we notice that the standard accuracy drops steeply when more than four attributes/actions and objects are present in the expression. In contrast, the standard accuracy does not degrade as steeply when the number of relations in an expression increases.

Figs.~\ref{fig:prompt_corr} and~\ref{fig:prompt_corr_percentage} show the mean absolute and percentage improvement in $\mathcal{I}_{\mathrm{ACC}}$ provided by visual prompting over the standard setting, as a function of the composition of the referring expressions. The main trend suggests that the larger the count of the three components in the expressions, the larger the gains provided by visual prompting. Taken together with the findings from Fig.~\ref{fig:prompt_corr_norm}, this highlights how the volume of subtasks posed by these components, whether in isolation or in combination, substantially hinders MLLMs' performance, and how providing relief via visual prompting is especially beneficial for the most complex expressions. In particular, expressions with higher component counts exhibit both lower standard accuracy and larger improvements with visual prompting, indicating that referring expressions in the MIMO setting are a key bottleneck for current models.

\begingroup
\subsubsection{Grounding and Localization}
\label{app:grounding}

We directly evaluate referring-expression (RE) grounding on the same examples
used for attribute binding (AB). Box-choice grounding asks the model to select
among four marked candidates (25\% chance), whereas coordinate grounding leaves
the image unchanged and requires a bounding-box output in normalized 0--1000
coordinates (approximately
9\% chance under a uniform choice among annotated instances). Each interface is
run jointly with the AB question and separately in isolation. A coordinate
prediction is correct when the annotated instance with maximum IoU is the target.
The four joint cells and conditional probabilities come from the joint run;
isolated grounding accuracy comes from the separate grounding-only run and is
therefore not the sum of the joint $G^+A^+$ and $G^+A^-$ cells. $G^-$ includes
both failures to return a usable location and selections of the wrong instance. Table~\ref{tab:grounding_joint} reports the joint grounding and answer outcomes, and Table~\ref{tab:grounding_conditional} the isolated grounding accuracy together with AB accuracy conditioned on grounding.

\begin{table}[!htbp]

\centering
\caption{Explicit RE-grounding outcomes and their correspondence with AB answers on matched examples. Cells are percentages. $G^+$/$G^-$ denote correct/incorrect grounding and $A^+$/$A^-$ correct/incorrect AB answers.}
\label{tab:grounding_joint}
\scriptsize
\setlength{\tabcolsep}{3.5pt}
\renewcommand{\arraystretch}{1.05}
\begin{tabular}{@{}llrrrr@{}}
\toprule
\textbf{Model} & \textbf{Grounding} & $\mathbf{G^+A^+}$ & $\mathbf{G^+A^-}$ & $\mathbf{G^-A^+}$ & $\mathbf{G^-A^-}$ \\
\midrule
Qwen3-VL-8B & Box-choice & 32.6 & 35.8 & 12.9 & 18.7 \\
 & Coordinates & 14.5 & 9.9 & 33.3 & 42.2 \\
Qwen2.5-VL-7B & Box-choice & 19.7 & 36.8 & 13.7 & 29.8 \\
 & Coordinates & 2.6 & 3.2 & 34.5 & 59.7 \\
InternVL3-8B & Box-choice & 23.2 & 27.6 & 21.6 & 27.6 \\
 & Coordinates & 2.5 & 2.2 & 45.8 & 49.4 \\
InternVL3.5-8B & Box-choice & 18.9 & 33.5 & 15.6 & 32.0 \\
 & Coordinates & 6.0 & 8.6 & 30.2 & 55.3 \\
InternVL3-14B & Box-choice & 26.5 & 23.2 & 21.3 & 29.1 \\
 & Coordinates & 6.3 & 4.4 & 49.6 & 39.7 \\
InternVL3.5-14B & Box-choice & 24.5 & 28.5 & 13.7 & 33.3 \\
 & Coordinates & 5.5 & 5.7 & 35.9 & 52.9 \\
\bottomrule
\end{tabular}%
\end{table}

\begin{table}[!htbp]

\centering
\caption{Grounding-only accuracy and conditional AB accuracy on the same matched examples. Isolated G-Acc. comes from a separate grounding-only run. $\Delta=P(A^+\mid G^+)-P(A^+\mid G^-)$. Joint box-choice AB values use an overlaid image and are not directly comparable with the main benchmark $\mathcal{I}_{\mathrm{ACC}}$ values.}
\label{tab:grounding_conditional}
\scriptsize
\setlength{\tabcolsep}{3.0pt}
\renewcommand{\arraystretch}{1.05}
\begin{tabular}{@{}llrrrr@{}}
\toprule
\textbf{Model} & \textbf{Grounding} & \textbf{Isolated G-Acc.} & $\mathbf{P(A^+\mid G^+)}$ & $\mathbf{P(A^+\mid G^-)}$ & $\boldsymbol{\Delta}$ \\
\midrule
Qwen3-VL-8B & Box-choice & 59.3 & 47.6 & 40.8 & +6.9 \\
 & Coordinates & 25.1 & 59.4 & 44.1 & +15.3 \\
Qwen2.5-VL-7B & Box-choice & 53.8 & 34.8 & 31.4 & +3.4 \\
 & Coordinates & 8.3 & 44.5 & 36.6 & +7.9 \\
InternVL3-8B & Box-choice & 52.9 & 45.7 & 44.0 & +1.7 \\
 & Coordinates & 9.1 & 53.2 & 48.1 & +5.1 \\
InternVL3.5-8B & Box-choice & 56.4 & 36.1 & 32.8 & +3.2 \\
 & Coordinates & 17.5 & 41.0 & 35.3 & +5.7 \\
InternVL3-14B & Box-choice & 54.4 & 53.3 & 42.3 & +11.0 \\
 & Coordinates & 15.3 & 58.7 & 55.5 & +3.2 \\
InternVL3.5-14B & Box-choice & 54.8 & 46.3 & 29.1 & +17.1 \\
 & Coordinates & 17.9 & 48.8 & 40.5 & +8.4 \\
\bottomrule
\end{tabular}%
\end{table}

Isolated box-choice grounding reaches 52.9--59.3\%, while isolated coordinate
grounding reaches 8.3--25.1\%. In the joint runs, correct grounding is associated
with a 1.7--17.1 point improvement in AB accuracy, but
$P(A^+\mid G^+)$ remains only 34.8--53.3\%. Thus, grounding failures and
target-conditioned downstream errors both remain; the association is not
interpreted as a causal decomposition because grounding difficulty can correlate
with answering difficulty and each interface introduces its own requirements.

The two interfaces serve different analytical roles. The isolated coordinate
rate exceeds the corresponding joint coordinate rate for all six models by
0.7--6.7 points, so the format gap is not created by the joint prompt. Relative
to chance, isolated box choice achieves 2.1--2.4 times its 25\% chance level,
whereas isolated coordinates achieve 0.9--2.8 times the approximately 9.0\%
chance level; Qwen2.5-VL-7B and InternVL3-8B are at chance under coordinates.
The coordinate chance level corresponds to a uniform choice among the annotated
instances in an image, whose median count is approximately 14. Much of the raw
accuracy gap between the two interfaces therefore reflects the answer space
shrinking from roughly 14 annotated instances to four marked candidates, rather
than a corresponding change in underlying localization ability.
We therefore use isolated coordinates as the less intrusive localization
readout, while using joint box choice for the conditional analysis: correctly
grounded cases comprise 49.7--68.4\% of the box-choice examples but only
4.7--24.4\% of the coordinate examples, giving a substantially more stable
estimate of $P(A^+\mid G^+)$. This elicitation dependence is also why we did not
make explicit localization a mandatory intermediate benchmark stage.
The released benchmark materials will include the target masks and bounding-box
coordinates so that alternative grounding interfaces can be evaluated.
The annotations already contain bounding-polygon coordinates for every target
object, allowing developers to evaluate either overlaid bounding boxes or raw
coordinate outputs.
\endgroup

\subsection{Graying and Distractor-Reduction Controls}

\begin{figure}[tbp]
\centering
\begin{subfigure}{0.42\linewidth}
    \centering
    \includegraphics[width=\linewidth,height=0.28\textheight,keepaspectratio]{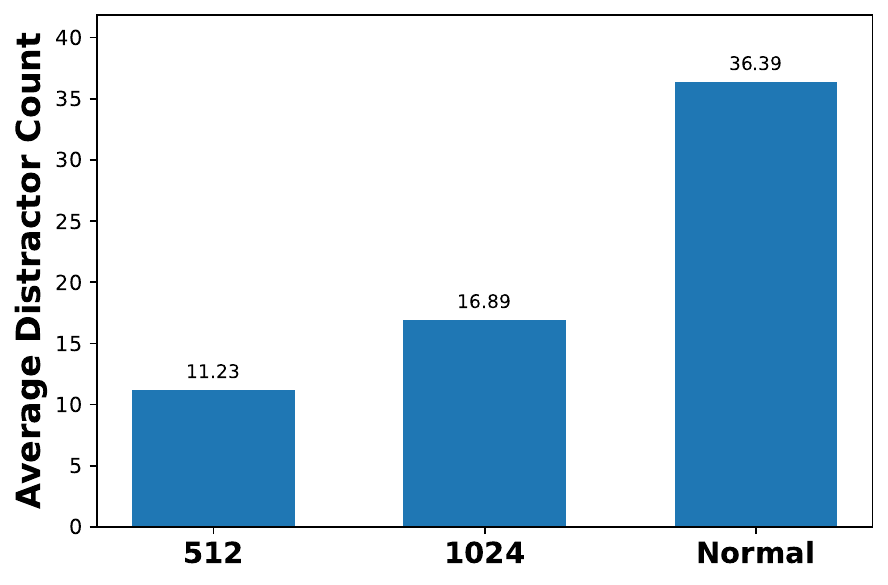}
    \caption{Average distractor count across window sizes.}
    \label{fig:distractor_count}
\end{subfigure}
\hfill
\begin{subfigure}{0.55\linewidth}
    \centering
    \includegraphics[width=\linewidth,height=0.28\textheight,keepaspectratio]{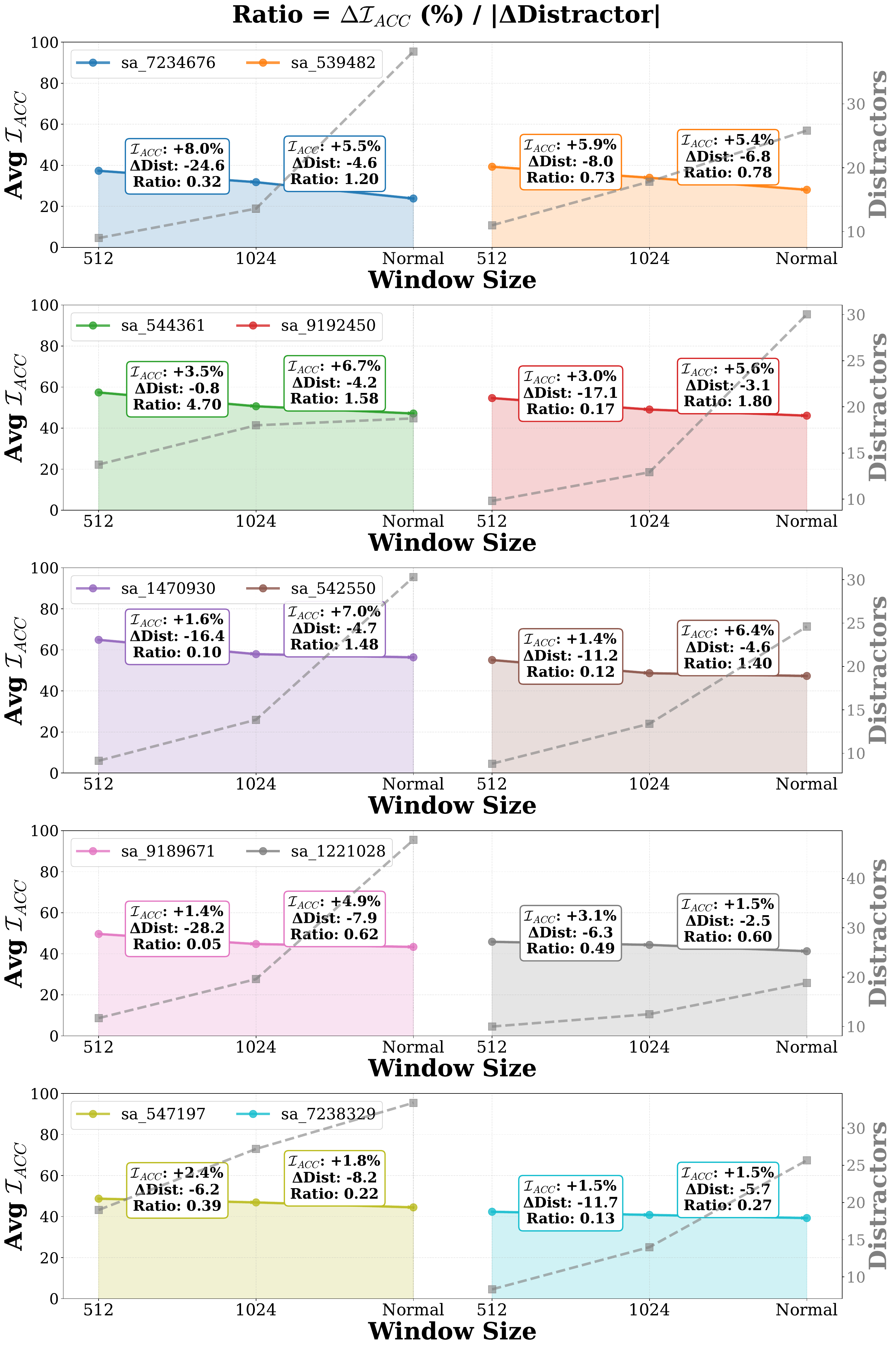}
    \caption{Image-wise relationship between distractor reduction and $\mathcal{I}_{\mathrm{ACC}}$ gain.}
    \label{fig:gray_dist_iacc}
\end{subfigure}
\caption{Graying experiment diagnostics. The first panel summarizes average distractor count; the second shows that the accuracy gain per removed distractor varies substantially across images and window sizes.}
\label{fig:graying_compact}
\end{figure}

We first quantify how the graying intervention changes the number of visible
same-category distractors. We use SAM3~\citep{carion2025sam3segmentconcepts},
a promptable concept segmentation model, to count instances of the target
object category within each retained window. For each window, we supply the
object-category name as a text concept prompt, and SAM3 returns one
instance-level segmentation mask per detected instance. We use the number of
returned masks as the distractor count. Fig.~\ref{fig:distractor_count}
reports the resulting average distractor count for each window size.

Fig.~\ref{fig:gray_dist_iacc} relates this change in distractor burden to model
performance for IVL3-8B, 14B, and 38B; IVL3.5-8B, 14B, and 38B; and
Q3-8B-Instruct and 32B-Instruct over a set of 10 images. For each window size
$w \in \{512, 1024, \text{Normal}\}$, we denote the average
$\mathcal{I}_{\mathrm{ACC}}$ across models for image $i$ by $I_i(w)$ and the
corresponding distractor count by $D_i(w)$.

The callouts in Fig.~\ref{fig:gray_dist_iacc} summarize the change relative to
the unmodified image:
\begin{align}
\Delta I_i(w) &= I_i(w) - I_i(\text{Normal}), \\
\Delta D_i(w) &= D_i(w) - D_i(\text{Normal}), \\
R_i(w) &= \frac{\Delta I_i(w)}{\lvert \Delta D_i(w) \rvert},
\end{align}
where $R_i(w)$ can be interpreted as the accuracy gain per distractor removed.

The trajectories of $I_i(w)$ and $D_i(w)$ are not aligned across images or
window sizes. Although accuracy generally improves as fewer distractors remain
visually active, $R_i(w)$ varies substantially across images and between the
512 and 1024 conditions for the same image. Thus, there is no single global
slope relating distractor count to accuracy, and distractor count alone does
not predict the magnitude of the performance change. Other properties, such as
the spatial configuration of distractors, may also affect the relationship
between distractor reduction and model performance.

\begingroup
\label{app:distractor_controls}

Importantly, graying is not a clean intervention on distractor count alone.
Suppressing the surrounding region also removes contextual information and
increases the relative salience of the retained region. We therefore interpret
graying as an intervention that reduces \emph{effective visual distractor
burden}, rather than as a causal manipulation of distractor count in isolation.

Reducing distractors in a natural image necessarily operates through some
change to the visual input. Cropping removes surrounding content while also
changing target scale and effective resolution. Graying and blurring preserve
the image extent and, for a given model, the visual-token count, but reduce
visual competition by suppressing the surrounding region and consequently
increasing the relative salience of the retained region. These accompanying
changes are not incidental side effects; they are the mechanisms through which
each intervention reduces visual distractor burden. Clean causal identification
of distractor count alone would instead require changing the number of
distractors while holding target appearance, scale, position, background, and
scene structure fixed.

To test whether the observed benefit is specific to grayscale suppression, we
compare graying with two complementary operators, blur and inpainting, over the
same retained regions. Blur suppresses the surrounding region without replacing
it with grayscale values. Inpainting provides the closest practical
approximation to selectively removing same-category distractors while leaving
the remainder of the image unchanged.

For inpainting, the edit mask is defined as the intersection of the suppressed
region with SAM3-segmented polygons of target-category instances. The original
image, region guide, and mask are supplied to the Codex image-generation model,
and the generated content is composited through the same binary mask. Pixels
outside the edit mask are therefore preserved exactly. However, the masks
depend on SAM3 and the generated content inside them may contain segmentation
or inpainting artifacts, so this remains a practical approximation rather than
an exact counterfactual intervention.

Qwen2.5-VL-7B, which was not part of the graying comparison, changes by +0.7
under blur and +0.3 under inpainting in the 1024 condition.
Table~\ref{tab:distractor_operators} reports the $512\times512$ condition and
Table~\ref{tab:distractor_operators_1024} the $1024\times1024$ condition.

\begin{table}[!htbp]

\centering
\caption{Distractor-reduction interventions at $512\times512$ on the 800-sample subset. Base is unmodified-image $\mathcal{I}_{\mathrm{ACC}}$; intervention entries are changes in percentage points. Qwen2.5-VL-7B was not part of the graying comparison and is omitted.}
\label{tab:distractor_operators}
\footnotesize
\setlength{\tabcolsep}{5pt}
\renewcommand{\arraystretch}{1.05}
\begin{tabular}{@{}lrrrr@{}}
\toprule
\textbf{Model} & \textbf{Base} & \textbf{Gray 512} & \textbf{Blur 512} & \textbf{Inpaint 512} \\
\midrule
Qwen3-VL-8B & 39.2 & +16.9 & +10.1 & +0.0 \\
InternVL3-14B & 36.1 & +17.1 & +12.7 & +18.4 \\
InternVL3.5-14B & 35.4 & +7.1 & +10.1 & +12.1 \\
InternVL3-8B & 37.3 & +6.7 & +6.3 & +3.5 \\
InternVL3.5-8B & 39.2 & +7.9 & +4.2 & +5.5 \\
\textbf{Mean} & 37.4 & \textbf{+11.1} & \textbf{+8.7} & \textbf{+7.9} \\
\bottomrule
\end{tabular}%
\end{table}

\begin{table}[!htbp]

\centering
\caption{Distractor-reduction interventions at $1024\times1024$ on the same subset. Entries are changes in percentage points from the same unmodified-image baseline.}
\label{tab:distractor_operators_1024}
\footnotesize
\setlength{\tabcolsep}{5pt}
\renewcommand{\arraystretch}{1.05}
\begin{tabular}{@{}lrrrr@{}}
\toprule
\textbf{Model} & \textbf{Base} & \textbf{Gray 1024} & \textbf{Blur 1024} & \textbf{Inpaint 1024} \\
\midrule
Qwen3-VL-8B & 39.2 & +11.4 & +6.6 & +0.0 \\
InternVL3-14B & 36.1 & +12.2 & +9.9 & +16.6 \\
InternVL3.5-14B & 35.4 & +1.6 & +6.5 & +12.8 \\
InternVL3-8B & 37.3 & +1.5 & +1.9 & +2.4 \\
InternVL3.5-8B & 39.2 & +2.3 & +4.1 & +4.8 \\
\textbf{Mean} & 37.4 & \textbf{+5.8} & \textbf{+5.8} & \textbf{+7.3} \\
\bottomrule
\end{tabular}%
\end{table}

At $512\times512$, the mean gains are 11.1 points for graying, 8.7 for blur,
and 7.9 for inpainting; at $1024\times1024$, the corresponding gains are
5.8, 5.8, and 7.3 points. The model-level ranking varies, and Qwen3-VL-8B
shows no gain under inpainting. Nevertheless, improvements under both blur and
inpainting show that the observed benefit is not specific to grayscale
highlighting. Taken together, these interventions support the broader
interpretation that reducing visual distractor burden can improve MIMO
performance, while not establishing a clean causal effect of distractor count
alone.

\endgroup

\subsection{Naive Relevant-Region Extraction: Mechanism and Controls}\label{app:naive_zooming}

The zooming strategy discussed in Sec.~\ref{sec:what_enhances} of the main paper operates via a two-turn MLLM inference pipeline. Here we describe the mechanism in detail.

\noindent \textbf{Turn 1 --- Region Proposal.} The model is presented with the original image together with the original question. Rather than being asked to answer the question directly, the model is instructed to identify the \emph{region of interest} needed to answer the question and output its spatial extent as a set of bounding-box coordinates (top-left and bottom-right corners, normalized to the image dimensions). The intuition is that the model, even without explicit crop-and-re-query, has sufficient global context to localize the relevant region within a dense, multi-instance scene.

\noindent \textbf{Cropping.} The bounding-box coordinates returned by the model in Turn~1 are used to crop the original image. This step is intended to (i)~\emph{reduce the search space} by discarding irrelevant portions of the image and (ii)~\emph{exclude same-category distractors} that fall outside the selected region. {Cropping also changes target scale, effective resolution and the amount of surrounding context, so these factors are not varied independently.}

\noindent \textbf{Turn 2 --- Answer Generation.} The cropped image is fed to the model along with the original question in a second inference call. Because the crop retains only the selected region, the model now operates on a narrower visual context containing fewer candidate instances.

\noindent \textbf{Interpretation.} MIMO-Bench places multiple same-category instances in close proximity, making it easy for the model to attend to the wrong object. By committing to a single spatial region before answering, the naive strategy narrows the set of candidate instances the model must consider. As reported in Table~\ref{tab:zooming_delta} of the main paper, this two-turn strategy yields consistent accuracy gains across all tested models --- for example, $+16.90$ points for IVL3-14B and $+11.75$ points for IVL3.5-14B . {These gains indicate that successful target-region selection can substantially improve downstream performance in dense MIMO scenes; the experiment does not isolate distractor interference, localization, recognition and attribute binding as independent causal contributors.}

\begingroup
\label{app:region_controls}

We evaluate whether the relevant-region results can be explained by generic
cropping, rescaling, or an additional inference call. Fixed and reproducibly
sampled random windows are evaluated at $512\times512$ and $1024\times1024$.
A target is visible when more than half of its ground-truth box lies inside the
window. Table~\ref{tab:fixed_random_crops} reports these fixed and random crop controls.

\begin{table}[!htbp]

\centering
\caption{Fixed and random crop controls on the 800-sample intervention subset. $\Delta$ is the change from the matched full-image baseline; visibility columns condition accuracy on whether the target remains in the crop.}
\label{tab:fixed_random_crops}
\footnotesize
\setlength{\tabcolsep}{3.0pt}
\renewcommand{\arraystretch}{1.04}
\begin{tabular}{@{}lrrrr@{}}
\toprule
Control & $\Delta$ vs. Base & Target visible & Target absent & Visibility $\Delta$ \\
\midrule
Fixed 1024 & +1.0 & 49.7 & 34.1 & \textbf{+15.6} \\
Random 1024 & +0.8 & 53.5 & 31.9 & \textbf{+21.6} \\
Fixed 512 & -5.3 & 58.3 & 33.2 & \textbf{+25.1} \\
Random 512 & -8.7 & 49.0 & 31.3 & \textbf{+17.7} \\
\bottomrule
\end{tabular}%
\end{table}

The $1024$ crops are essentially unchanged overall, while the $512$ crops reduce
accuracy because they exclude the target more often. Within every condition,
retaining the target is associated with a 15.6--25.1 point advantage. This
direction holds for all 24 model-by-condition comparisons; when the target is
absent, accuracy falls to approximately 31--34\%, close to the majority-answer
baseline. Because resolution is identical within each visibility split,
resolution alone cannot explain this difference.

MIMO-Crop is an annotation-derived crop centered on the target that retains the
objects needed to resolve the referring expression and is independent of the
evaluated MLLM. On the full benchmark it raises mean accuracy from 39.5\% to
48.2\%. Table~\ref{tab:mimo_crop_full} reports the MIMO-Crop reference on the full benchmark.

\begin{table}[!htbp]

\centering
\caption{Full-benchmark performance with the externally supplied
MIMO-Crop$_{512}$ relevant region. This is not a matched baseline for the
800-sample predicted-crop analysis.}
\label{tab:mimo_crop_full}
\small
\setlength{\tabcolsep}{7pt}
\renewcommand{\arraystretch}{1.05}
\begin{tabular}{@{}lrr@{}}
\toprule
\textbf{Model} & \textbf{Base} & \textbf{MIMO-Crop$_{512}$} \\
\midrule
InternVL3-14B & 37.2 & 50.7 \\
InternVL3-8B & 35.9 & 46.8 \\
InternVL3.5-8B & 40.5 & 50.7 \\
Qwen3-VL-32B & 47.5 & 48.9 \\
Qwen3-VL-8B & 40.3 & 44.4 \\
Qwen2.5-VL-7B & 35.8 & 47.6 \\
\textbf{Mean} & \textbf{39.5} & \textbf{48.2} \\
\bottomrule
\end{tabular}
\end{table}

We also condition performance on the IoU of model-selected regions with the
ground-truth target box. Across predicted crops, the mean region is
$459\times479$ pixels (12.43\% of the original image area). The base and IoU
columns use the 800-sample intervention subset. MIMO-Crop$_{512}$ uses the full
benchmark and is shown as an external relevant-region reference rather than a
matched subset baseline. The final column reports the change under naive
model-driven zooming; for Qwen3-VL-32B, the corresponding subset accuracy is
46.3 before and 49.3 after zooming. Table~\ref{tab:iou_localization} reports accuracy conditioned on predicted-crop IoU, and Table~\ref{tab:iou_localization_summary} the corresponding reference and zooming gain per model.

\begin{table}[!htbp]

\centering
\caption{Final-answer accuracy conditioned on predicted-crop IoU with the ground-truth target box. Base and IoU bins use the 800-sample intervention subset. All values are percentages.}
\label{tab:iou_localization}
\scriptsize
\setlength{\tabcolsep}{4.0pt}
\renewcommand{\arraystretch}{1.05}
\begin{tabular}{@{}lrrrrr@{}}
\toprule
\textbf{Model} & \textbf{Base} & \textbf{0--25} & \textbf{25--50} & \textbf{50--75} & \textbf{75--100} \\
\midrule
InternVL3-14B & 36.1 & 48.8 & 51.6 & 56.5 & 52.0 \\
InternVL3-8B & 37.3 & 35.4 & 44.9 & 51.5 & 51.0 \\
InternVL3.5-8B & 39.2 & 37.2 & 42.7 & 36.4 & 45.8 \\
Qwen3-VL-32B & 46.3 & 41.2 & 45.5 & 49.0 & 52.9 \\
Qwen3-VL-8B & 39.2 & 39.3 & 45.5 & 39.0 & 50.8 \\
Qwen2.5-VL-7B & 35.8 & 29.7 & 40.5 & 28.6 & 39.2 \\
\textbf{Mean} & \textbf{39.0} & \textbf{38.6} & \textbf{45.1} & \textbf{43.5} & \textbf{48.6} \\
\bottomrule
\end{tabular}%
\end{table}

\begin{table}[!htbp]

\centering
\caption{Relevant-region reference and zooming gain for the same models. MIMO-Crop$_{512}$ is evaluated on the full benchmark and is therefore an external reference rather than a matched subset baseline. Zoom $\Delta$ is the change under naive model-driven relevant-region extraction.}
\label{tab:iou_localization_summary}
\small
\setlength{\tabcolsep}{7pt}
\renewcommand{\arraystretch}{1.05}
\begin{tabular}{@{}lrr@{}}
\toprule
\textbf{Model} & \textbf{MIMO-Crop$_{512}$} & \textbf{Zoom $\Delta$} \\
\midrule
InternVL3-14B & 50.7 & +16.90 \\
InternVL3-8B & 46.8 & +5.46 \\
InternVL3.5-8B & 50.7 & +2.14 \\
Qwen3-VL-32B & 48.9 & +3.0 \\
Qwen3-VL-8B & 44.4 & +0.89 \\
Qwen2.5-VL-7B & 47.6 & +1.50 \\
\textbf{Mean} & \textbf{48.2} & \textbf{+4.98} \\
\bottomrule
\end{tabular}
\end{table}

Low-IoU regions (0--25\%) achieve 38.6\%, close to the 39.0\% full-image
baseline, whereas high-IoU regions (75--100\%) reach 48.6\%, close to the
48.2\% MIMO-Crop$_{512}$ result. IoU is only an approximate measure of crop
usefulness because a compositional RE can require neighboring context beyond the
target box; we therefore interpret this as an association between better region
selection and higher downstream accuracy.

Finally, a two-pass control keeps the region-proposal turn and second inference
call but reuses the unmodified image in the second turn. Table~\ref{tab:two_pass_no_crop} reports this two-pass control.

\begin{table}[!htbp]

\centering
\caption{Two-pass no-crop control on the 800-sample intervention subset. $\Delta$ is the change in $\mathcal{I}_{\mathrm{ACC}}$ from the matched single-pass full-image condition.}
\label{tab:two_pass_no_crop}
\small
\setlength{\tabcolsep}{8pt}
\renewcommand{\arraystretch}{1.04}

\begin{tabular}{@{}lr@{}}
\toprule
Model & $\Delta$ \\
\midrule
Qwen2.5-VL-7B & +1.0 \\
Qwen3-VL-8B & +1.5 \\
InternVL3-8B & +2.0 \\
InternVL3.5-8B & +1.1 \\
InternVL3-14B & -1.5 \\
InternVL3.5-14B & +3.5 \\
\textbf{Mean} & \textbf{+1.3} \\
\bottomrule
\end{tabular}%
\end{table}

The second inference call alone yields only a 1.3 point mean gain, with no model
gaining more than 3.5 points. Together, these controls show that generic
cropping/rescaling by itself and a second call by itself do not account for the
observed relevant-region gains; the benefit is associated with successfully
retaining useful target-relevant information.

These experiments do not isolate localization, rescaling, distractor reduction,
and additional inference as independent causal contributors. Here, ``dynamic
visual access'' denotes a small exploratory single-pass narrowing study in which
the first inference selects the visual region supplied in the second; it is not
a complete iterative visual-search system. We present it as a proof of concept
that re-accessing a useful region can improve MIMO reasoning and as motivation
for future learned multi-step visual access.
\endgroup

\subsection{Zero-Shot and Self-Consistency CoT: Inference Mechanism}\label{app:cot_mechanism}

Sec.~\ref{sec:what_enhances} of the main paper (``Does CoT Enhance MIMO Performance?'') evaluates Zero-Shot CoT (ZS-CoT) and Self-Consistency CoT (SC-CoT). Both build on the base prompts of Figs.~\ref{fig:attribute_binding_prompt} and~\ref{fig:numerical_counting_prompt}.

\noindent \textbf{ZS-CoT.} The base prompt is appended with \textit{``Let's think step by step,''} which precedes the model's answer. Decoding is greedy (\texttt{do\_sample=False}), and a single generation is used to score accuracy.

\noindent \textbf{SC-CoT.} The same ZS-CoT prompt is reused, but generation is stochastic: \texttt{do\_sample=True} with temperature $0.7$, top-$k=40$, top-$p=1.0$. Five generations are sampled per question, and the final prediction is the \textbf{majority-vote} answer across them (ties broken by first-encountered order).

\begingroup
\subsection{Uncertainty and Statistical Analysis}
\label{app:uncertainty}

We report 95\% nonparametric percentile bootstrap intervals using 20,000
replicates. For the base bootstrap intervals, we resample the unit on which each
metric is defined: image--object--RE--attribute quadruplets for
$\mathcal{I}_{\mathrm{ACC}}$, target groups for $G_{\mathrm{ACC}}$, counting
queries for $\mathcal{I}_{\mathrm{MRA}}$, and parent groups for
$\mathcal{L}_{\mathrm{MRA}}$ and $\mathcal{G}_{\mathrm{MRA}}$. Because questions
sharing an image, target object, or parent query are
not independent, we evaluate progressively coarser resampling units. The central
point estimate is unchanged across columns; only the bootstrap unit differs, with
coarser units giving wider intervals.

Pooled scores use the metric-appropriate aggregation unit: the AB questions for $\mathcal{I}_{\mathrm{ACC}}$, target groups for
$G_{\mathrm{ACC}}$, the NC questions for
$\mathcal{I}_{\mathrm{MRA}}$, and the parent groups for
$\mathcal{L}_{\mathrm{MRA}}$ and $\mathcal{G}_{\mathrm{MRA}}$. Table~\ref{tab:ab_clustered_ci} reports $\mathcal{I}_{\mathrm{ACC}}$ under sample-, object- and image-level resampling, and Table~\ref{tab:ab_clustered_ci_group} the corresponding $G_{\mathrm{ACC}}$ intervals under target-group and image-level resampling. For numerical counting, Tables~\ref{tab:nc_clustered_ci}, \ref{tab:nc_clustered_ci_local} and \ref{tab:nc_clustered_ci_global} report $\mathcal{I}_{\mathrm{MRA}}$, $\mathcal{L}_{\mathrm{MRA}}$ and $\mathcal{G}_{\mathrm{MRA}}$ intervals at the units each metric is defined over.

\begin{table}[!htbp]

\centering
\caption{Attribute-binding $\mathcal{I}_{\mathrm{ACC}}$ point estimates and 95\% bootstrap intervals under sample-, object-, and image-level resampling.}
\label{tab:ab_clustered_ci}
\scriptsize
\setlength{\tabcolsep}{2.6pt}
\renewcommand{\arraystretch}{1.03}
\begin{tabular}{@{}lrrr@{}}
\toprule
Model & Sample & Object & Image \\
\midrule
Gemini-3.5 Flash Low & 54.7 [52.0, 57.3] & 54.7 [51.6, 57.9] & 54.7 [49.7, 62.3] \\
Gemini-3.1 Pro Preview Low & 55.0 [52.3, 57.7] & 55.0 [51.8, 58.3] & 55.0 [50.0, 62.4] \\
Gemini-2.5 Flash & 50.2 [48.1, 52.4] & 50.2 [47.8, 52.7] & 50.2 [47.3, 53.7] \\
Gemini-2.5 Pro & 48.5 [46.7, 50.3] & 48.5 [46.3, 50.6] & 48.5 [45.5, 51.6] \\
GPT-5 minimal & 30.7 [26.8, 35.2] & 30.7 [28.9, 33.1] & 30.7 [29.1, 32.2] \\
GPT-5 medium & 51.8 [49.9, 53.6] & 51.8 [49.6, 53.9] & 51.8 [48.6, 55.0] \\
GLM-4.5V & 47.4 [44.7, 50.0] & 47.4 [44.0, 50.9] & 47.4 [38.7, 55.4] \\
InternVL3.5-241B-A28B & 46.2 [44.3, 48.0] & 46.2 [43.9, 48.3] & 46.2 [43.1, 49.5] \\
InternVL3-78B & 39.3 [37.7, 41.0] & 39.3 [37.4, 41.2] & 39.3 [36.3, 42.8] \\
Qwen3-VL-32B-Instruct & 47.5 [45.8, 49.2] & 47.5 [45.6, 49.4] & 47.5 [44.9, 50.2] \\
Qwen3-VL-8B-Instruct & 40.3 [38.6, 42.1] & 40.3 [38.5, 42.2] & 40.3 [38.0, 42.8] \\
Qwen3-VL-30B-A3B-Instruct & 33.8 [31.9, 35.6] & 33.8 [31.7, 35.9] & 33.8 [28.9, 38.3] \\
Qwen2.5-VL-72B-Instruct & 42.6 [39.5, 45.7] & 42.6 [38.7, 46.6] & 42.6 [36.3, 49.9] \\
Qwen2.5-VL-32B-Instruct & 39.3 [36.7, 41.9] & 39.3 [36.3, 42.5] & 39.3 [35.7, 43.8] \\
Qwen2.5-VL-7B-Instruct & 35.8 [33.8, 37.8] & 35.8 [33.7, 37.9] & 35.8 [32.5, 39.1] \\
InternVL3.5-38B & 42.8 [40.8, 44.8] & 42.8 [40.5, 45.0] & 42.8 [40.0, 45.9] \\
InternVL3.5-14B & 34.4 [32.8, 36.0] & 34.4 [32.7, 36.1] & 34.4 [31.4, 37.6] \\
InternVL3.5-8B & 40.5 [38.9, 42.1] & 40.5 [38.8, 42.1] & 40.5 [37.1, 44.4] \\
InternVL3-38B & 44.3 [42.5, 46.1] & 44.3 [42.2, 46.5] & 44.3 [39.0, 51.3] \\
InternVL3-14B & 37.2 [35.5, 38.9] & 37.2 [35.3, 39.0] & 37.2 [34.7, 40.0] \\
InternVL3-8B & 35.9 [34.3, 37.5] & 35.9 [34.1, 37.7] & 35.9 [33.1, 38.5] \\
GLM-4.1V-9B-Thinking & 36.0 [33.4, 38.5] & 36.0 [33.5, 38.5] & 36.0 [32.5, 38.9] \\
\bottomrule
\end{tabular}%
\end{table}

\begin{table}[!htbp]

\centering
\caption{Attribute-binding $G_{\mathrm{ACC}}$ point estimates and 95\% bootstrap intervals under target-group and image-level resampling.}
\label{tab:ab_clustered_ci_group}
\scriptsize
\setlength{\tabcolsep}{4pt}
\renewcommand{\arraystretch}{1.03}
\begin{tabular}{@{}lrr@{}}
\toprule
Model & Group & Image \\
\midrule
Gemini-3.5 Flash Low & 62.3 [59.0, 65.6] & 62.3 [56.0, 69.7] \\
Gemini-3.1 Pro Preview Low & 61.8 [58.4, 65.3] & 61.8 [54.1, 69.3] \\
Gemini-2.5 Flash & 46.3 [43.7, 48.9] & 46.3 [43.9, 51.4] \\
Gemini-2.5 Pro & 46.9 [44.2, 49.5] & 46.9 [43.7, 55.3] \\
GPT-5 minimal & 27.2 [24.3, 31.2] & 27.2 [23.1, 30.2] \\
GPT-5 medium & 50.4 [47.9, 52.9] & 50.4 [46.4, 54.4] \\
GLM-4.5V & 40.8 [36.9, 44.6] & 40.8 [37.0, 47.2] \\
InternVL3.5-241B-A28B & 43.9 [41.3, 46.5] & 43.9 [39.8, 47.7] \\
InternVL3-78B & 37.3 [35.0, 39.6] & 37.3 [32.3, 44.2] \\
Qwen3-VL-32B-Instruct & 46.5 [44.5, 48.6] & 46.5 [43.3, 50.6] \\
Qwen3-VL-8B-Instruct & 36.6 [34.6, 38.7] & 36.6 [34.8, 41.9] \\
Qwen3-VL-30B-A3B-Instruct & 29.7 [27.3, 32.0] & 29.7 [26.2, 35.4] \\
Qwen2.5-VL-72B-Instruct & 42.3 [38.1, 46.5] & 42.3 [31.1, 50.2] \\
Qwen2.5-VL-32B-Instruct & 39.2 [35.9, 42.4] & 39.2 [35.7, 45.2] \\
Qwen2.5-VL-7B-Instruct & 34.7 [32.2, 37.2] & 34.7 [31.3, 37.7] \\
InternVL3.5-38B & 42.7 [40.2, 45.3] & 42.7 [39.6, 48.0] \\
InternVL3.5-14B & 36.5 [34.7, 38.4] & 36.5 [34.2, 41.2] \\
InternVL3.5-8B & 41.3 [39.4, 43.2] & 41.3 [37.3, 42.9] \\
InternVL3-38B & 43.8 [41.4, 46.2] & 43.8 [37.4, 46.0] \\
InternVL3-14B & 37.3 [35.2, 39.4] & 37.3 [33.9, 41.9] \\
InternVL3-8B & 36.5 [34.6, 38.4] & 36.5 [33.5, 39.3] \\
GLM-4.1V-9B-Thinking & 35.7 [32.6, 38.8] & 35.7 [32.0, 40.0] \\
\bottomrule
\end{tabular}%
\end{table}

\begin{table}[!htbp]

\centering
\caption{Numerical-counting $\mathcal{I}_{\mathrm{MRA}}$ point estimates and 95\% bootstrap intervals under sample-, parent-group-, and image-level resampling.}
\label{tab:nc_clustered_ci}
\scriptsize
\setlength{\tabcolsep}{2.6pt}
\renewcommand{\arraystretch}{1.03}
\begin{tabular}{@{}lrrr@{}}
\toprule
Model & Sample & Group & Image \\
\midrule
Gemini-3.5 Flash Low & 61.5 [56.3, 66.5] & 61.5 [54.3, 68.3] & 61.5 [49.7, 72.1] \\
Gemini-3.1 Pro Preview Low & 61.3 [56.0, 66.5] & 61.3 [54.4, 67.9] & 61.3 [50.3, 71.6] \\
Gemini-2.5 Flash & 40.8 [35.0, 46.5] & 40.8 [34.2, 47.5] & 40.8 [30.9, 51.2] \\
Gemini-2.5 Pro & 49.0 [43.5, 54.5] & 49.0 [42.0, 56.0] & 49.0 [39.9, 57.9] \\
GPT-5 minimal & 44.0 [38.6, 49.3] & 44.0 [37.4, 50.6] & 44.0 [34.5, 53.5] \\
GPT-5 medium & 49.0 [43.8, 54.2] & 49.0 [42.8, 55.3] & 49.0 [41.1, 56.6] \\
GLM-4.5V & 42.3 [37.0, 47.7] & 42.3 [36.4, 48.2] & 42.3 [33.1, 51.3] \\
InternVL3.5-241B-A28B & 49.9 [45.9, 54.4] & 49.9 [42.3, 59.0] & 49.9 [37.4, 65.1] \\
InternVL3-78B & 37.5 [32.2, 42.9] & 37.5 [31.0, 44.3] & 37.5 [28.0, 47.6] \\
Qwen3-VL-32B-Instruct & 47.7 [42.3, 53.1] & 47.7 [41.1, 54.3] & 47.7 [38.2, 57.2] \\
Qwen3-VL-8B-Instruct & 42.6 [37.0, 48.2] & 42.6 [35.9, 49.2] & 42.6 [33.2, 52.1] \\
Qwen3-VL-30B-A3B-Instruct & 35.0 [29.8, 40.3] & 35.0 [28.4, 41.9] & 35.0 [26.3, 43.4] \\
Qwen2.5-VL-72B-Instruct & 40.4 [35.0, 45.7] & 40.4 [34.1, 46.8] & 40.4 [30.3, 50.7] \\
Qwen2.5-VL-32B-Instruct & 37.3 [32.1, 42.5] & 37.3 [30.9, 43.8] & 37.3 [27.6, 47.8] \\
Qwen2.5-VL-7B-Instruct & 31.4 [26.6, 36.4] & 31.4 [25.0, 38.0] & 31.4 [21.2, 42.4] \\
InternVL3.5-38B & 36.8 [31.6, 42.1] & 36.8 [30.6, 43.1] & 36.8 [28.8, 44.5] \\
InternVL3.5-14B & 38.2 [32.9, 43.6] & 38.2 [32.3, 44.1] & 38.2 [28.4, 48.5] \\
InternVL3.5-8B & 41.2 [35.7, 46.7] & 41.2 [35.5, 47.2] & 41.2 [32.3, 51.0] \\
InternVL3-38B & 41.0 [35.5, 46.6] & 41.0 [34.4, 47.6] & 41.0 [32.9, 48.7] \\
InternVL3-14B & 33.2 [26.8, 39.8] & 33.2 [26.1, 41.0] & 33.2 [24.0, 42.4] \\
InternVL3-8B & 33.3 [27.5, 39.0] & 33.3 [27.3, 38.4] & 33.3 [26.5, 39.4] \\
GLM-4.1V-9B-Thinking & 25.8 [16.8, 35.3] & 25.8 [17.6, 34.7] & 25.8 [17.4, 34.3] \\
\bottomrule
\end{tabular}%
\end{table}

\begin{table}[!htbp]

\centering
\caption{Numerical-counting $\mathcal{L}_{\mathrm{MRA}}$ point estimates and 95\% bootstrap intervals under parent-group and image-level resampling.}
\label{tab:nc_clustered_ci_local}
\scriptsize
\setlength{\tabcolsep}{4pt}
\renewcommand{\arraystretch}{1.03}
\begin{tabular}{@{}lrr@{}}
\toprule
Model & Group & Image \\
\midrule
Gemini-3.5 Flash Low & 60.8 [52.2, 68.9] & 60.8 [47.6, 73.0] \\
Gemini-3.1 Pro Preview Low & 60.5 [52.2, 68.3] & 60.5 [51.0, 70.1] \\
Gemini-2.5 Flash & 37.1 [28.9, 45.5] & 37.1 [27.0, 48.0] \\
Gemini-2.5 Pro & 44.0 [35.7, 52.6] & 44.0 [32.2, 55.8] \\
GPT-5 minimal & 42.9 [35.1, 50.7] & 42.9 [32.8, 52.9] \\
GPT-5 medium & 46.7 [40.0, 53.7] & 46.7 [37.7, 55.9] \\
GLM-4.5V & 44.6 [37.8, 51.5] & 44.6 [35.0, 54.4] \\
InternVL3.5-241B-A28B & 50.4 [42.1, 60.1] & 50.4 [33.9, 69.6] \\
InternVL3-78B & 37.1 [29.5, 44.9] & 37.1 [28.1, 45.9] \\
Qwen3-VL-32B-Instruct & 48.7 [40.8, 56.5] & 48.7 [38.0, 59.1] \\
Qwen3-VL-8B-Instruct & 42.4 [34.6, 50.3] & 42.4 [30.8, 53.9] \\
Qwen3-VL-30B-A3B-Instruct & 41.6 [34.0, 49.6] & 41.6 [31.2, 52.9] \\
Qwen2.5-VL-72B-Instruct & 43.5 [36.9, 50.2] & 43.5 [34.5, 52.2] \\
Qwen2.5-VL-32B-Instruct & 36.3 [29.0, 43.6] & 36.3 [26.8, 46.4] \\
Qwen2.5-VL-7B-Instruct & 32.4 [24.8, 40.2] & 32.4 [22.6, 42.7] \\
InternVL3.5-38B & 41.5 [33.5, 49.4] & 41.5 [30.9, 51.7] \\
InternVL3.5-14B & 38.5 [31.2, 46.1] & 38.5 [29.9, 46.6] \\
InternVL3.5-8B & 42.6 [35.1, 50.2] & 42.6 [34.9, 50.0] \\
InternVL3-38B & 46.5 [38.3, 55.0] & 46.5 [36.5, 57.0] \\
InternVL3-14B & 34.1 [25.6, 43.3] & 34.1 [24.4, 43.8] \\
InternVL3-8B & 39.8 [22.6, 46.2] & 39.8 [28.4, 47.7] \\
GLM-4.1V-9B-Thinking & 30.2 [19.4, 41.9] & 30.2 [18.8, 41.7] \\
\bottomrule
\end{tabular}%
\end{table}

\begin{table}[!htbp]

\centering
\caption{Numerical-counting $\mathcal{G}_{\mathrm{MRA}}$ point estimates and 95\% bootstrap intervals under sample- and image-level resampling.}
\label{tab:nc_clustered_ci_global}
\scriptsize
\setlength{\tabcolsep}{4pt}
\renewcommand{\arraystretch}{1.03}
\begin{tabular}{@{}lrr@{}}
\toprule
Model & Sample & Image \\
\midrule
Gemini-3.5 Flash Low & 60.7 [52.5, 68.5] & 60.7 [50.1, 70.0] \\
Gemini-3.1 Pro Preview Low & 63.8 [55.6, 71.7] & 63.8 [53.2, 73.0] \\
Gemini-2.5 Flash & 47.3 [37.1, 57.5] & 47.3 [32.7, 61.4] \\
Gemini-2.5 Pro & 66.4 [58.0, 74.5] & 66.4 [56.8, 76.1] \\
GPT-5 minimal & 43.8 [34.8, 52.8] & 43.8 [31.8, 55.8] \\
GPT-5 medium & 55.1 [46.1, 63.8] & 55.1 [42.2, 67.2] \\
GLM-4.5V & 38.5 [30.0, 46.9] & 38.5 [27.1, 49.5] \\
InternVL3.5-241B-A28B & 46.8 [44.3, 51.7] & 46.8 [44.7, 51.0] \\
InternVL3-78B & 35.6 [26.5, 45.1] & 35.6 [22.0, 49.8] \\
Qwen3-VL-32B-Instruct & 51.4 [42.5, 60.4] & 51.4 [39.4, 63.5] \\
Qwen3-VL-8B-Instruct & 38.0 [28.4, 48.4] & 38.0 [25.3, 51.3] \\
Qwen3-VL-30B-A3B-Instruct & 32.9 [23.6, 42.4] & 32.9 [21.9, 44.5] \\
Qwen2.5-VL-72B-Instruct & 36.8 [26.8, 46.8] & 36.8 [23.5, 50.2] \\
Qwen2.5-VL-32B-Instruct & 41.2 [32.4, 50.0] & 41.2 [28.6, 53.8] \\
Qwen2.5-VL-7B-Instruct & 28.6 [20.5, 37.0] & 28.6 [17.6, 41.1] \\
InternVL3.5-38B & 35.1 [26.1, 44.2] & 35.1 [25.1, 44.8] \\
InternVL3.5-14B & 37.8 [28.6, 46.9] & 37.8 [25.2, 50.4] \\
InternVL3.5-8B & 35.1 [25.4, 44.7] & 35.1 [23.3, 47.0] \\
InternVL3-38B & 33.4 [24.5, 42.7] & 33.4 [19.5, 47.7] \\
InternVL3-14B & 30.0 [18.6, 41.7] & 30.0 [16.5, 44.6] \\
InternVL3-8B & 18.4 [10.4, 27.0] & 18.4 [5.6, 25.3] \\
GLM-4.1V-9B-Thinking & 29.5 [13.3, 48.7] & 29.5 [9.5, 53.9] \\
\bottomrule
\end{tabular}%
\end{table}

Coarser clustering generally widens the intervals by approximately 1.3--1.8
times, consistent with within-image dependence, although the ordering is not
monotone for every model.

For matched intervention conditions, we preserve example-level pairing and
report the paired change with its bootstrap interval. Visual prompting replaces
the textual RE with a ground-truth visual cue. We define the paired visual-prompting
change as
\[
\Delta_{\mathrm{VP}}
=
\mathcal{I}_{\mathrm{ACC}}(\text{with visual prompting})
-
\mathcal{I}_{\mathrm{ACC}}(\text{without visual prompting}).
\]

Table~\ref{tab:visual_prompting_ci} reports these paired visual-prompting changes.

\begin{table}[!htbp]

\centering
\caption{Paired change in $\mathcal{I}_{\mathrm{ACC}}$ under visual prompting on the matched 800-sample subset. All intervals exclude zero.}
\label{tab:visual_prompting_ci}
\footnotesize
\setlength{\tabcolsep}{3.5pt}
\renewcommand{\arraystretch}{1.04}
\begin{tabular}{@{}lrr@{}}
\toprule
Model & $\Delta$ $\mathcal{I}_{\mathrm{ACC}}$ & 95\% CI of $\Delta$ \\
\midrule
Qwen3-VL-8B-Instruct & \textbf{+25.5} & [19.84, 31.59] \\
GLM-4.5V & \textbf{+18.5} & [13.42, 23.96] \\
Qwen3-VL-32B-Instruct & \textbf{+18.0} & [13.85, 22.12] \\
Qwen2.5-VL-32B-Instruct & \textbf{+22.5} & [18.11, 26.98] \\
Qwen3-VL-30B-A3B-Instruct & \textbf{+24.0} & [16.20, 29.15] \\
InternVL3-38B & \textbf{+10.8} & [6.95, 14.97] \\
InternVL3.5-38B & \textbf{+12.4} & [8.04, 17.17] \\
InternVL3-14B & \textbf{+13.2} & [9.81, 17.01] \\
InternVL3.5-14B & \textbf{+11.2} & [7.09, 15.44] \\
InternVL3-8B & \textbf{+8.6} & [5.29, 12.37] \\
InternVL3.5-8B & \textbf{+6.0} & [3.87, 8.48] \\
\bottomrule
\end{tabular}%
\end{table}

Every visual-prompting interval excludes zero. The positive result also survives
image-clustered two-sided sign-flip tests with Benjamini--Hochberg correction
($q\leq0.0312$ for $\mathcal{I}_{\mathrm{ACC}}$ and $q\leq0.0202$ for
$G_{\mathrm{ACC}}$).

For paired significance, condition labels are exchanged jointly within the
corresponding image or object group, and Benjamini--Hochberg FDR correction is
applied within each family of model comparisons. We attach randomization-test
$p$-values to visual prompting, for which we make a positive claim. For reasoning
and Chain-of-Thought, the paired intervals overwhelmingly include zero and
already support the stated non-improvement conclusion, so additional
randomization-test values would only restate that result. Table~\ref{tab:reasoning_mode_ci} reports the paired reasoning-mode changes for attribute binding and Table~\ref{tab:reasoning_mode_ci_counting} for numerical counting; Table~\ref{tab:cot_ci} reports the corresponding chain-of-thought changes.

\begin{table}[!htbp]

\centering
\caption{Paired reasoning-mode changes and 95\% bootstrap intervals for attribute binding. For Qwen models, $\Delta$ is Thinking minus Instruct; for Gemini models, it is High minus Low reasoning effort. Bold entries have intervals excluding zero.}
\label{tab:reasoning_mode_ci}
\footnotesize
\setlength{\tabcolsep}{4pt}
\renewcommand{\arraystretch}{1.05}
\begin{tabular}{@{}lrr@{}}
\toprule
Model & $\Delta\mathcal{I}_{\mathrm{ACC}}$ (95\% CI) & $\Delta G_{\mathrm{ACC}}$ (95\% CI) \\
\midrule
Q3-8B & \textbf{+5.2 [+0.8, +9.9]} & +4.7 [-2.7, +12.1] \\
Q3-30B-A3B & +1.8 [-0.9, +4.4] & \textbf{+5.0 [+1.9, +7.6]} \\
Q3-32B & -0.1 [-4.4, +7.4] & -2.5 [-9.5, +10.5] \\
G3.1-Pro-Preview & -0.6 [-2.5, +1.1] & +0.1 [-2.4, +2.4] \\
G3.5-Flash & \textbf{+3.6 [+0.6, +5.7]} & -1.9 [-4.8, +0.4] \\
\bottomrule
\end{tabular}%
\end{table}

\begin{table}[!htbp]

\centering
\caption{Paired reasoning-mode changes and 95\% bootstrap intervals for numerical counting, using the same reasoning-mode comparison. Bold entries have intervals excluding zero.}
\label{tab:reasoning_mode_ci_counting}
\scriptsize
\setlength{\tabcolsep}{3pt}
\renewcommand{\arraystretch}{1.05}
\begin{tabular}{@{}lrrr@{}}
\toprule
Model & $\Delta\mathcal{I}_{\mathrm{MRA}}$ (95\% CI) & $\Delta\mathcal{L}_{\mathrm{MRA}}$ (95\% CI) & $\Delta\mathcal{G}_{\mathrm{MRA}}$ (95\% CI) \\
\midrule
Q3-8B & +3.4 [-6.0, +13.5] & \textbf{+22.3 [+11.2, +36.2]} & +5.9 [-4.4, +15.3] \\
Q3-30B-A3B & +1.4 [-4.5, +3.1] & +0.4 [-3.9, +3.3] & +0.8 [-6.2, +7.6] \\
Q3-32B & +2.5 [-8.9, +11.9] & +2.8 [-10.4, +14.3] & -6.2 [-23.1, +7.6] \\
G3.1-Pro-Preview & -4.8 [-8.6, +1.3] & -5.9 [-12.4, +4.1] & +1.2 [-10.7, +11.9] \\
G3.5-Flash & \textbf{+6.4 [+3.2, +10.6]} & \textbf{+13.3 [+9.3, +18.0]} & \textbf{+5.0 [+0.4, +12.3]} \\
\bottomrule
\end{tabular}%
\end{table}

\begin{table}[!htbp]

\centering
\caption{Paired changes under zero-shot and self-consistency chain-of-thought prompting relative to the same base condition. Bold entries have 95\% intervals excluding zero.}
\label{tab:cot_ci}
\footnotesize
\setlength{\tabcolsep}{5pt}
\renewcommand{\arraystretch}{1.04}
\begin{tabular}{@{}lrr@{}}
\toprule
Model & $\Delta$ ZS-CoT vs. Base (95\% CI) & $\Delta$ SC-CoT vs. Base (95\% CI) \\
\midrule
IVL3-14B & \textbf{+4.0 [+0.8, +7.5]} & +6.0 [-2.3, +11.6] \\
IVL3-8B & -1.0 [-4.0, +2.3] & -1.0 [-5.1, +6.5] \\
IVL3.5-14B & +1.0 [-4.3, +5.4] & +0.0 [-14.2, +9.1] \\
IVL3.5-8B & \textbf{+7.0 [+4.3, +9.6]} & +3.0 [-4.8, +8.3] \\
QVL2.5-7B & +0.0 [-4.9, +6.0] & +1.0 [-14.8, +9.6] \\
QVL2.5-32B & -2.0 [-13.2, +6.7] & -3.0 [-10.1, +2.2] \\
QVL3-8B & +4.0 [-5.0, +8.9] & +4.0 [-8.4, +12.0] \\
QVL3-32B & -2.0 [-12.3, +3.6] & -6.0 [-10.3, +1.3] \\
\bottomrule
\end{tabular}%
\end{table}

Reasoning effort does not improve every model--metric combination, and most CoT
intervals contain zero. The only CoT intervals excluding zero are positive
zero-shot effects for IVL3-14B and IVL3.5-8B. For clustered global-versus-local
counting, only Qwen3-VL-8B-Thinking remains significant after correction
($\Delta_{G-L}=-24.6$, 95\% CI $[-39.9,-12.8]$, $p=0.0020$, $q=0.0488$).

Six participants who did not annotate MIMO-Bench each completed 25 AB and 15
NC groups. Human intervals resample groups rather than treating individual
questions as independent. Table~\ref{tab:human_ci} reports the human point estimates and group-resampled intervals.

\begin{table}[!htbp]

\centering
\caption{Human baseline point estimates and group-resampled 95\% bootstrap intervals. The lowest lower bound (77.1\%) remains above the strongest corresponding model result.}
\label{tab:human_ci}
\small
\setlength{\tabcolsep}{7pt}
\renewcommand{\arraystretch}{1.04}

\begin{tabular}{@{}lrr@{}}
\toprule
\textbf{Metric} & \textbf{Human} & \textbf{95\% CI} \\
\midrule
Real $\mathcal{I}_{\mathrm{ACC}}$ & 89.3 & [81.7,93.2] \\
Real $G_{\mathrm{ACC}}$ & 86.7 & [83.2,89.8] \\
Syn. $\mathcal{I}_{\mathrm{ACC}}$ & 85.7 & [77.1,89.6] \\
Syn. $G_{\mathrm{ACC}}$ & 87.5 & [84.9,90.1] \\
Real $\mathcal{I}_{\mathrm{MRA}}$ & 87.7 & [85.1,88.2] \\
Real $\mathcal{L}_{\mathrm{MRA}}$ & 86.4 & [83.2,88.5] \\
Real $\mathcal{G}_{\mathrm{MRA}}$ & 90.1 & [87.9,92.6] \\
Syn. $\mathcal{I}_{\mathrm{MRA}}$ & 89.6 & [86.9,91.9] \\
Syn. $\mathcal{L}_{\mathrm{MRA}}$ & 87.4 & [85.7,90.2] \\
Syn. $\mathcal{G}_{\mathrm{MRA}}$ & 92.0 & [90.1,94.4] \\
\bottomrule
\end{tabular}%
\end{table}
\clearpage
\endgroup

\section{Limitations}\label{app:limitations}

We discuss some of the limitations of our work.

\begin{itemize}
    \item Scaling MIMO-Bench is challenging because of its annotation pipeline. The entire process from bounding box and referring expression creation to question answering and verification relies heavily on humans. Our attempts to semi-automate the generation of referring expressions were unsuccessful, as existing models failed to produce unambiguous descriptions.

    \item {Currently, we attribute the difficulty of MIMO-Bench to the joint effect of same-category distractors and the referring-expression requirement, without isolating either as the dominant factor. Distractors increase the difficulty of the underlying visual reasoning task, while the referring-expression requirement introduces multiple subtasks that the model must solve in order to correctly localize the target object.} However, beyond the mere presence of distractors, their distribution within the scene also plays a significant role in influencing model performance. In future work, we aim to study how the number and spatial arrangement of distractors jointly affect the performance of MLLMs.

\end{itemize}

\end{document}